%% file: colm2026_conference.tex
\documentclass{article} % For LaTeX2e
\usepackage[preprint]{colm2026_conference} % 

\newif\ifarxiv
\arxivtrue

\ifarxiv
  \usepackage[margin=1in]{geometry}
\fi

\usepackage{microtype}
\usepackage[T1]{fontenc}
\usepackage{hyperref}
\usepackage{url}
\usepackage{booktabs}
\usepackage{makecell}
\usepackage[table]{xcolor}
\usepackage{graphicx}
\usepackage{lineno}
\usepackage{amsmath}
\usepackage{amssymb}
\usepackage{wrapfig}
\usepackage{caption}
\usepackage{subcaption}
\usepackage{tikz}
\usepackage{enumitem}

\input{shortcut}

\title{
LLMs Corrupt Your Documents When You Delegate
}

\author{Philippe Laban \quad Tobias Schnabel \quad Jennifer Neville \\
Microsoft Research \\
\texttt{\{plaban, tobias.schnabel, jenneville\}@microsoft.com}
}

\begin{document}

\ifcolmsubmission
\linenumbers
\fi

\maketitle

\begin{abstract}
    Large Language Models (LLMs) are poised to disrupt knowledge work, with the emergence of \textit{delegated work} as a new interaction paradigm (e.g., vibe coding). Delegation requires trust -- the expectation that the LLM will faithfully execute the task without introducing errors into documents. We introduce \benchmark{} to study the readiness of AI systems in delegated workflows. \benchmark{} simulates long delegated workflows that require in-depth document editing across 52 professional domains, such as coding, crystallography, and music notation. Our large-scale experiment with 19 LLMs reveals that current models degrade documents during delegation: even frontier models (Gemini 3.1 Pro, Claude 4.6 Opus, GPT~5.4) corrupt an average of 25\% of document content by the end of long workflows, with other models failing more severely. Additional experiments reveal that agentic tool use does not improve performance on \benchmark{}, and that degradation severity is exacerbated by document size, length of interaction, or presence of distractor files. Our analysis shows that \textbf{current LLMs are unreliable delegates: they introduce sparse but severe errors that silently corrupt documents, compounding over long interaction.}
\end{abstract}

\vspace{-2em}
\begin{center}
\href{https://github.com/microsoft/DELEGATE52}{\symbolimg[0.35cm]{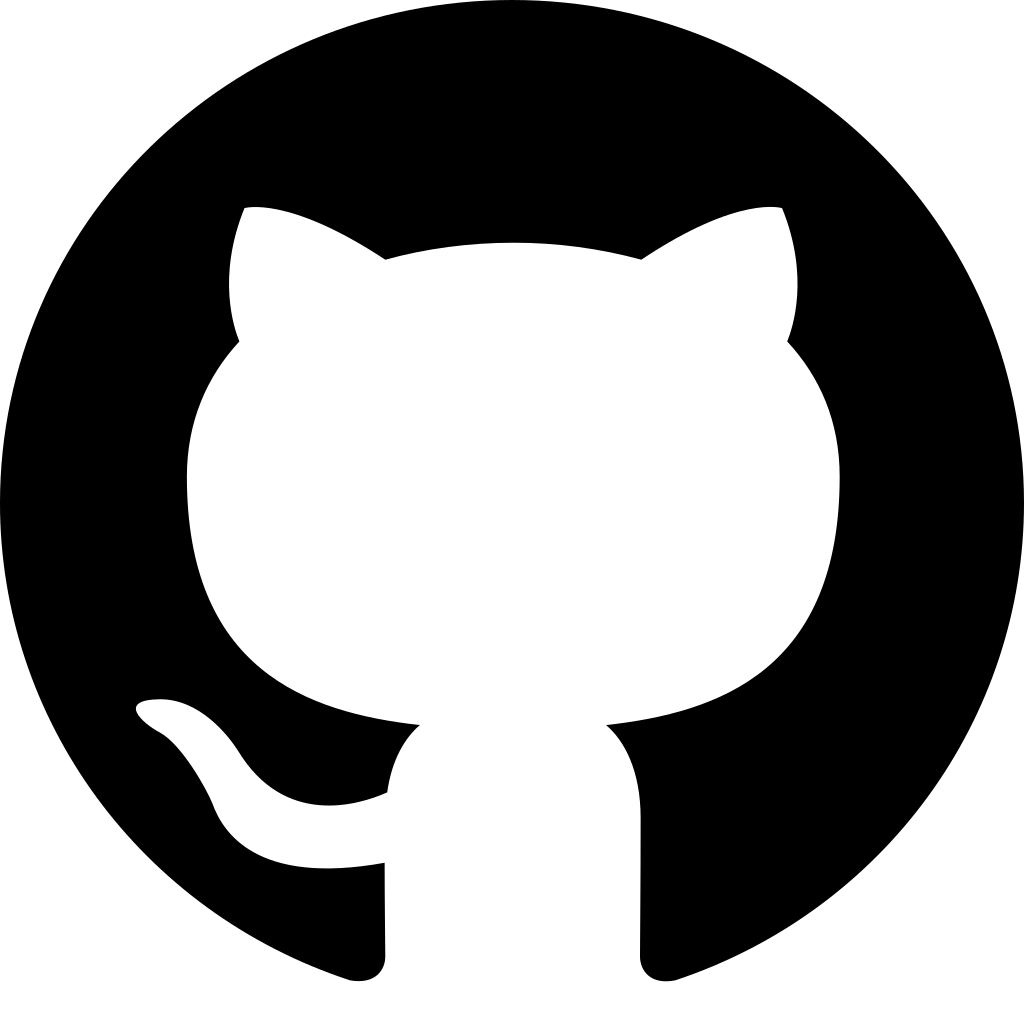}~\texttt{microsoft/DELEGATE52}} \quad \quad \quad \quad  \quad \href{https://huggingface.co/datasets/microsoft/DELEGATE52}{\symbolimg[0.35cm]{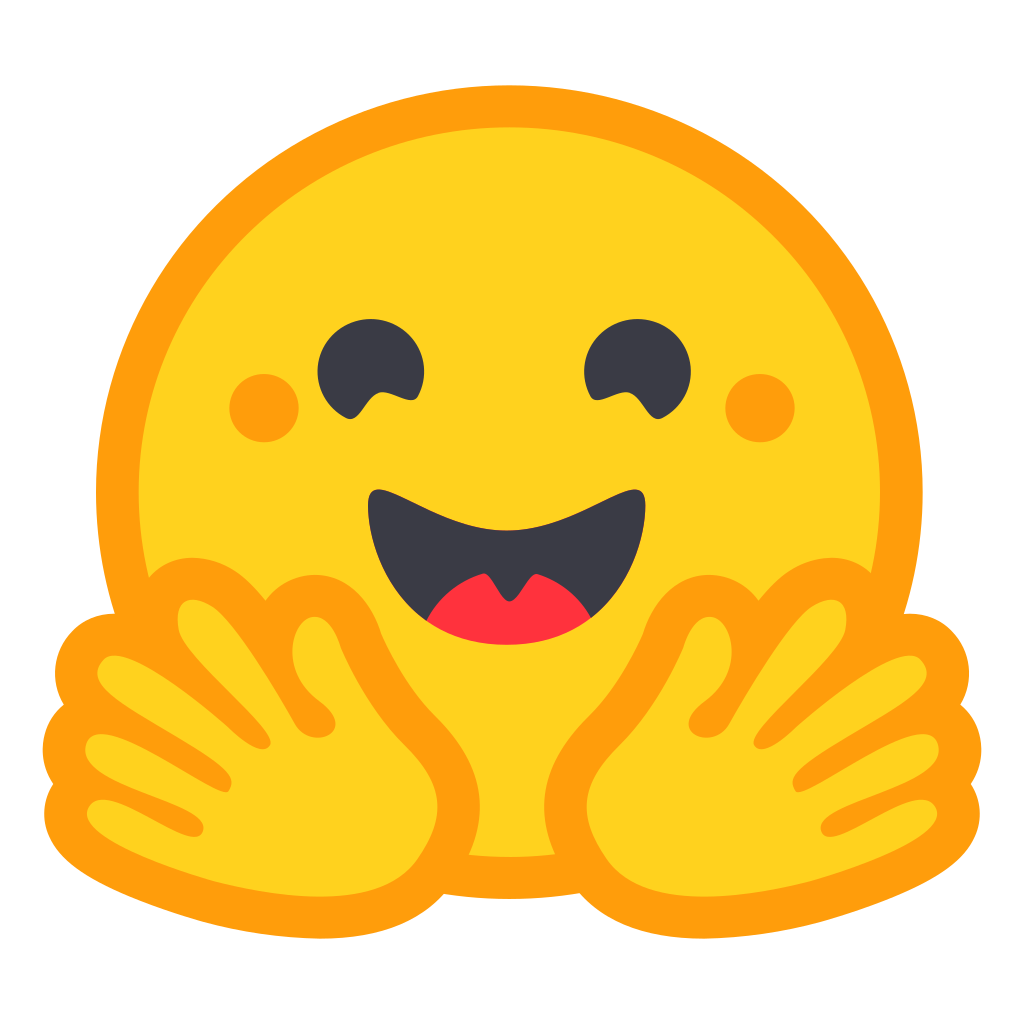}~\texttt{datasets/microsoft/DELEGATE52}}
\end{center}

\vspace{-2mm}
\begin{figure*}[h!]
    \centering
    \includegraphics[width=\textwidth]{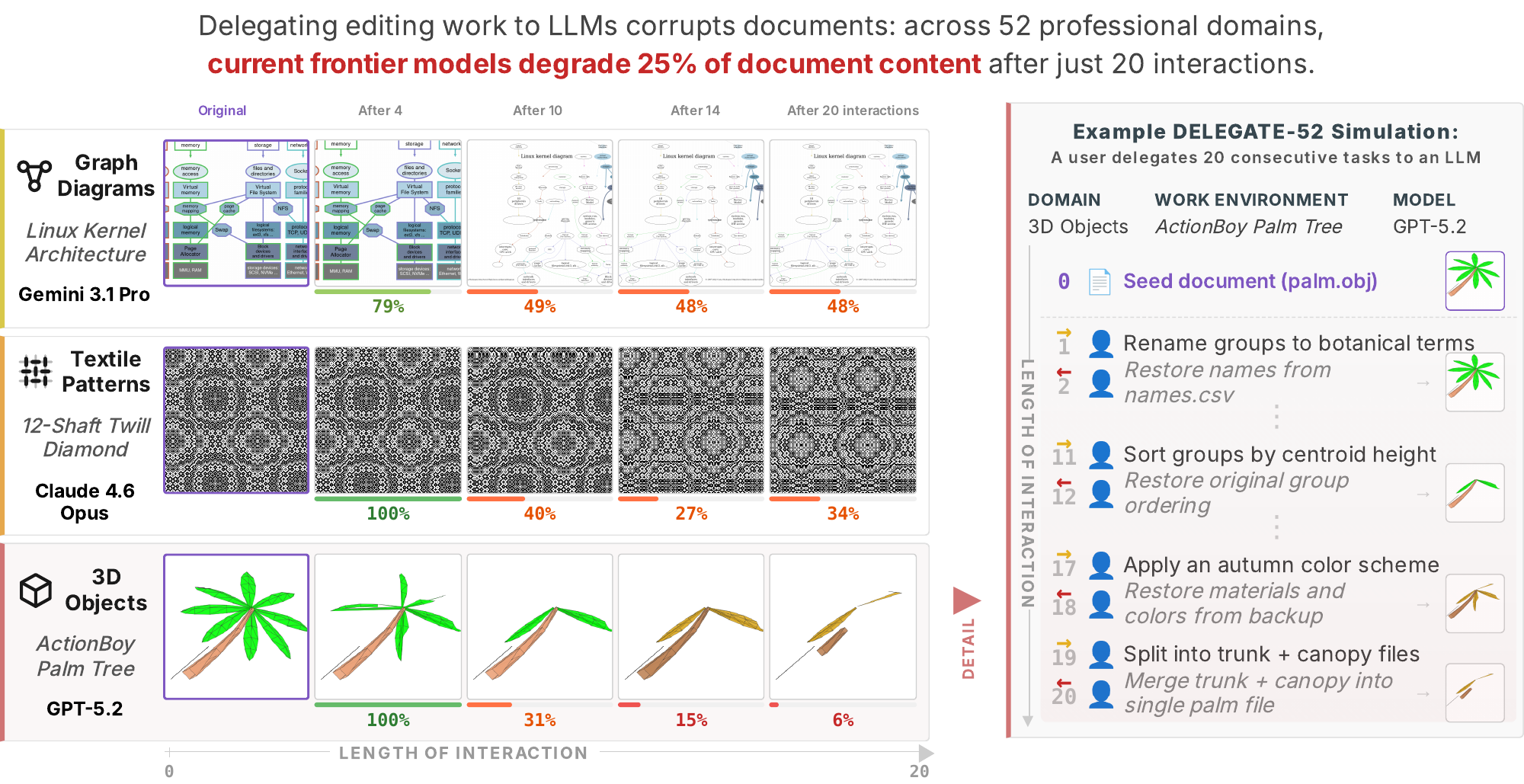}
    \caption{Illustrative examples of how LLMs corrupt documents over long workflows in the \benchmark{} benchmark. As LLMs edit files that represent graph diagrams, textile patterns or 3D objects, they introduce sparse but severe errors that silently corrupt documents, compounding over long interaction.\protect\footnotemark[1]}
    \label{fig:teaser}
\end{figure*}
% \vspace{-2mm}
\footnotetext[1]{\benchmark{} is a text-only benchmark, visual renderings are for illustrative purposes.}

% \vspace{-1.0\intextsep}
\section{Introduction} \label{sec:intro}
% \vspace{-1.0\intextsep}

Recent LLM progress is enabling new interaction paradigms such as \textit{delegated work} \citep{Shao2025FutureOW, Ulloa2025ProductMP}, where knowledge workers supervise LLMs as they complete tasks on their behalf (e.g., ``vibe coding''). Crucially, users delegating work might lack the expertise or time to review changes implemented by the LLM, and must trust that the LLM does not introduce unchecked errors, such as hallucinations, or deletions.

The viability of delegated work hinges on LLMs' ability to carry out tasks and manipulate domain documents without introducing errors. We study, through simulation, the readiness of current LLMs for delegated work across a wide range of professions.

The first contribution of our work is \benchmark{}, a benchmark with 310 work environments across 52 professional domains, including coding, crystallography, genealogy, and music sheet notation. Each environment consists of real documents totaling around 15k tokens in length, and 5-10 complex editing tasks that a user might ask an LLM to carry out. This substantially differs from past work that focuses on tasks within a single domain (e.g., code editing \citep{Cassano2023CanIE} or text editing \citep{Spangher2022NewsEditsAN}).

Our second contribution is the round-trip relay simulation method, which enables us to simulate long-horizon delegated interaction and evaluate LLM performance without requiring annotation or reference solutions. Specifically, we assume every editing task is reversible, defined by a forward instruction and its inverse. Applying both in sequence forms a backtranslation round-trip that, under a perfect model, recovers the original documents exactly. This lets us evaluate performance by measuring document similarity before and after a round-trip. Round-trips can further be composed sequentially, forming a \textit{relay}. Backtranslation originated as a data augmentation and evaluation technique in machine translation \citep{Sennrich2015ImprovingNM,Somers2005RoundtripTW}, and has recently been adapted to evaluate LLM consistency through chained reversible transformations \citep{Hong2025ConsistencyCheckerTE,Allamanis2024UnsupervisedEO}. We repurpose the technique to study long-horizon delegated interaction. % Prior work on evaluating long-horizon AI interactions has focused on assessing the model's ability to recall facts or user preference from previous sessions (e.g., \citep{Xu2021BeyondGM,Maharana2024EvaluatingVL}). In contrast, we focus on the how long-horizon interactions affect {\em document contents}.

Our third contribution is a large-scale simulation with 19 LLMs on \benchmark{}. Our findings show that current LLMs introduce substantial errors when editing work documents, with frontier models (Gemini 3.1 Pro, Claude 4.6 Opus, and GPT~5.4) losing on average 25\% of document content over 20 delegated interactions, and an average degradation across all models of 50\%. Degradation depends on the domain: LLMs perform better in programmatic domains (Python, Database) and worse in natural language and niche domains (e.g., earning statements, music notation). We define a model as ``ready'' for delegated work in a domain if it achieves a score of 98\% or higher after 20 interactions. Python is the only domain (out of 52) where most models are ready, highlighting the significant gap that remains.

Finally, targeted experiments refine our understanding of current LLM capabilities. We confirm that known factors such as document size, interaction length, and distractor context contribute to degradation \citep{Liu2023LostIT, Shi2023LargeLM}, but these negative effects compound over time, meaning short simulations underestimate their severity. We also find that using a basic agentic harness does not improve the performance of LLMs we test on \benchmark{}, and that performance after two interactions is not predictive of long-horizon performance (20 interactions), validating the importance of long-horizon evaluation. We release \benchmark{} publicly as a tool to monitor AI readiness for delegated work and drive research on long-horizon Human-AI interaction.

\vspace{-1.0\intextsep}
\section{The \benchmark{} Benchmark} \label{sec:delegate52}
% we need to leave it here, otherwise it goes into the bottom of the page
\begin{wrapfigure}{r}{0.35\textwidth}
    \centering
    \vspace{-4\intextsep}
    \includegraphics[width=0.35\textwidth]{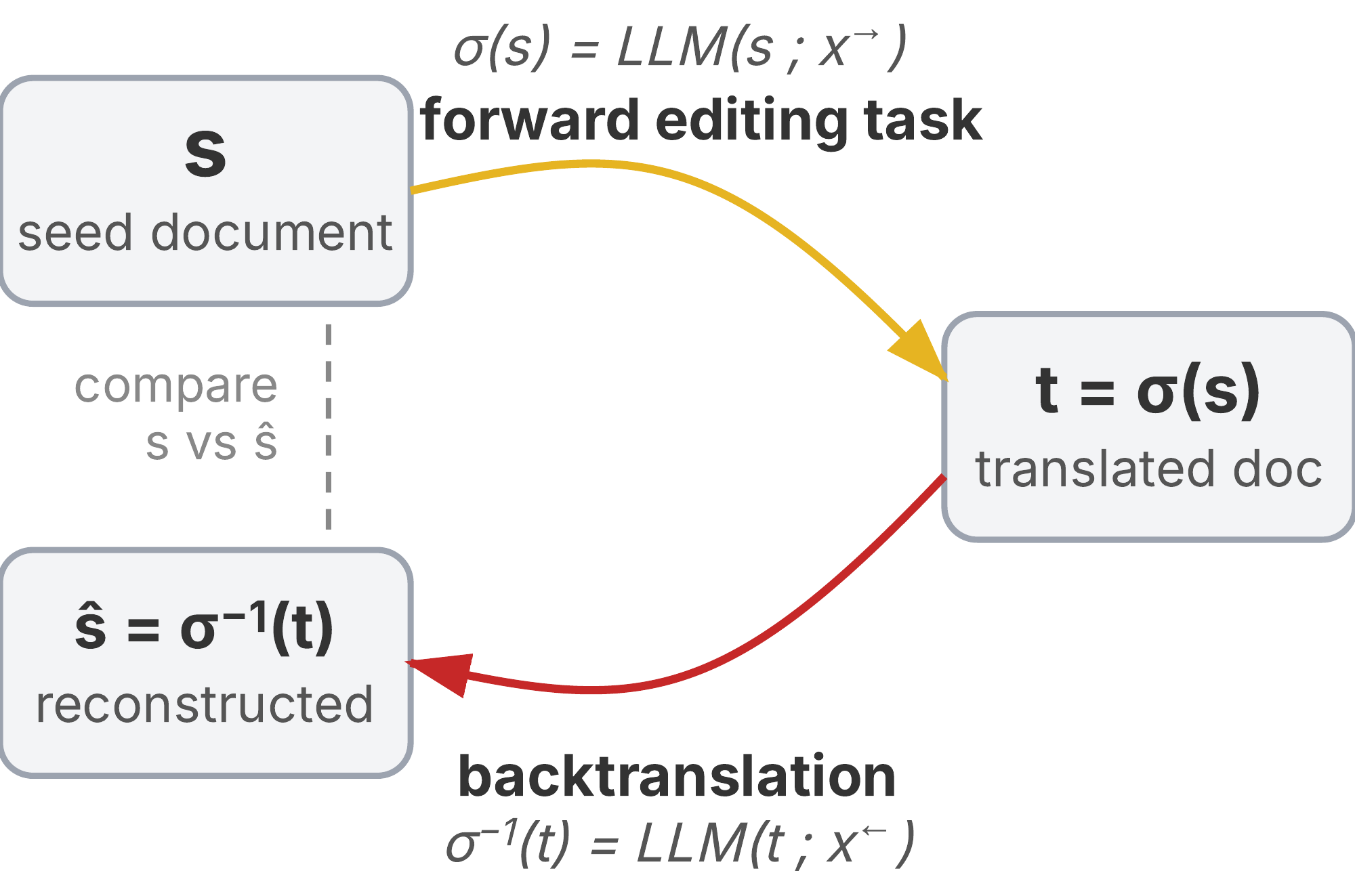}
    \caption{The backtranslation round-trip primitive.}
    \label{fig:round_trip_primitive}
    \vspace{-2\intextsep}
\end{wrapfigure}

In \benchmark{} we simulate long workflows that could be part of a knowledge worker's tasks. A workflow consists of seed documents along with other content that are transformed via a sequence of complex editing tasks, mirroring the iterative nature of delegated work. We now introduce the framework that allows us to (i) perform evaluation automatically and (ii) scale the length of workflows.

\subsection{Evaluating Without References} \label{sec:def_round_trip}

Figure~\ref{fig:round_trip_primitive} illustrates the round-trip primitive made up from a pair of editing tasks, inspired by backtranslation~\citep{Somers2005RoundtripTW}. Given a seed document $s$, we can define a pair of forward and backward edit instructions ($x^{\rightarrow}, x^{\leftarrow}$) that describe in natural language a transformation of the seed document and its inverse ($\sigma$, $\sigma^{-1}$). First, an LLM applies a forward instruction to the seed document, producing a transformed document $t=\sigma(s)=\textsf{LLM}(s; x^{\rightarrow})$. Second, the LLM applies the backward instruction to the transformed document, producing a reconstructed document $\hat{s}=\sigma^{-1}(t) = \textsf{LLM}(t; x^{\leftarrow})$. Each step is conducted as an independent, single-turn session.

% At the heart of the benchmark lies the idea of backtranslation on a seed document $s$ as illustrated in Figure~\ref{fig:round_trip_primitive}: First, the document is translated forward with an edit instruction $x^{\rightarrow}$ via $t=\sigma(s)=\textsf{LLM}(s; x^{\rightarrow})$, then, a back translation is done with edit instruction $\hat{s}=\sigma^{-1}(t) = \textsf{LLM}(t; x^{\leftarrow})$ resulting in a reconstructed document $\hat{s}$. An important assumption we make is that the forward translation an LLM does cannot be a no-op, i.e., $\sigma(s) \neq s$, otherwise the backtranslation would be trivial as well. This is ensured by xxx.

To measure reconstruction quality, we implement a domain-specific similarity function $\mathrm{sim}(s_i, s_j)$. A perfect model yields $\mathrm{sim}(s, \hat{s}) = 1$, reducing evaluation to semantic equivalence without reference annotations. For backtranslation to be aligned with model performance, models need to genuinely attempt the editing instructions rather than taking shortcuts; we validate this in Appendix~\ref{app:instruction_compliance}. Appendix~\ref{app:backtranslation} discusses other properties, assumptions, and limitations of this framework.

\paragraph{Simulating long workflows.}
Since each round-trip is designed to return to the seed document $s$, round-trips can be chained into longer workflows. We sample $N$ pairs of forward and backward instructions $(x_1^{\rightarrow}, x_1^{\leftarrow}), \ldots, (x_N^{\rightarrow}, x_N^{\leftarrow})$ from the set of available options, each representing a transformation $\sigma_i(s)$. We simulate an $n$-relay by applying $n$ round-trip edits in sequence:
% \vspace{-\baselineskip}
$$\hat{s}_k = \left(\sigma_1 \circ \sigma_1^{-1} \circ \cdots \circ \sigma_n \circ \sigma_n^{-1} \right)(s), \;\;\; 1\leq n\leq N.$$
Our main metric is the reconstruction score after $k$ interactions (i.e., $k/2$ round-trips):
$$\textrm{RS}@k(s) = \mathrm{sim}(s, \hat{s}_{k/2}).$$
% \begin{equation}
%     \textrm{RS}@k(s) = \mathrm{sim}(s, \hat{s}_k).
% \end{equation}

\subsection{Benchmark Construction}

\begin{figure*}[t]
    \centering
    \includegraphics[width=\textwidth]{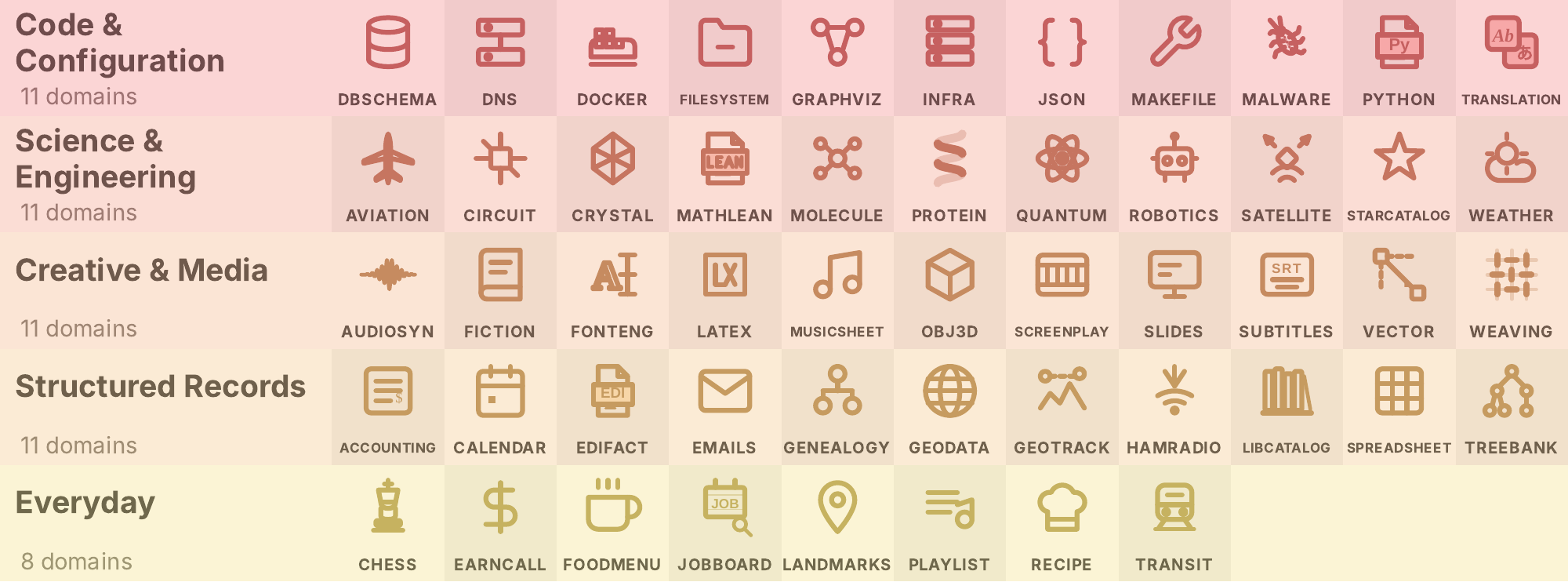}
    \caption{\benchmark{} includes work environments from 52 professional domains in five categories: Science \& Engineering, Code \& Configuration, Creative \& Media, Structured Records, and Everyday.}
    \label{fig:domain_categories}
\end{figure*}

%\subsubsection{Domains} \label{sec:domain_selection}
We selected 52 professional domains to simulate workflows (listed in Figure~\ref{fig:domain_categories}), representing diverse knowledge work professions across five categories: Science \& Engineering, Code \& Configuration, Creative \& Media, Structured Record, and Everyday. A key criterion for inclusion is the existence of a \textit{standard document type} that is textual and unencoded (e.g., .srt for subtitles, .cif for crystallography). Secondary considerations in domain selection are listed in Appendix~\ref{app:domain_brainstorm}.

\subsubsection{Work Environments} \label{sec:work_environments}

For each domain, we construct six work environments consisting of a seed document, a set of 5-10 possible edit tasks, and a distractor context. An example environment for the accounting domain is presented in Figure~\ref{fig:sample_viz}, and environment creation is detailed in Appendix~\ref{app:dataset_creation}.

\begin{figure*}[t]
    \centering
    % \vspace{-2\intextsep}
    \includegraphics[width=1.0\textwidth]{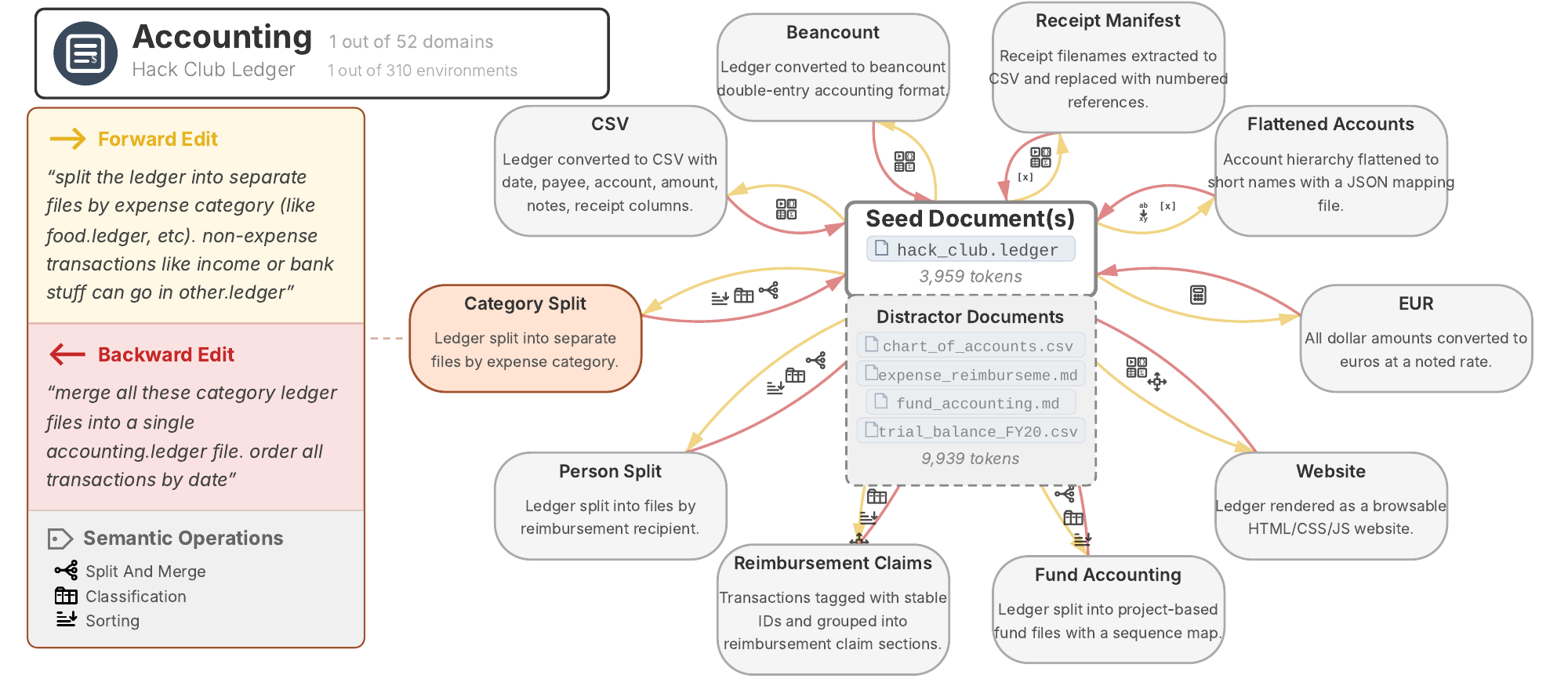}
    \caption{Example work environment from the accounting domain in \benchmark{}. The seed document is an accounting ledger of Hack Club, a non-profit organization. The highlighted edit (Category Split) consists in first splitting the seed document \texttt{hack\_club.ledger} into separate files by expense category (forward edit task), and merging it back chronologically into one file (backward edit task).}
    \label{fig:sample_viz}
    % \vspace{-1\intextsep}
\end{figure*}

\vspace{-1.0\intextsep}
\paragraph{Seed Documents.} The seed document is the starting point for all simulations. Seed documents are real documents found online (no synthetic data, exemplars, or templates), range from 2--5k tokens\footnote{Based on the GPT~4 tiktoken encoder}, and have a permissive license for redistribution. Secondary requirements are listed in \autoref{app:document_desiderata}. The simulations in Figure~\ref{fig:teaser} use three seed documents: a Linux Kernel Architecture Diagram (graph), a 12-shaft Twill Diamond Pattern (textile), and the ActionBoy Palm Tree (3D objects).

\vspace{-1.0\intextsep}
\paragraph{Edit Tasks.} Edit tasks are pairs of forward and backward instructions defining invertible transformations. The instructions must: (1) represent realistic work tasks that a stakeholder might perform given the document, (2) require in-depth, non-trivial transformation of the context that goes beyond expansion. In other words, $\sigma(s)$ cannot be decomposed into $[s, \sigma'(s)]$ (concatenation), as this would make the backward edit trivial (cropping). Each edit task is tagged with the semantic operations required to perform the edit (e.g., numerical reasoning, classification, splitting). The accounting work environment in Figure~\ref{fig:sample_viz} has 10 edit tasks, including tasks that require splitting the ledger into separate files by expense category or reimbursement recipient, converting the amounts to Euro, or formatting the ledger in Beancount format. Appendix~\ref{app:env_scaling} describes the edit creation and tagging process.

\vspace{-1.0\intextsep}
\paragraph{Distractor Context.} In realistic work settings, retrieved or available documents are not always relevant to the task at hand (i.e., retrieval precision is imperfect). To simulate this, each work environment includes a distractor context: topically related documents that do not interfere with any of the editing tasks. In the accounting example of Figure~\ref{fig:sample_viz}, the distractor context includes a chart of accounts, the organization expense reimbursement policy, and three other documents from the organization. Distractor contexts range from 8--12k tokens per environment, and are included by default in experiments to enhance simulation realism. Distractor construction and non-interference validation are detailed in Appendix~\ref{app:distractor_qa}.

\subsubsection{Domain-Specific Evaluation} \label{sec:domain_specific_evaluation}

\begin{figure*}[t]
    % \vspace{-2\intextsep}
    \centering
    \includegraphics[width=\textwidth]{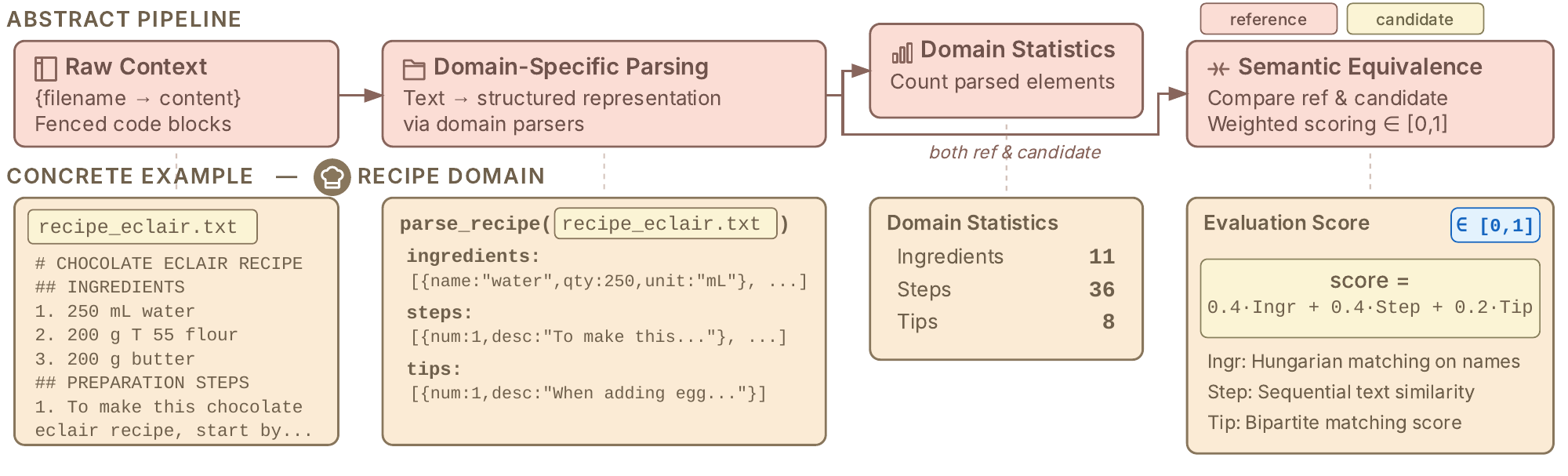}
    \caption{Top: Domains in \benchmark{} implement a parsing function that converts text documents into a structured representation which is then used by a similarity function to score two parsed instances. Bottom: concrete example for the recipe domain.}
    \label{fig:domain_anatomy}
    % \vspace{-1\intextsep}
\end{figure*}

Common textual similarity methods consider either low-level overlap (e.g., Levenshtein ratio \citep{Levenshtein1965BinaryCC}) or semantic distance in a generic embedding space~\citep{Neelakantan2022TextAC}. These do not adequately capture fine-grained semantic changes, so we implement a custom similarity function for each domain, illustrated in \autoref{fig:domain_anatomy}.

Semantic equivalence is measured in two steps: parsing and evaluation. A parsing function converts documents into a structured representation. In \autoref{fig:domain_anatomy}, a recipe is parsed into ingredients (names, quantities, units), steps, and tips. A similarity function then compares two parsed representations and outputs a score in $[0, 1]$. In the recipe domain, similarity is a weighted sum over ingredient lists (40\%), steps (40\%), and tips (20\%). Per-domain component combination and relative weights are calibrated through ablation testing to ensure proportional sensitivity to content loss or corruption (Appendix~\ref{app:domain_creation}).

This flexibility allows for a domain-appropriate weighing of various components of the scoring function. For instance a small surface-level change in an ingredient (e.g., 200 $\to$ 800 g of butter) can severely impact the overall score (as desired). Conversely, the domain-specific parsing allows for robustness in the scoring function: surface-level changes that do not impact semantics (e.g., 200g vs. 0.2kg of butter, or shuffling the order of the ingredient list) do not affect the score.

Implementing robust semantic equivalence for 52 domains is central to our methodology. In Appendix~\ref{app:alt_eval}, we show that generic similarity measures (including LLM-as-a-judge with GPT~5.4) fail to capture nuanced semantic differences, only moderately correlating with our parsing-based metric and capturing at most 25\% of the variance.

\subsubsection{Quality Assurance.} \label{sec:benchmark_qa}

To ensure experimental validity, we performed quality assurance at each stage of the construction process (Appendix~\ref{app:dataset_creation}), evaluating (1) parsing robustness, (2) evaluation sensitivity, (3) edit testing, and (4) distractor interference.

\section{Experiments} \label{sec:experiments}

\begin{wrapfigure}{l}{0.35\textwidth}
    \centering
    \vspace{-1\intextsep}
    \includegraphics[width=0.35\textwidth]{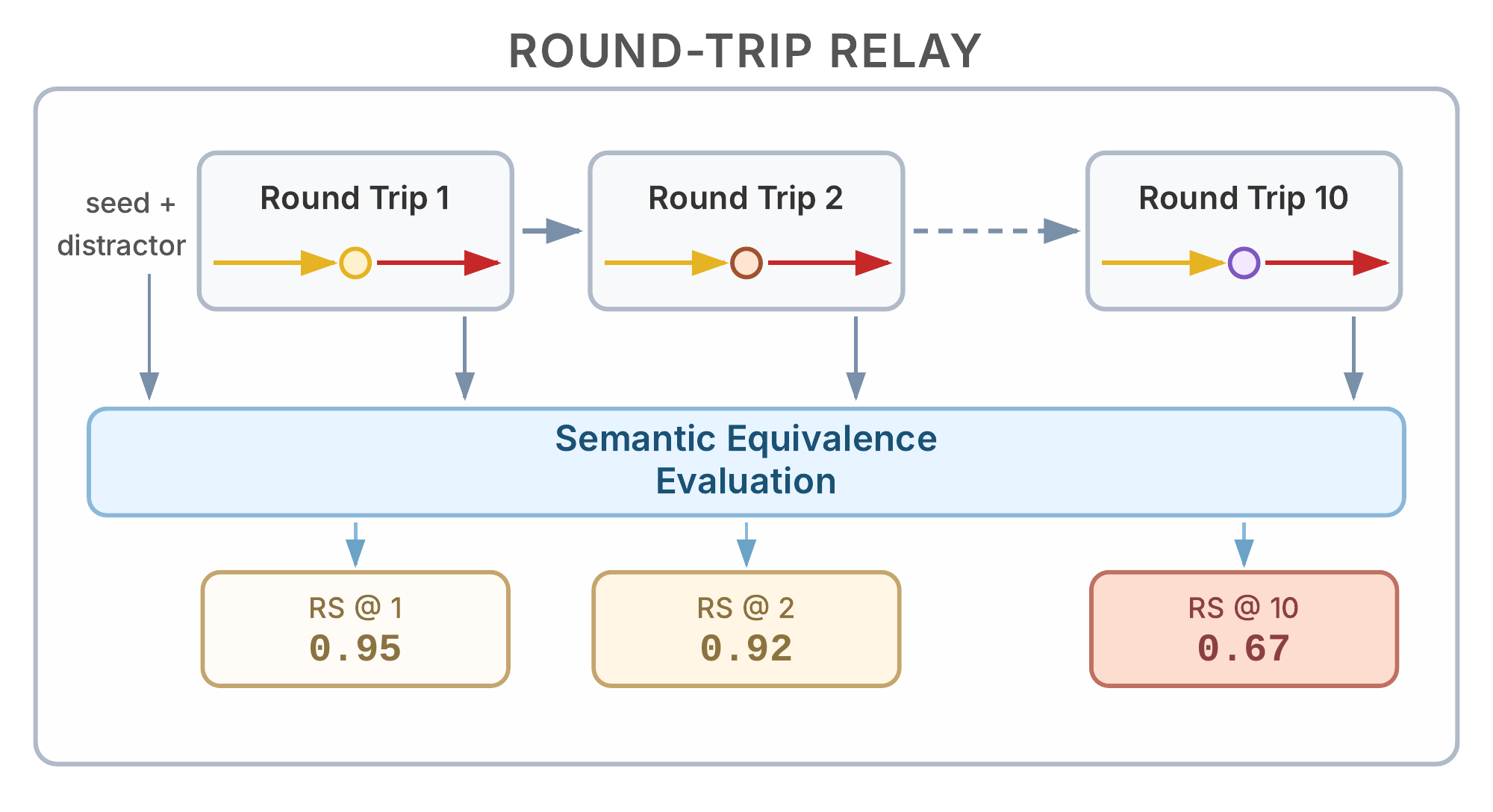}
    \caption{A \textit{round-trip relay}: a sequence of 10 consecutive round-trip tasks, total: 20 interactions.}
    \label{fig:twenty_interaction_simulation}
    \vspace{-1\intextsep}
\end{wrapfigure}

\paragraph{Experimental Setup.} \label{sec:exp_design}
Our main experiment is a \textit{round-trip relay} with $N=10$ consecutive round-trips per environment, simulating 20 delegated interactions. In each interaction, the model receives all work environment documents as text in its context window in a single turn (unless stated otherwise in the agentic experiments of Section~\ref{sec:agentic}). Since most of the constructed environments have fewer than 10 editing tasks, we repeat edits in round-robin fashion (shuffling order at each epoch) to reach 10 round-trips. We compute reconstruction scores RS@$k$ after each round-trip, estimating degradation every two interactions. We validate the use of round-robin scheduling in Appendix~\ref{app:round_robin}, showing it is more realistic and leads to more degradation than repeating the same edit across all rounds of a relay.

\paragraph{Model Selection.} \label{sec:model_selection}
We select 19 LLMs from six model families: \openai~OpenAI (GPT~4o, GPT~4.1, GPT~5 Nano, GPT~5 Mini, GPT~5 Chat, GPT~5, GPT~5.1, GPT~5.2, GPT~5.4, o1, o3, and GPT~OSS 120B), \claude~Anthropic (Claude 4.6 Sonnet and Claude 4.6 Opus), \gemini~Google Gemini (3 Flash and 3.1 Pro), \mistral~Mistral (Large 3), \xai~xAI (Grok 4), and \moonshot~Moonshot (Kimi K2.5). The selection spans a wide range of capabilities, from smaller to frontier models, enabling us to study how model scale and architecture influence degradation in delegated work. Exact model versions in Appendix~\ref{app:model_details}.

\vspace{3\intextsep}

\section{Results} \label{sec:results}

\begin{wraptable}{r}{0.6\textwidth}
    \vspace{-7\intextsep}
    \input{tables/overall_by_round}
    \caption{Round-trip relay results for 19 LLMs on \benchmark{} (20 interactions). All models accumulate errors, leading to significant document corruption. Background color: degradation from the seed document.}
    \label{tab:overall_by_round}
    \vspace{-1\intextsep}
\end{wraptable}

\subsection{Main Results} \label{sec:main_results}

Table~\ref{tab:overall_by_round} details simulation results. At a high level, \textbf{every model sees its performance degrade over the course of interaction, with average degradations of 50\% across tested models by the end of simulation}. Even frontier models (Gemini 3.1 Pro, Claude 4.6 Opus, GPT~5.4) degrade documents by 25\% on average over 20 interactions.
% The degradation is concave:
% The degradation dynamics are generally non-linear and concave: earlier interactions cause more severe degradation, and as the document degrades, the marginal degradation from each additional interaction shrinks. However, stronger models exhibit more gradual degradation (round 1-5 / round 5-10 degradation ratios of 3-5x), whereas weaker models often see more extreme degradation from the onset (11-17x).

A per-domain breakdown of end-of-simulation scores (Table~\ref{tab:round10_by_domain}) reveals that \textbf{models are not ready for delegated workflows in the vast majority of domains, with models severely corrupting documents (at least -20\% degradation) in 80\% of our simulated conditions.} The Python domain is an outlier: a majority of tested models (17/19) achieve lossless manipulation, resonating with recent findings on delegated coding workflows \citep{Pimenova2025GoodVA}. The top model (Gemini 3.1 Pro) is designated as ready (RS@20 $\geq$ 98\%) in 11 of 52 domains.

We find that short-term performance (after 2 interactions) is not always predictive of long-horizon performance. For instance, GPT~5 and Kimi K2.5 achieve near-identical performance after two interactions (91.5 vs. 91.1) but diverge sharply over time (ending at 48.3 vs. 64.1). Conversely, Gemini 3 Flash trails Mistral Large 3 by 6.4 points early on (76.0 vs. 82.4) but overtakes it by end of simulation (35.8 vs. 35.5). \textbf{In other words, short interaction simulations are insufficient to understand long-horizon LLM performance, validating the importance of benchmarks that simulate extended interactions.}

% The results should be interpreted for the scale
We caution the reader to interpret the absolute scores with respect to the scale of our experimental setting. LLMs are tested in a simulation environment that requires completing work on documents with 3-5k tokens, as well as a distractor context of 8-12k tokens, over the course of 20 interactions. Upcoming subsections use a subset of GPT-family models to study how tool use, document size, interaction length, and distractors affect degradation.

\begin{table}[h!]
\vspace{-1\intextsep}
\input{tables/round10_by_domain}
\caption{Visual histogram of end of simulation scores (after 20 interactions), broken down across the 52 domains in \benchmark{}. Scores are binned into buckets: 
\chbox{D6EEFF}{\checkmark\,$\geq$98} (``\textit{ready}'')\;
\chbox{FFF9EC}{95--98}\;
\chbox{FEF2CC}{90--95}\;
\chbox{FFE3D3}{80--90}\;
\chbox{F9C6C6}{70--80}\;
\chbox{E8A0A0}{55--70}\;
\chbox{D48A8A}{$<$55}. Catastrophic corruption (80 and below) occurs in more than 80\% of model, domain combinations. Python is the only domain where a majority of models achieve the ready status, and the best model (Gemini 3.1 Pro) is ready for delegated workflows in only 11 out of 52 domains.}
\label{tab:round10_by_domain}
\vspace{-1\intextsep}
\end{table}

\subsection{Agent (With Tools) vs. LLM (Without Tools)} \label{sec:agentic}

In the main experiment, models operate \textit{without tools}, directly outputting modified files. In principle, tool use could reduce degradation by enabling models to make targeted, programmatic modifications (e.g., via search-and-replace or code execution) rather than regenerating entire documents, reducing the risk of inadvertent content corruption. To test this, we implemented a basic agentic harness \citep{Yao2022ReActSR} with file reading, writing, and code execution tools (Appendix~\ref{app:agentic_harness}). We note this is not an optimized state-of-the-art agent system; future work could explore more sophisticated harnesses.

The results are summarized in Table~\ref{tab:agentic_scores}. The four tested models perform worse when operated agentically with tools than without, incurring an average additional degradation of 6\% by the end of simulation. The best-performing model (GPT~5.4) narrows the gap with an additional degradation of only 3\% (71.5\% vs. 68.3\%).

At first glance, this seems counter-intuitive: tools should give LLMs an advantage. However, there are several factors at play. First, models incur an overhead when using tools (see Table~\ref{tab:agentic_behavior}) due to the interactive nature of the agentic harness; they invoke 8-12 tools on average to complete a task, consuming 2-5x more input tokens than the no-tool alternative. Preserving LLM performance in long context settings is a known challenge for current LLMs~\citep{Liu2023LostIT, Laban2024SummaryOA}. Second, \benchmark{} does not contain tasks that can be trivially completed by executing a short program (such as sorting a spreadsheet) as this would not be representative of a task a user would delegate to an LLM. Tasks can involve computation, but must also require textual understanding and reasoning over the documents. This explains why even in the agentic settings, LLMs favor the file writing tool over code execution (see Table~\ref{tab:agentic_behavior}), which limits the benefits of the agentic harness. Looking at the trend: better models rely more on code execution (10\% for GPT~4.1 vs. 45\% for GPT~5.4), leading to more efficiency in the use of the agentic harness.

In short, \textbf{under our basic harness, the tested LLMs do not benefit from agentic tool use when completing complex editing tasks in diverse textual domains.} This suggests \benchmark{} can serve agentic system developers: it provides diverse domains with complex editing tasks where current LLMs struggle to leverage tooling for precise manipulation.

\subsection{Document Size Effect} \label{sec:context_size}

\begin{table*}[t]
% \vspace{-2\intextsep}
\begin{minipage}[t]{0.48\textwidth}
\vspace{0pt}
\centering
\input{tables/agentic_scores}
\caption{Tool-use effect on degradation for four LLMs, comparing agentic (\iconagentic) and direct (\icondirect) operation. All models degrade documents more with tools than without.}
\label{tab:agentic_scores}
\end{minipage}\hfill
\begin{minipage}[t]{0.48\textwidth}
\vspace{0pt}
\centering
\input{tables/agentic_behavior}
\caption{Tool-use behavior. Tool use leads to overhead (with tools / no tools): models consume more input tokens (Inp$\times$), produce fewer output tokens (Out$\times$), cost more (\$$\times$), with typically higher latency (Lat$\times$). To edit documents, models can execute code (\iconexec) or write files manually (\iconwrite). \%Dist: the portion of distractor files read.}
\label{tab:agentic_behavior}
\end{minipage}
% \vspace{-2\intextsep}
\end{table*}

The main experiment used 3--5k token documents to isolate degradation from long-context effects. We now study how document size affects degradation (details in Appendix~\ref{app:context_size_exp}).

\begin{table*}[t]
% \vspace{-2\intextsep}
\begin{minipage}[t]{0.48\textwidth}
\vspace{0pt}
\centering
\input{tables/context_size}
\caption{Document size effect on degradation for GPT~5.4. Larger documents degrade more, with the gap widening from 2 to 20 interactions.}
\label{tab:context_size}
\end{minipage}\hfill
\begin{minipage}[t]{0.48\textwidth}
\vspace{0pt}
\centering
\input{tables/fifty_rounds}
\caption{Interaction length effect on degradation, extending relays to 100 interactions. All models show monotonic decline, with no signs of plateauing.}
\label{tab:fifty_rounds}
\end{minipage}
\vspace{-1\intextsep}
\end{table*}

Results from document size variation are summarized in Table~\ref{tab:context_size}. In short, as document size is increased from 1k to 10k tokens, GPT~5.4 degradation levels worsen gradually, with scores at the 10k scale of 59.9\% by end of simulation. Each additional 1,000 tokens in a document degrades GPT~5.4's ability to preserve content by roughly 0.7\% after two interactions, but 3.6\% after 20 interactions: a {\raise.17ex\hbox{$\scriptstyle\sim$}}5-fold increase over the course of interaction. In a nutshell, \textbf{document size and interaction length compound multiplicatively: the degradation from increased document size snowballs over the course of the interaction.}

\subsection{Length of Interaction} \label{sec:fifty_rounds}

The main experiment uses a 10-round-trip relay (20 interactions). In Table~\ref{tab:fifty_rounds}, we extend relays for a subset of models to 50 round-trips (100 interactions). We did not create additional edits, and simply repeat existing edits in round-robin fashion.

We find that degradations continue to accumulate in longer relays, with none of the models showing signs of plateauing. The rate of degradation decelerates: the first half of the extended relay (round-trips 5--25) accounts for roughly 2--3$\times$ more loss than the second half (25--50), but even the strongest model (GPT~5.4) drops below 60\% by the end of a 50-round-trip relay. In summary, \textbf{as we extend relays from 10 to 50 round-trips, performance continues to degrade, with models introducing novel errors even when tasks repeat.}

\subsection{Distractor Effect} \label{sec:distractor}

Experiments so far include distractor documents during simulation: this enacts a realistic work environment where retrieved documents are not all necessary to complete the task (i.e., imperfect retrieval precision). We ablate the experiment by running simulations that exclude distractor documents. This simplifies the setting: the LLM is provided exactly the documents it must edit, without having to judge information relevance.

Table~\ref{tab:distractor_effect} summarizes the results for four models, contrasting performance of each model with distractor documents included or excluded. Looking at initial steps in the simulation (2 interactions), removing distractor documents has a small positive effect, improving scores by 0.4--4\%. However, over the course of interaction the effect of distractors widens, and we observe improvements of 2--8\% by the end of the simulation. In other words, \textbf{distractor harm compounds with interaction length, and measuring short-term effect of distractors likely underestimates their effects in long, realistic interactions.} This finding echoes prior work on irrelevant context distraction \citep{Shi2023LargeLM}, and extends it to a long-horizon setting. This is relevant to retrieval system evaluation: long-horizon benchmarks can capture lasting effects of improved retrieval on performance.

% A second interesting observation from the results is that stronger models benefit the most from the removal of distractors compared to weakest tested model (GPT5.4: +6.3\% gain; GPT~4.1: +2.8\% gain). Multiple hypotheses could explain this finding. First, the weaker model's performance degrades rapidly independent of distractors, and might be reaching a floor that is less susceptible to further degradation from distractors. Second, the finding could be related to recent findings from the community finding that stronger models are more susceptible to being jail-broken, due to their improved instruction-following capabilities~\citep{Wei2023JailbrokenHD}. In other words, stronger models might be digesting distractor documents more deeply, and attempting to connect them to the task in a way weaker models do not, leading to deepened degradation. We however caution that the small scale of this experiment does not allow for strong conclusions.

\subsection{Delegation Beyond Textual Documents} \label{sec:image_domain}

\begin{table*}[t]
\vspace{-2\intextsep}
\begin{minipage}[t]{0.48\textwidth}
\vspace{0pt}
\centering
\input{tables/distractor_effect}
\caption{Distractor effect on degradation, with distractor (top) and without (indented). Removing distractors consistently improves scores across all models and rounds.}
\label{tab:distractor_effect}
\end{minipage}\hfill
\begin{minipage}[t]{0.48\textwidth}
\vspace{0pt}
\centering
\input{tables/image_domain}
\caption{Image editing relay results for nine image generation models. Current models degrade images significantly faster than LLMs degrade text.}
\label{tab:image_domain}
\end{minipage}
\vspace{-1\intextsep}
\end{table*}

To test whether our methodology extends beyond text, we implemented six visual work environments simulating image editing workflows (details in Appendix~\ref{app:image_domain}), testing 9 models with image generation capabilities across up to 20 interactions.

Example edit relay outputs are shown in a gallery in Figure~\ref{fig:image_domain_gallery}, and reconstruction scores are summarized in Table~\ref{tab:image_domain}. We observe that degradations for image manipulation are severely more pronounced than for textual domains. The best models achieve final reconstruction scores of 28-30\%, compared to 70--80\% for textual domains. Even after two interactions, no image generation model exceeds 65\%, worse than text models after 20 interactions. This small-scale experiment suggests that \textbf{image editing models degrade documents far more severely than text models, and are not ready for delegated work.} This proof-of-concept shows our methodology extends to non-textual modalities.

\section{Analysis} \label{sec:understanding_degradation}

\begin{wraptable}{r}{0.5\textwidth}
    \vspace{-8\intextsep}
    \input{tables/critical_errors_compact}
    \vspace{-0.4\intextsep}
    \caption{Critical error analysis: cumulative \% of runs with at least one critical error ($\geq$10pt drop) by interaction, and share of total degradation from critical errors.}
    \label{tab:critical_errors_main}
    \vspace{-1\intextsep}
\end{wraptable}

\paragraph{Critical Failures (Appendix~\ref{app:critical_errors}).} The main results (Table~\ref{tab:overall_by_round}) average degradations across hundreds of simulations for each model, giving an impression of smooth degradation curves, with each interaction leading to a small amount of additional degradation. To look beyond this aggregate view, we analyzed the dynamics of individual relay simulations. We categorize each round-trip as introducing a critical failure if it led to a drop in score of at least 10\%. Table~\ref{tab:critical_errors_main} summarizes the result of the analysis, reporting for each model the likelihood of a critical error after $N$ interactions, and the proportion of total error attributable to critical errors. We find that models are not failing due to ``death by a thousand cuts'' (i.e., many small errors). Instead, they maintain near-perfect reconstruction in some rounds, and experience critical failures in a few rounds, typically losing 10-30+ points in a single round trip. \textbf{These sparse critical failures explain about 80\% of total document degradation we observe.} Stronger models do not avoid small errors better; they delay critical failures and experience them in fewer interactions. \vspace{6\intextsep}

\begin{wrapfigure}{l}{0.45\textwidth}
    \centering
    \vspace{-2\intextsep}
    \includegraphics[width=0.45\textwidth]{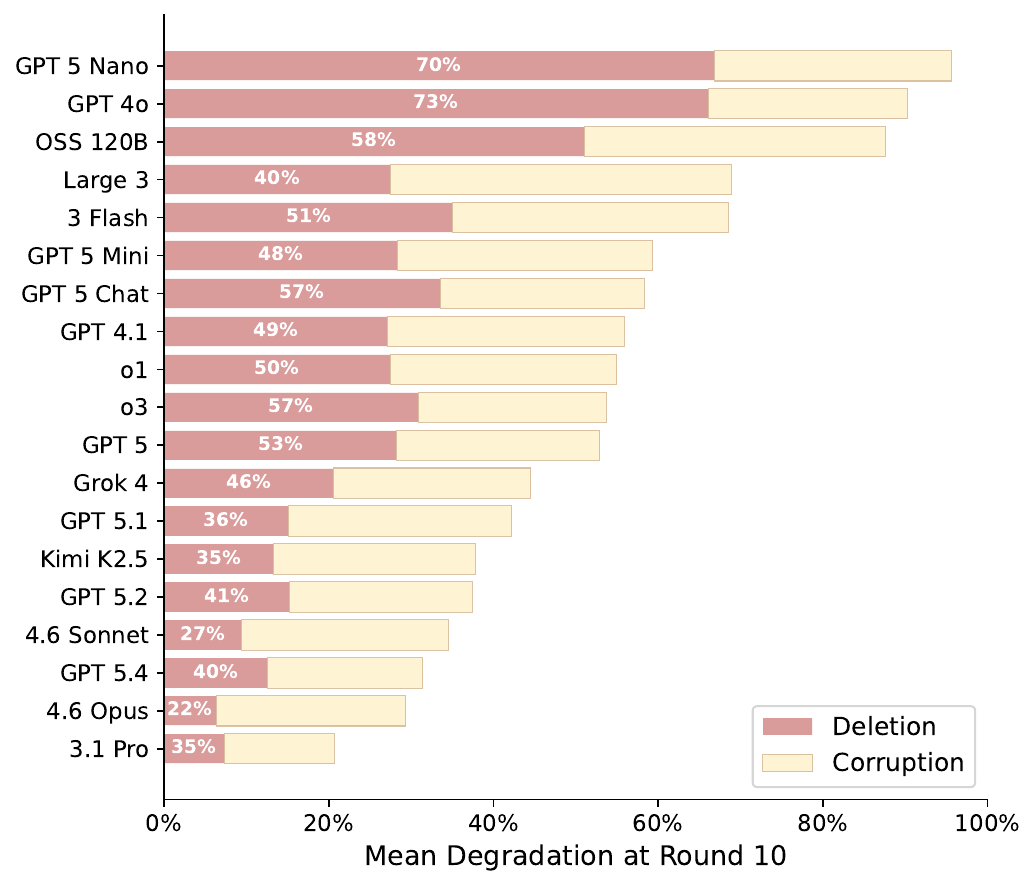}
    \vspace{-1\intextsep}
    \caption{Decomposition of degradation into deletion (missing elements) and corruption (present but incorrect).}
    \label{fig:del_cor_main}
    \vspace{-2\intextsep}
\end{wrapfigure}

\paragraph{Deletion vs.\ Corruption (Appendix~\ref{app:deletion_corruption}).} So far, the paper primarily discusses overall degradation that occurs during simulation. Yet, degradation can be caused by several underlying phenomena. To explore this further, we decompose model degradation into two components: deletion of content vs. corruption of existing content. For this analysis, we leverage the Domain Statistics component of domains in the benchmark (see Figure~\ref{fig:domain_anatomy}). For each domain, we count the structured elements (e.g., ingredients, steps) before and after a round trip: any reduction in count is attributed to deletion, and the remaining degradation is attributed to corruption. Analysis results are summarized for each model in Figure~\ref{fig:del_cor_main}. We find that \textbf{weaker models' degradation originates primarily from content deletion, while frontier models' degradation is attributable to corruption of content.}

\begin{wrapfigure}{r}{0.4\textwidth}
    \centering
    \vspace{-1.5\intextsep}
    \includegraphics[width=0.4\textwidth]{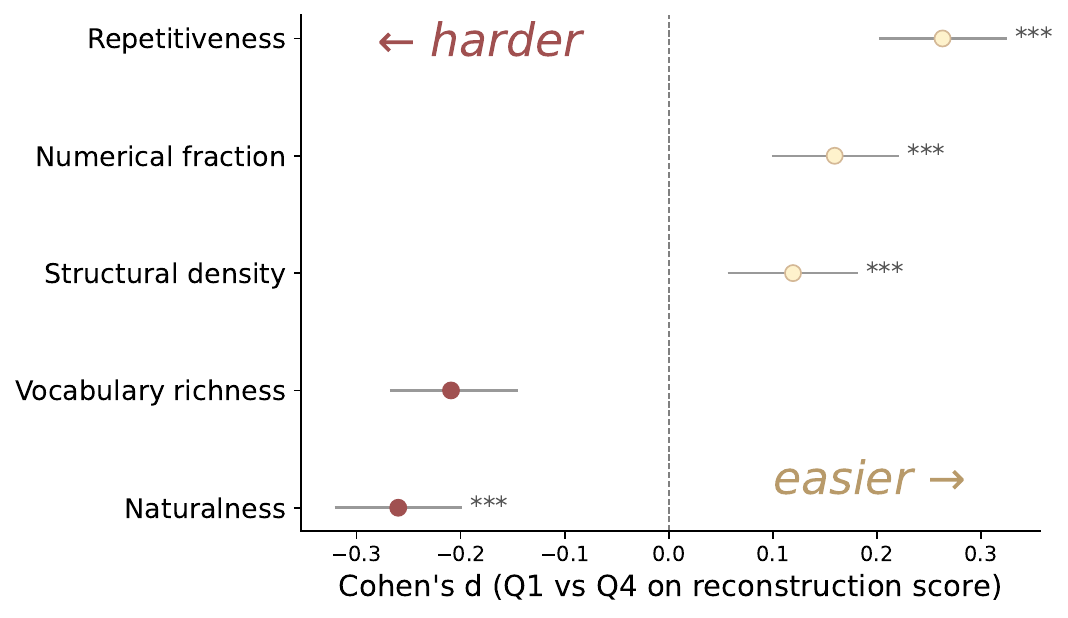}
    \caption{Cohen's $d$ effect sizes for document characteristics on scores.}
    \label{fig:domain_characteristics_main}
    \vspace{-1\intextsep}
\end{wrapfigure}

\hspace{2\intextsep}
\paragraph{Document Characteristics (Appendix~\ref{app:domain_chars}).} We analyzed how various document characteristics affect model performance, finding that \textbf{models perform better in programmatic domains (Python, DBSchema) compared to natural language domains (e.g., Recipe, Fiction)}. Performance is also higher in domains with high repetitiveness and structural density (e.g., Molecule, Chess), and lower in domains with rich unrepeated vocabulary (e.g., Transit, Textile). This echoes prior findings that LLM performance is highest in programmatic or structured domains where verifiable rewards can be defined \citep{Suma2025DeepSeekR1IR}. Through this lens, our work can be interpreted as a process to create verifiable rewards for a wide variety of domains, by building domain-specific parsing and evaluation.

\begin{wrapfigure}{l}{0.45\textwidth}
    \centering
    \vspace{-1\intextsep}
    \includegraphics[width=0.45\textwidth]{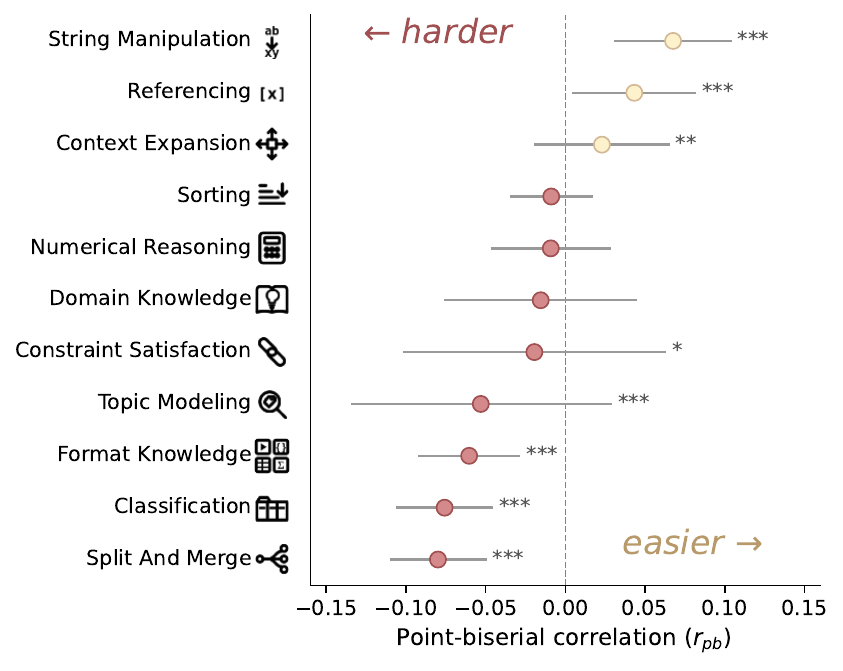}
    \vspace{-1.5\intextsep}
    \caption{Operation difficulty: point-biserial correlation with reconstruction score (GPT~5.2).}
    \label{fig:semantic_ops_main}
    \vspace{-5\intextsep}
\end{wrapfigure}

\paragraph{Semantic Operations (Appendix~\ref{app:semantic_ops}).} Each editing task in \benchmark{} was tagged with semantic operation tags that represent actions a model must take to successfully complete the task (such as sorting, merging, or string manipulation). The 11 semantic operations are listed in Figure~\ref{fig:semantic_ops_main} as well as a point-biserial correlation analysis between the presence of a tag and the reconstruction score of GPT~5.2 during a single round trip. We find editing tasks that require global document restructuring (e.g., split and merge, classification) are significantly harder than tasks involving local operations (e.g., string manipulation, referencing). In Appendix~\ref{app:semantic_ops}, we also show that tasks that require coordinating multiple operations are more challenging than tasks that involve only one.

% Focusing on the dynamics of degradation, we observe short-term divergence narrows to a tight band of scores by end of simulation, ranging in the 10-30\% content preservation. The initial advantage of larger models (such as Gemini 3 Pro at 63.2) is largely erased over long interaction, as it gets overtaken by Gemini 3.1 Flash after 20 interactions.

% We note that though \benchmark{} comprises of tasks on textual documents, some of these documents represent data from other modalities. For example, the Vector (.svg) and GraphViz (.dot), and Weaving (.wif) domains represent visual data, and Music sheet (.ly) and Audiosyn (.csd) represent audio data. Therefore, we believe that \benchmark{} still evaluates capabilities of LLMs for non-textual knowledge work tasks, due to the breadth of structured textual formats that are used in various knowledge work professions. % this can probably be skipped, it's kind of meta

\vspace{3\intextsep}
\section{Implications} \label{sec:implications}

% Implications for Model Buildings: Can we use this for training? --> Each domain is a mini-gym, with  Cycle GAN / discuss reward hacking.
\paragraph{Implications for LLM Developers.} In this work, we use \benchmark{} primarily as an evaluation tool to understand the capabilities of current LLMs. The work environments we developed could be repurposed to train models, with literature on cycle consistency training \citep{Zhu2017UnpairedIT} providing a potential training framework. Each of the 52 domains can be considered a ``mini-gym'' for online reinforcement learning, a simulation environment where an agent (LLM) can be trained to complete task cycles losslessly. Careful reward design is required to avoid agents learning misaligned behavior (i.e., reward hacking~\citep{Skalse2022DefiningAC}), such as learning to perform the no-op operation (i.e., not editing the document), or concatenating copies of original input to facilitate reconstruction. In short, combining rewards that capture both instruction-following and content preservation jointly could be a promising direction to leverage \benchmark{} to train models in diverse domains where reference solutions are not available.

% Implications for NLP Practitioners: (1) work on long interaction benchmarks; (2) work on domains beyond math & code (3) with & without tool comparison is important
\paragraph{Implications for NLP Practitioners.} Our simulation experiments provide several underexplored research directions that warrant more attention from the community, which we summarize succinctly. First, model performance in short interaction is not always predictive of long-horizon performance, and studying model capabilities for long interaction (beyond memory management) is essential to understanding readiness for realistic delegated workflows: we need more long-horizon benchmarks. Second, efforts to understand model capabilities have been unevenly distributed across domains, disproportionately studying math and code capabilities. Yet, a large proportion of knowledge work occurs in other domains: we need wider benchmarks to close this gap, studying capabilities across diverse professions and domains. Third, the community at times frames ``agent benchmarks'' and ``LLM benchmarks'' as separate fields, but they should be seen as two modes of operations to accomplish tasks: when benchmarking an LLM, we need to consider various modes of operations of the LLM to better understand its capabilities and limitations.

% Implications for Users of LLMs --  jagged frontier. Vibe Coding, but not Vibe work Yet. Trend of Progress
\paragraph{Implications for Users of AI Systems.} When delegating work to AI systems, users of LLMs should be cautious not to generalize the capabilities of the LLM in one domain to other domains. Model capabilities follow a \textit{jagged frontier} \citep{DellAcqua2023NavigatingTJ}, with models exhibiting strong (and sometimes surprising) performance at certain tasks, while making severe errors in others. Current LLMs are ready for delegated workflows in some domains such as Python coding, but not in other less common domains. In general, users still need to closely monitor LLM systems as they operate and complete tasks on their behalf. Our experiments indicate an encouraging trend, for example looking at the GPT family: 16 months separate the GPT~4o and GPT~5.4 models we tested\footnote{Respectively: November, 2024 and March, 2026}, but benchmark performance increased from 14.7\% to 71.5\%, indicative of rapid progress.

\section{Related Work} \label{sec:rel_work}

Our work sits at the intersection of four research areas.

\paragraph{Evaluating AI Systems for Knowledge Work.}

AI systems are increasingly adopted in knowledge work professions, with \citet{Bick2024TheRA} reporting that around 40\% of working-age Americans used generative AI at work in late 2024, and surveys finding knowledge workers actively integrating LLMs into their workflows \citep{Brachman2024HowKW,Ulloa2025ProductMP}. Yet, existing evaluation benchmarks have been shown to be misaligned with real-world use \citep{Wang2026HowWD}.

The community has been hard at work building benchmarks that better capture real-world work, building industry-specific benchmarks for customer service \citep{Huang2024CRMArenaUT,Yao2024taubenchAB}, enterprise knowledge work \citep{Drouin2024WorkArenaHC,Xu2024TheAgentCompanyBL}, IT operations \citep{Jha2025ITBenchEA}, or spanning multiple professions \citep{Chen2025xbenchTA,Patwardhan2025GDPvalEA,Mazeika2025RemoteLI}. However, such benchmarks require costly expert annotation, often limiting the scope of the benchmark.

Another vein of work has analyzed logs interactions, for example from users of OpenAI's ChatGPT \citep{Chatterji2025HowPU}, Anthropic's Claude \citep{Handa2025WhichET}, or Microsoft's Bing Copilot \citep{Tomlinson2025WorkingWA}. Researchers then can connect interactions with work task taxonomies such as O*NET \citep{Peterson2001UNDERSTANDINGWU}, gaining perspective on current work practices. This research however requires careful handling of privacy-sensitive data, and is limited to the few organizations that have access to interaction logs at scale.

\paragraph{Benchmarking AI Systems for Document Editing.}

Document editing is among the most common tasks in knowledge work \citep{Siu2025AugmentingEC}, and one of the primary use cases of LLM-based systems \citep{Handa2025WhichET,Eloundou2023GPTsAG}. This has spurred active research communities that study AI system capabilities to edit documents.

An established community has built methodologies to study code editing, creating evaluation benchmarks such as CanItEdit \citep{Cassano2023CanIE}, SWE-bench \citep{Jimenez2023SWEbenchCL}, CodeEditorBench \citep{Guo2024CodeEditorBenchEC}, SWE-Refactor \citep{Xu2026SWERefactorAR}.

For non-programmatic domains, where evaluation cannot rely on verifiable execution, more targeted benchmarks have been proposed, for instance to evaluate capabilities for news article editing \citep{Spangher2022NewsEditsAN}, text simplification \citep{Laban2023SWiPEAD}, fiction creative writing \citep{Chakrabarty2024CanAW}, or instruction-following \citep{Raheja2023CoEdITTE,Dwivedi-Yu2022EditEvalAI}. For structured textual domains, prior work has looked at editing graphics files (SVGEditBench \citep{Nishina2024SVGEditBenchAB}, SVGenius \citep{Chen2025SVGeniusBL}), charts and tables (ChartEditBench \citep{Kapadnis2026ChartEditBenchEG}, WikiTableEdit \citep{Li2024WikiTableEditAB}, ChartE3 \citep{Li2026ChartE3AC}), slide decks (PPTArena \citep{Ofengenden2025PPTArenaAB}, DECKBench \citep{Jang2026DECKBenchBM}), or structured output generation across multiple formats \citep{Yang2025StructEvalBL}.

This prior work typically focuses on a single domain, for which custom evaluation is curated. With \benchmark{}, we take a more generalizable approach that enables us to extend our methodology to 52 domains: we develop programmatic domain-specific parsers for each domain, and leverage a backtranslation-based evaluation that circumvents the need for references.

\paragraph{Backtranslation.}

Backtranslation (a.k.a., round-trip translation) has its roots in the neural machine translation (NMT) community, with early work showing that round-trip translation on monolingual corpora could be effectively leveraged to augment data and improve translation performance \citep{Sennrich2015ImprovingNM,Lample2017UnsupervisedMT}. Beyond data augmentation, backtranslation has been used as a direct training signal through dual learning, where forward and backward models are jointly optimized via round-trip consistency \citep{He2016DualLF,Hoang2018IterativeBF}, and as a reference-free evaluation method, where round-trip fidelity serves as a proxy for translation quality \citep{Somers2005RoundtripTW,Zhuo2022RethinkingRT}.

Backtranslation has then been successfully applied in other domains, for instance to the code domain, where it was leveraged to train unsupervised code translation models across programming languages \citep{Lachaux2020UnsupervisedTO,Rozière2021LeveragingAU}, and to jointly train code generation and summarization as dual tasks \citep{Wei2019CodeGA}. More recently, back-translation has been applied to instruction following to improve LLM alignment \citep{Li2023SelfAlignmentWI,Nguyen2024BetterAW}.

Some work has looked at chaining consecutive backtranslation cycles as a way to evaluate consistency or robustness of LLMs, measuring whether models preserve information through sequences of reversible transformations \citep{Hong2025ConsistencyCheckerTE,Min2023BeyondAE,Allamanis2024UnsupervisedEO,Maveli2026CanLC}.

We extend backtranslation-as-evaluation \citep{Zhuo2022RethinkingRT,Allamanis2024UnsupervisedEO} from single round-trips in individual domains to chained sequences across 52 diverse professions, simulating long delegated workflows where errors compound. This reduces evaluation to measuring semantic equivalence with the original document, allowing us to scale evaluation across domains without requiring annotation.

\paragraph{Evaluating Long, Multi-Session Interaction.}

AI systems are most commonly evaluated on independent conversations (single sessions) without prior history or context. \citet{Xu2021BeyondGM} introduced the first multi-session conversation dataset, showing that models trained on single sessions fail to maintain coherent long-term dialogue, and \citet{Jang2023ConversationCT} scaled this to 1M dialogues with diverse temporal dynamics.

Since then, the community has built benchmarks to evaluate memory in LLMs across sessions. \citet{Maharana2024EvaluatingVL} evaluated very long-term conversational memory, and \citet{Wu2024LongMemEvalBC} proposed LongMemEval, benchmarking core memory abilities (retention, retrieval, synthesis) in chat assistants, extended by more recent benchmarks such as EverMemBench \citep{Hu2026EverMemBenchBL} and LifeBench \citep{Cheng2026LifeBenchAB} testing memory across hundreds of interactions and diverse information sources.

Beyond memory, other work has studied LLM personalization across sessions: \citet{Jiang2025KnowMR} benchmarked dynamic user profiling across 60+ sessions, \citet{Li2025TowardMP} studied implicit preference reasoning, and \citet{Mehri2026MultiSessionCollabLU} evaluated how agents learn collaborative preferences over time. Recent work has also extended multi-session evaluation to agentic systems: \citet{Zheng2025LifelongAgentBenchEL} benchmarked lifelong learning in LLM agents, \citet{He2026MemoryArenaBA} tested memory in interdependent multi-session tasks, and \citet{Du2025MemGuideIM} introduced the first multi-session task-oriented dialogue benchmark.

Prior work frames multi-session interaction as fundamentally a \textit{memory} problem: can the system remember, retrieve, or adapt based on past interactions? With \benchmark{}, we study an orthogonal and understudied failure mode: whether repeated LLM interaction \textit{degrades the artifacts being worked on}. We study how model errors in early sessions compound and affect long-horizon performance.

\section{Limitations} \label{sec:limitations}

\vspace{-0.5\intextsep}
\paragraph{Single-Turn Interaction.} Our simulations use single-turn sessions where each instruction fully specifies a task without needing clarification. In practice, users underspecify instructions and iteratively refine intent through multi-turn conversation \citep{Herlihy2024OnOM,kim2026discoverllm}, and LLM performance degrades significantly in multi-turn settings \citep{Laban2025LLMsGL}. Extending \benchmark{} to multi-turn, multi-session simulations (e.g., via instruction sharding or user simulation \citep{naous2025flipping}) would likely amplify degradation.

\vspace{-0.5\intextsep}
\paragraph{Practical Constraints.} Our simulation parameters -- document size (3--5k tokens), distractor context (8--12k tokens), relay length (20 interactions) -- were chosen based on practical cost and context-window limits, and underestimate real-world scale. Experiments show increasing these parameters worsens degradation.

\vspace{-0.5\intextsep}
\paragraph{Conceptual Constraints.} Our framework relies on (1) backtranslation and (2) domain-specific parsing for reference-free evaluation, which constrains scope in three ways: tasks are limited to document editing (excluding other knowledge work like communication or planning); edits must be reversible (see Appendix~\ref{app:bt_limitations}); and evaluation favors structured domains where parsing is tractable. We explore expansion of the framework to more open-ended generation tasks by including the Fiction domain as one of the domains in the benchmark, though it requires adapting the evaluation to leverage a specialized evaluation method \citep{Chakrabarty2025AISlopTA} catered to creative writing.

\section{Conclusion} \label{sec:conclusion}

%In this work, we conduct a large-scale simulation of how users might delegate work to LLMs across 52 professional domains. We find that current LLMs are \textit{lossy delegates}: even frontier models lose an average of 25\% of document content after 20 interactions, with Python coding as the only domain where most models achieve near-lossless delegation. Ablation experiments show that degradation compounds with document size, interaction length, and distractor context, and that agentic tool use does not mitigate it. We release \benchmark{} as a public tool to monitor the readiness of AI systems for delegated work across the long tail of knowledge work professions.

In this work, we conduct a large-scale simulation of how users might delegate work to LLMs across 52 professional domains. We find that current LLMs are \textit{unreliable delegates}: even frontier models corrupt an average of 25\% of document content over long workflows, with sparse but severe errors that silently compound over time. Our analysis shows that degradation worsens with document length, interaction horizon, and distractor context, and is not mitigated by agentic tool use. These results highlight a fundamental gap in reliability that undermines trust in delegation. We release \benchmark{} as a public tool to monitor the readiness of AI systems for delegated work in knowledge work professions.

\section*{Acknowledgements}

We thank Hiroaki Hayashi, Yoonjoo Lee, Tarek Naous, Jihoon Tack, Michel Galley, Tanya Goyal, and Kiran Tomlinson for great feedback along the way.

\section*{Ethics Statement} \label{sec:ethics}

\paragraph{AI Use Disclosure.} AI was used in multiple stages of this project. First, AI assisted the authors with developing the codebase and curating benchmark work environments. Second, AI was used in some annex evaluations in LLM-as-a-judge setups (see Appendix~\ref{app:instruction_compliance}), though the evaluation protocol of our main simulation experiment relies on domain-specific parsing and not LLM-based evaluation. Third, AI was used to assist with the writing of the Appendix of the paper, helping document precisely various details of our work. AI was not used extensively for writing the main text, beyond minor typo, fluency fixes, and compression of content to adhere with submission space constraints.

\bibliographystyle{colm2026_conference}
\bibliography{colm2026_conference}

\clearpage
\appendix
\setcounter{figure}{0}
\setcounter{table}{0}
\renewcommand{\thefigure}{A\arabic{figure}}
\renewcommand{\thetable}{A\arabic{table}}
\renewcommand{\theHfigure}{A\arabic{figure}}
\renewcommand{\theHtable}{A\arabic{table}}

\section{Instruction Compliance Validation} \label{app:instruction_compliance}

A major risk of backtranslation-based evaluation is over-estimating model capability due to model non-compliance with the instruction. We assume when conducting backtranslation evaluation (a sequence of forward and backward edit steps) that the model makes best-effort attempts to follow instructions. If instead, the model completes a simpler task (such as copying the input, or an easy subset of the instruction) that lead to high reconstruction scores, the evaluation is over-estimating model capabilities.

We conduct an analysis to quantify instruction compliance of models during our simulation. The objective is to determine whether the LLMs are majoritarily making attempts to complete the editing tasks as instructed, or whether they might be taking shortcuts that invalidate the evaluation. Crucially, the analysis does not consider whether the LLM's attempt is correct, but rather whether it attempted the editing task or not. We found through small-scale analysis that measuring instruction compliance is more tractable than judging correctness (the evaluation task, see Appendix~\ref{app:alt_eval}), and we leverage an LLM judge (GPT~5.4) to classify compliance.

\subsection{Methodology}

Each editing step attempted by a model is classified by an LLM judge into one of eight compliance categories: \textit{fully executed} (attempted the full instruction, possibly with errors), \textit{partially executed} (attempted some but not all aspects of the instruction), \textit{truncated attempt} (started executing the task but output was cut short), \textit{hallucinated output} (produced unrelated content), \textit{superficial attempt} (the attempt introduces cosmetic changes only and does not address the full scope of the instruction), \textit{not executed} (returned input unchanged), \textit{empty response} (returned empty or near-empty output), and \textit{instruction infeasible} (the editing instruction is no longer feasible given corrupted input from prior rounds). The judge receives the editing instruction, the input document, and the output document, and must select exactly one category with a justification.

An important confounder of the analysis is the performance of the model on the instruction: we want to study whether the model is attempting the instruction both when it achieved high and low reconstruction scores. For this reason, we drew a stratified sample of round trips from our main experiment (Section~\ref{sec:experiments}), sampling across all models and five reconstruction score bins: \textit{collapsed} $[0, 20)$, \textit{low} $[20, 50)$, \textit{mid} $[50, 80)$, \textit{high} $[80, 98)$, and \textit{perfect} $[98, 100]$. For each sampled round trip, both the forward and backward editing steps are evaluated independently, yielding 12{,}409 individual step-level judgments across 52 domains.

\subsection{Findings}

\begin{wrapfigure}{r}{0.5\textwidth}
    \vspace{-16pt}
    \centering
    \includegraphics[width=0.48\textwidth]{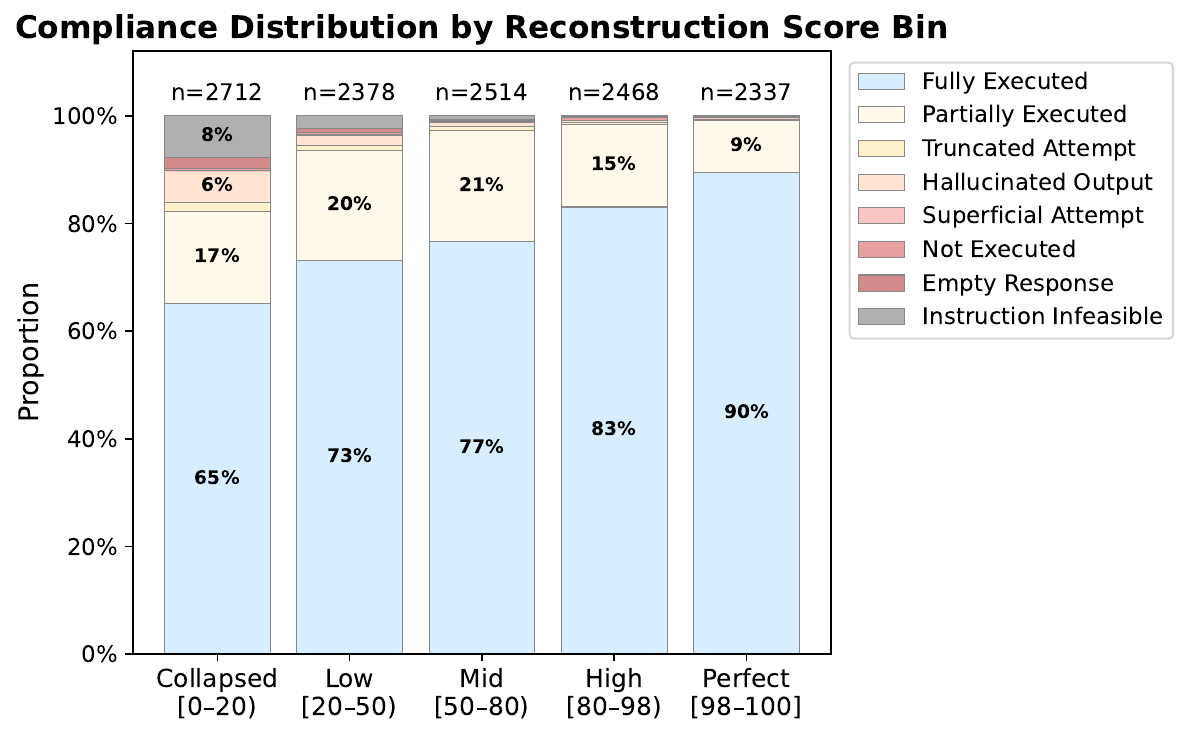}
    \caption{Instruction compliance distribution by reconstruction score bin. Models overwhelmingly attempt to complete instructions in our simulation, validating that round-trip reconstruction aligns with measuring model capabilities at completing the editing tasks.}
    \label{fig:compliance_by_score_bin}
    \vspace{-12pt}
\end{wrapfigure}

Figure~\ref{fig:compliance_by_score_bin} presents the compliance distribution across score bins. Three key findings emerge.

\paragraph{Models overwhelmingly attempt instructions.} Across all score bins, 93.8\% are fully or partially executed. Non-compliance categories (not executed, empty response, hallucinated output) collectively account for only 3.0\% of steps, concentrated in the worst-performing models; among the top-10 best-performing models, non-compliance drops to 1.7\%. This confirms that instruction-following LLMs make genuine best-effort attempts at the editing tasks in \benchmark{}.

\paragraph{Low scores reflect execution errors, not non-compliance.} The critical test is whether steps that produce low reconstruction scores are dominated by non-compliant behaviors. Among steps in the collapsed score bin ($<$20), 82.3\% are classified as fully or partially executed. In other words, models attempted the instruction but made errors severe enough to prevent reconstruction. As reconstruction scores rise, compliance rates also rise.

\paragraph{Non-compliance reasons are model failures.} The lowest score bin (collapsed) shows that the two major non-compliance categories are hallucinated output (6\%) and instructions being infeasible (8\%). The former is a known limitation of LLMs, while the latter is a consequence of the setting of our simulation: errors introduced in early rounds (such as hallucinations) can render later editing tasks infeasible, in cases where the relevant document elements are not present anymore. These two categories do not indicate models taking shortcuts to achieve higher scores, but rather reflect realistic failure modes that arise during delegated workflows.

\paragraph{Partial execution may overstate model capabilities.} Across all judgments, 16.7\% of steps are classified as partially executed, meaning the model attempted some but not all aspects of the instruction. The rate of partial execution rises from 9.5\% in the perfect score bin to approximately 20\% in the mid and low bins. Since a partially executed edit is a simpler transformation than the intended one, it is easier likely easier to reverse, and the resulting round-trip score may lead us to overestimate the model's capability at the full intended task. In roughly one in six steps, models complete a simplified version of the requested edit, and the reconstruction score captures their performance on that simpler task rather than on the original instruction.

Taken together, these results validate the use of backtranslation-based evaluation in \benchmark{}: models predominantly make genuine attempts at the editing tasks, and when they deviate---through hallucination, truncation, or infeasible instructions---the reconstruction score effectively captures the resulting degradation. The non-trivial rate of partial execution further suggests that our reported scores may slightly overstate true model capabilities at the intended tasks.

\section{Backtranslation Properties, Assumptions, and Limitations} \label{app:backtranslation}

In Section~\ref{sec:delegate52}, we introduced the round-trip backtranslation primitive as the backbone of evaluation in \benchmark{}. This appendix provides a more thorough discussion of the properties, assumptions, and limitations of this evaluation scheme.

We use the following notation, consistent with the main text. A work environment contains a seed document $s$ and a set of edit tasks, each consisting of a forward instruction $x^{\rightarrow}$ and a backward instruction $x^{\leftarrow}$ defining a transformation $\sigma$ and its inverse $\sigma^{-1}$. An LLM $M$ produces a transformed document $t = M(s; x^{\rightarrow})$ and a reconstructed document $\hat{s} = M(t; x^{\leftarrow})$. A domain-specific similarity function $\mathrm{sim}(s, \hat{s}) \in [0, 1]$ measures reconstruction faithfulness.

\subsection{Properties} \label{app:bt_properties}

We first describe three properties that are inherent to backtranslation-as-evaluation and hold by construction, independent of any assumptions about the LLM being tested.

\paragraph{Reference-free evaluation.} Backtranslation reduces the problem of evaluating LLM editing capability to measuring semantic equivalence between the original document $s$ and the reconstructed document $\hat{s}$. No reference annotations of intermediate states (i.e., the forward-edited document $t$) are required. This is the key property that enables \benchmark{} to scale to 52 domains without per-task annotation, in contrast to benchmarks that require expert-curated reference outputs for each task.

\paragraph{Composability.} Each round-trip is an independent cycle that ideally returns the document to its original state. Round-trips can therefore be chained in any order to simulate arbitrarily long interactions, as formalized in Eq.~1. This composability is what enables us to simulate 20-interaction delegated workflows from a set of 5--10 edit tasks per work environment, by applying round-trips in sequence.

\paragraph{Round-trip as atomic measurement unit.} The evaluation measures the composite operation $M(x^{\leftarrow}; M(x^{\rightarrow}; s))$, i.e., the full round-trip from $s$ through $t$ and back to $\hat{s}$. In other words, the evaluation captures the \textit{net} effect of both editing steps, losing the ability to decompose whether errors were introduced in the forward step, the backward step, or both (see \S\ref{app:bt_limitations}).

\subsection{Assumptions} \label{app:bt_assumptions}

The validity of backtranslation-as-evaluation rests on several assumptions about the edits and the models being tested. We enumerate these assumptions, discuss how our benchmark design enforces or encourages them, and note where they hold empirically.

\paragraph{Stateless execution.} Each call to the LLM is independent: $M(x; s)$ depends only on the instruction $x$ and the document $s$, not on prior interactions. In our experiments, every editing step is conducted as a separate, single-turn session with no conversational history. We argue this design reflects realistic delegated work, where tasks may span different sessions, days, or stakeholders, and the full history of prior work is not always available in the environment. A consequence is that our simulations do not capture multi-turn refinement within a single session (e.g., ``actually, change the third ingredient instead''), which we discuss as a limitation below. In short, \benchmark{} is a multi-session but single-turn benchmark.

\paragraph{Non-triviality.} The model must not treat edit instructions as no-ops: $M(x^{\rightarrow}; s) \neq s$. If a model simply returns the input unchanged, it trivially achieves $\mathrm{sim}(s, \hat{s}) = 1$ without performing any work. We validated empirically in Appendix~\ref{app:instruction_compliance} that models make best-effort attempts at executing instructions in close to 94\% of analyzed cases, confirming that low reconstruction scores reflect genuine execution errors rather than non-compliance. As an additional safeguard, our edit design rules explicitly require transformative edits (not purely expansive operations), making it less likely for a model to produce a no-op that would satisfy the instruction.

\paragraph{Transformative edits.} Edits must require genuine transformation of the document, not merely expanding on the content. Formally, the forward edit $\sigma(s)$ cannot be decomposed into $[s, \sigma'(s)]$ (concatenation of the original with new content), as this would make the backward edit trivial (cropping the appended content). This assumption is enforced by our edit design rules (Rules~3, 4, 9 in Appendix~\ref{app:domain_creation}), which require that edits modify the existing content in place rather than purely extending it. We note that this does not exclude edits to have an expansive component (e.g., adding a new section to a report), as long as they also require modification of existing content (e.g., re-ordering sections).

\paragraph{Order independence.} Round-trips must be composable in any order without affecting the ideal outcome. That is, for any two edit tasks $(\sigma_i, \sigma_i^{-1})$ and $(\sigma_j, \sigma_j^{-1})$, a perfect model should achieve $\mathrm{sim}(s, \hat{s}) = 1$ regardless of whether $\sigma_i$ is applied before or after $\sigma_j$. This is a design constraint on edit creation: edit tasks within a work environment must be mutually independent, so that one edit's forward--backward cycle does not alter state that another edit depends on.

\paragraph{Error faithfulness.} For the round-trip score to be a meaningful measure of LLM capability, errors introduced during the forward step must survive through the backward step and surface in the evaluation. This requires two conditions. First, the backward step must approximately preserve errors: if the forward step introduces an error into $t$, the backward step should not coincidentally correct it. Formally, $M(x^{\leftarrow}; \cdot)$ should be approximately injective --- distinct inputs should produce distinct outputs, so errors are not collapsed. Second, error cancellation must be improbable: if $M(x^{\rightarrow}; s)$ introduces an error $\epsilon$, the probability that $M(x^{\leftarrow}; \cdot)$ introduces exactly $-\epsilon$ should be negligible. LLMs are stochastic and do not produce systematically self-canceling errors. We observe empirically that round-trip evaluation surfaces errors faithfully: in our experiments, low scores consistently correspond to genuine content loss or corruption upon manual inspection.

\subsection{Limitations} \label{app:bt_limitations}

We now discuss inherent limitations of the backtranslation approach: aspects of LLM capability or delegated work that our evaluation framework cannot capture.

\paragraph{Focus on edit-based tasks.} Knowledge work is complex and involves many types of tasks beyond document editing, including information retrieval, decision-making, communication, and planning. Our simulation focuses exclusively on document editing, because edits produce observable changes to artifacts that can be evaluated programmatically. This scope means \benchmark{} does not assess LLM capabilities in other aspects of delegated work, even though such capabilities are equally important for real-world deployment.

\paragraph{Round-trip opacity.} As noted in \S\ref{app:bt_properties}, the evaluation measures the composite round-trip and cannot observe individual editing steps. This has two consequences. First, a high score does not guarantee that the forward edit was performed well: a model could produce a poor or trivial forward edit that happens to be easy to reverse, achieving a high round-trip score without demonstrating genuine editing capability. Therefore, the round-trip score should be interpreted as measuring \textit{content preservation through editing cycles}, not the quality of any individual edit. Second, a low score cannot be decomposed: we cannot determine whether errors arose in the forward step, the backward step, or both. A further consequence is that our findings can only measure degradation every two interactions (i.e., every full round-trip), not after each individual edit.

\paragraph{Reversibility constrains the edit space.} By construction, all edit tasks in \benchmark{} must be reversible: each forward instruction has a corresponding backward instruction that, for a perfect model, returns the document to its original state. This constraint excludes inherently irreversible real-world operations such as lossy compression, content deletion, or stylistic rewriting that is not captured by the domain parsing implementation. We however note that lossy transformations can often be combined with record keeping to create an overall reversible edit. Consider the following: an edit that alters the order of sections in a document is not reversible as order information is lost. But the edit can be made reversible by instructing the model to keep track of the original order in a separate original\_order.csv book-keeping file. We use this book-keeping trick extensively in \benchmark{} to create reversible versions of edits that would otherwise be irreversible.

\paragraph{Single-turn interaction only.} Our stateless execution design means each editing step is a single-turn interaction: the model receives the document and an instruction, and produces a result. Real-world delegated work often involves multi-turn refinement within a session (e.g., a user reviewing an edit and requesting corrections). Our framework does not capture the ability of models to self-correct through iterative dialogue, nor does it test whether models can improve their edits when given feedback. Results from our agentic experiment (Section~\ref{sec:agentic}) offer partial insight, showing that even with tool-mediated iteration, models do not improve over the single-turn baseline. In Section~\ref{sec:limitations}, we discuss possible extensions to \benchmark{} that would allow for a multi-turn, multi-session simulation environment.

\paragraph{Evaluation Imperfection.} The domain-specific similarity function $\mathrm{sim}(\cdot, \cdot)$ serves as ground truth for all measurements. If a parser lacks robustness --- aspects of document semantics it fails to capture --- then errors in those dimensions go undetected, creating a ceiling on evaluation sensitivity. We mitigate this risk through the ablation testing described in Appendix~\ref{app:domain_creation} (removing $K$ of $N$ blocks must reduce the score proportionally) and through the multi-stage QA process that iteratively hardens parsers against edge cases. The post-hoc analysis in Appendix~\ref{app:alt_eval_method} further validates that our domain-specific metrics capture substantially more variance than generic alternatives. However, parsing robustness is a long-tail challenge, and it likely leads to small-scale bias in evaluation.

\paragraph{Subjective or many-to-many edits.} Edits where multiple valid outputs exist (e.g., ``make this recipe more appealing'' or ``improve the writing style'') cannot be reliably round-tripped, because the backward instruction cannot specify a unique inverse. Our edit design rules address this by requiring that edits be precisely specified: the forward and backward instructions must define a unique transformation path, leaving little room for subjective interpretation. This means \benchmark{} tests \textit{precise, task-oriented editing} rather than open-ended creative transformation, which is a distinct and complementary capability.

\section{Alternative Evaluation Methods} \label{app:alt_eval}

In Section~\ref{sec:def_round_trip}, we described the domain-specific parsing and evaluation approach used in \benchmark{}. A natural question is whether simpler, generic evaluation methods could serve as adequate substitutes. We present a post-hoc comparison of alternative evaluation methods against the domain-specific scores, assessing whether generic evaluation methods are sufficient for evaluation in the complex textual domains included in \benchmark{}.

\subsection{Evaluation Methodology} \label{app:alt_eval_method}

\paragraph{Sample Selection.} We draw a stratified sample of 9,851 backward evaluation entries from our main experiment results (Section~\ref{sec:experiments}). Entries are stratified by domain ($N{=}52$) and domain-specific score bucket (20 buckets of width 0.05 over $[0, 1]$), sampling up to 10 entries per cell with a fixed random seed. This ensures uniform coverage across both domains and score levels. Entries with evaluation errors are excluded.

\paragraph{Hard Subset.} Not all evaluation instances are equally informative: entries where the candidate is substantially shorter or longer than the reference (e.g., empty or truncated responses) are trivially scored by any method. To isolate genuinely challenging cases, we define a \textit{hard} subset consisting of entries where the character-level length ratio between the candidate and reference falls within $[0.8, 1.2]$. This subset ($N{\approx}4{,}600$, roughly 47\% of entries) filters out cases where length alone provides a strong scoring signal.

\paragraph{Alternative Methods.} We evaluate five classes of generic evaluation methods against the domain-specific scores:

\begin{enumerate}
    \item \textbf{Levenshtein Ratio}~\cite{Levenshtein1965BinaryCC}: Character-level edit distance normalized by document length, computed via Python's \texttt{SequenceMatcher}. Measures surface-level textual similarity.
    \item \textbf{ROUGE-L}: Longest common subsequence (LCS)-based F-measure~\cite{Lin2004ROUGEAP}, a standard metric in text generation evaluation. Captures token-level sequential overlap. 
    \item \textbf{Embedding-based Similarity}: Cosine similarity between document-level embeddings. We evaluate four embedding models: nomic-embed-text-v1.5~\cite{Nussbaum2024NomicET} (run locally), and OpenAI's text-embedding-small and text-embedding-large~\cite{Neelakantan2022TextAC}. All embedding models have capacity for at least 8,000 tokens, which is sufficient for the majority of documents (2--5k tokens). 
    \item \textbf{BERTScore}~\cite{Zhang2019BERTScoreET}: Token-level contextual embedding matching using RoBERTa-large~\cite{Liu2019RoBERTaAR}, with greedy alignment between candidate and reference tokens. Note that BERTScore's 512-token context window truncates most of our 2--5k token documents, evaluating only document prefixes.
    \item \textbf{LLM-as-Judge}~\cite{Zheng2023JudgingLW}: LLM prompting approach (GPT~5.4, GPT~5 Nano) to rate semantic equivalence on a 0--100 scale, accounting for domain-specific semantics.
\end{enumerate}

\noindent For each method, we compute Spearman's $\rho$ against the domain-specific score on both the full sample and the hard subset.

\subsection{Evaluation Findings} \label{app:alt_eval_findings}

\begin{figure*}[t]
    \centering
    \includegraphics[width=\textwidth]{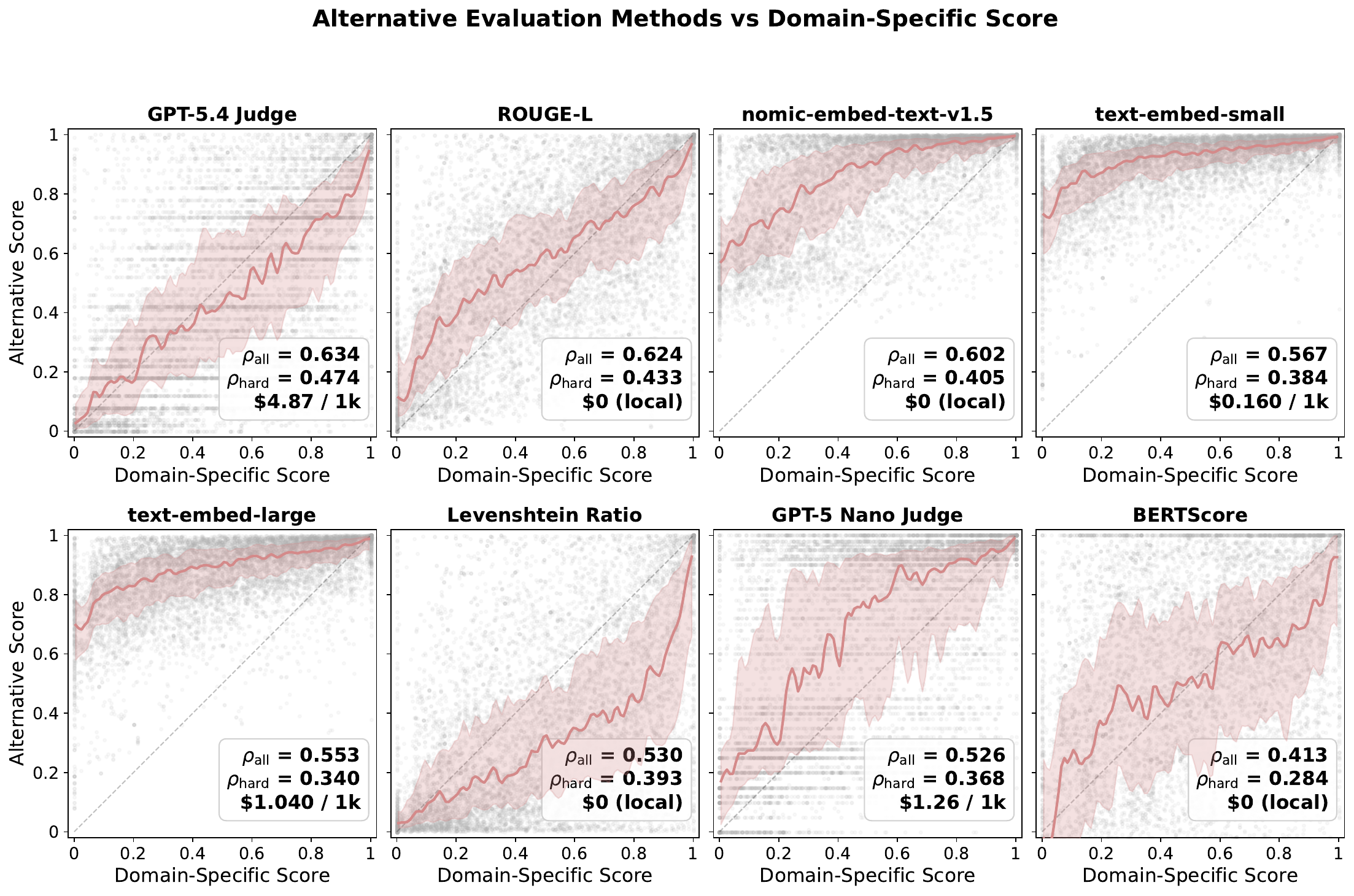}
    \caption{Comparison of alternative evaluation methods against domain-specific parsing-based scores. Each subplot shows the scatter of individual entries (gray) with a smoothed calibration curve (rose, median with IQR band). Methods are ranked by Spearman's $\rho$ on the full sample. The dashed diagonal represents perfect calibration.}
    \label{fig:alt_evals}
\end{figure*}

Figure~\ref{fig:alt_evals} presents the comparison of alternative evaluation methods against our domain-specific scores. Three key findings emerge:

\paragraph{(1) No method achieves strong correlation, especially on hard examples.} The best-performing method, GPT~5.4 as judge, achieves $\rho_{\mathrm{all}} = 0.634$ on the full sample but only $\rho_{\mathrm{hard}} = 0.474$ on the hard subset, explaining less than 25\% of variance in domain-specific scores ($\rho^2 \approx 0.22$). Surface-level methods (Levenshtein, ROUGE-L, local embedding) converge to $\rho_{\mathrm{hard}} \approx 0.40$, suggesting a ceiling for generic methods on this task. BERTScore performs worst ($\rho_{\mathrm{hard}} = 0.284$), likely due to its 512-token context window truncating most of our 2--5k token documents.

\paragraph{(2) Systematic calibration biases.} The calibration curves in Figure~\ref{fig:alt_evals} reveal characteristic biases. Surface-level methods (Levenshtein, ROUGE-L) tend to over-penalize: documents with high domain-specific scores can receive low surface-level similarity scores when the model makes surface-level changes that do not affect semantics. Conversely, embedding-based methods and LLM judges tend to over-estimate: documents with low domain-specific scores receive high similarity scores because they appear superficially similar despite containing semantically consequential errors. Both bias direction would impose limits on reliable evaluation.

\paragraph{(3) The best method is cost-prohibitive.} GPT~5.4 as judge is the only method with meaningfully higher hard-subset correlation than simple baselines, but at a cost that is impractical for large-scale evaluation. At an estimated cost of about \$4 per 1{,}000 judgments, the total cost would be on the order of the experiment itself, effectively doubling the budget without yielding reliable scores.

These results demonstrate the necessity of domain-specific evaluation in \benchmark{}. Generic evaluation methods fail to capture the fine-grained semantic distinctions that determine whether a document has been correctly preserved across domains. Though the domain-specific parsing approach requires upfront implementation effort, it is both free to run at scale and sensitive to the domain-specific semantics that matter for evaluating delegated work.

\section{Round-Robin Design Validation} \label{app:round_robin}

\begin{wraptable}{r}{0.5\textwidth}
\vspace{-4\intextsep}
\input{tables/round_robin}
\caption{Round-robin (\iconroundrobin) vs.\ single-edit (\iconsingleedit) degradation for four models. Single-edit repeats one task for all 10 round trips; round-robin cycles through 5--10 tasks. Task diversity drives the majority of observed degradation.}
\label{tab:round_robin}
\vspace{-1\intextsep}
\end{wraptable}

The main experiment uses round-robin task scheduling: edits cycle through all available tasks, shuffling order at each epoch. We validate this design choice by comparing against a \textit{single-edit} ablation in which a relay consists of a single forward-backward edit instruction repeated in all rounds.

\paragraph{Methodology.} For each work environment, the single-edit condition runs a separate 10-round-trip relay for every available edit task (5--10 per environment), repeating that single forward-backward pair in all rounds. Since results are averaged across all edit tasks within each environment, both conditions converge to the same set of attempted edits and the difference lies only in whether tasks are interleaved (round-robin) or isolated (single-edit). We tested four models (GPT~5.4, GPT~5.2, GPT~5.1, GPT~4.1) across 50 domains. Running the single-edit setting requires running a full simulation for each editing task rather than for each work environment, which would require 5-10x more compute. We therefore restrict our experiments to one work environment per domain, ensuring we still get a valid estimate of performance while keeping computational costs manageable.

\paragraph{Findings.} Table~\ref{tab:round_robin} summarizes the results. At RS@2 (a single round trip), the two conditions are nearly identical (within a few points due to sampling noise). This is expected, as in the first round-trip, all editing tasks are novel. However, degradation rates quickly diverge, and by RS@20, single-edit scores are 20--24 points higher than round-robin across all four models. For instance, GPT~5.4 retains 88.5\% under single-edit but only 66.9\% under round-robin. Degradation curves under single-edit are remarkably flat: GPT~5.4 loses only 6 points over 10 round trips (94.3$\to$88.5), compared to 28 points under round-robin.

This result indicates that \textbf{task diversity and not repetition are the primary driver of degradation in our simulations.} Each new edit type introduces distinct error modes that compound, whereas repeating the same edit allows the model to settle into a stable (though still imperfect) cycle.

In summary, we validate the importance of designing multiple editing tasks per work environment and using a round-robin schedule during simulation. Not only does this design better reflect realistic workflows, but it also leads to more observed degradation, revealing a more accurate picture of current LLM capabilities. Since task repetition substantially reduces degradation, we conjecture that if all work environments were constructed with 10 or more unique editing tasks (eliminating repetition entirely), reported degradation levels would be even higher than those in our main experiment.

\section{Critical Error Analysis} \label{app:critical_errors}

\begin{table*}[h]
    \centering
    \resizebox{0.8\textwidth}{!}{\input{tables/critical_errors}}
    \caption{Critical error analysis (10+ drop within single round). \textit{Left:} cumulative \% of runs with at least one critical error after N interactions. \textit{Right:} share of total degradation from critical errors and average per-run drop magnitudes. Models sorted by round-20 critical error rate (ascending).}
    \label{tab:critical_errors}
\end{table*}

We define a \textit{critical error} as a round trip that leads to a degradation of at least 10 points relative to the previous round. Table~\ref{tab:critical_errors} analyzes the prevalence of critical errors in the simulations we conducted.

The first section of the Table (\textit{\% Runs w/ Critical Error by Round}) reports, for each interaction length, the cumulative percentage of runs in which at least one critical error has occurred by that point. The second section (\textit{Degradation Breakdown}) quantifies the share of total degradation attributable to critical errors: \textit{\% Critical} is the pooled ratio of critical-error drops to all drops across runs, while \textit{Avg Drop} and \textit{Avg Crit Drop} report the mean per-run total and critical gross drop (in percentage points).

Across all models, critical errors account for 80--98\% of total degradation, confirming that score loss is dominated by critical single-step failures rather than gradual accumulation of small errors. By round 20, the majority of runs for all models except Gemini 3.1 Pro have experienced at least one such critical error.

In other words, our simulations indicate models are not failing due to ``death by a thousand cuts.'' LLMs don't slowly corrupt content through many small rounding errors. Instead, they maintain near-perfect reconstruction in some rounds, and experience critical failures in a few rounds — typically losing 10-30+ points in a single round-trip. These sparse critical failures explain document degradation in large part. The stronger models (Gemini 3.1 Pro, Claude 4.6, GPT~5.4) aren't avoiding small errors better, they delay critical failures to later rounds and experience them in fewer interactions.

\section{Deletion vs.\ Corruption Decomposition} \label{app:deletion_corruption}

When models degrade document content, there are two distinct error categories: \textit{deletion} (dropping content that was present in the original document) and \textit{corruption} (modifying, hallucinating, or otherwise distorting content). Understanding the relative contribution of each is important for designing mitigations: deletion might be more readily noticed as document size shrinks, while corruption requires more careful review to detect.

\subsection{Methodology}

\begin{wrapfigure}{r}{0.45\textwidth}
    \vspace{-1\intextsep}
    \centering
    \includegraphics[width=0.45\textwidth]{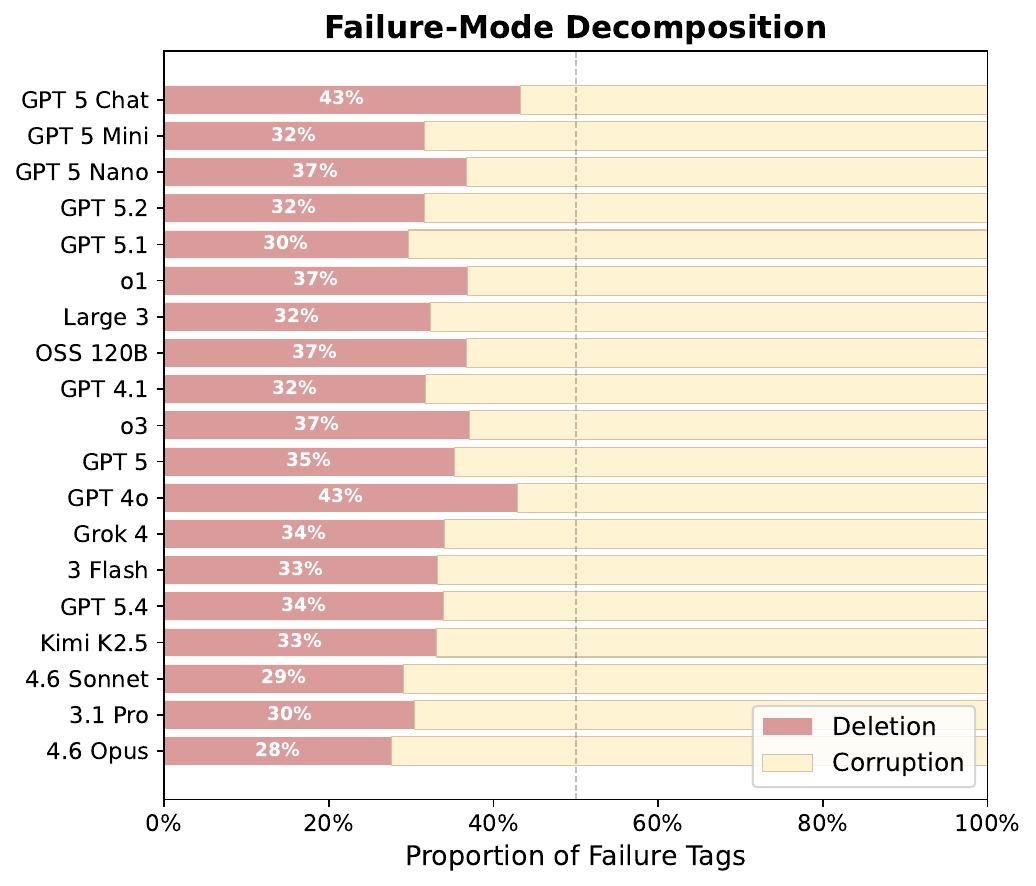}
    \caption{Failure-mode re-aggregation. Each tagged failure instance is classified as deletion or corruption. Deletion share ranges from 28\% (Claude 4.6 Opus) to 43\% (GPT~5 Chat, GPT~4o).}
    \label{fig:del_cor_fm}
    \vspace{-1\intextsep}
\end{wrapfigure}

\paragraph{Count-based decomposition.} Our domain-specific evaluators (Figure~\ref{fig:domain_anatomy}) parse documents into structural elements (e.g., ingredients in a recipe, entries in a ledger) and report both reference and generated element counts. Let $n_{\mathrm{ref}}$ and $n_{\mathrm{gen}}$ be the reference and generated element counts, and $s \in [0, 1]$ the reconstruction score. We define \textit{coverage} $= \min(n_{\mathrm{gen}} / n_{\mathrm{ref}},\, 1)$, which captures the fraction of expected elements that are present. The deletion component is $1 - \text{coverage}$ (elements that are missing), and the corruption component is $\text{coverage} - s$ (elements that are present but incorrect). The two components sum to $1 - s$, the total degradation. This analysis is conducted on a subset of 38 of the 52 domains for which element counts meaningfully represents the core element in the domain.

\paragraph{Failure-mode re-aggregation.} In a separate analysis, we tagged failure instances from our simulation with one of 11 failure-mode labels. We group these labels into two buckets: \textit{deletion} (\texttt{content\_loss} and \texttt{truncation}) and \textit{corruption} (the remaining 9 tags: \texttt{hallucination}, \texttt{structure\_change}, \texttt{skipped\_backward\_edit}, \texttt{syntax\_error}, \texttt{mathematical\_error}, \texttt{duplicated\_content}, \texttt{reordering}, \texttt{templated\_completion}, and \texttt{other}).

\subsection{Findings}

\paragraph{Weaker models delete more, corruption dominates for frontier models.} In the count-based analysis (Figure~\ref{fig:del_cor_main}, main text), for the models with worse performance (GPT~4o, GPT~5 Nano) 70--73\% of degradation is attributable to deletion, while for current frontier models (Claude 4.6 Opus, Claude 4.6 Sonnet) deletion only explains 22--27\% of observed degradation. The failure-mode analysis (Figure~\ref{fig:del_cor_fm}) corroborates this: deletion accounts for 35\% of all tagged failures, with a narrower range (28--43\%) across models.

In short, current LLMs primarily corrupt user documents in delegated workflows. Degradations observed over repeated editing interaction is primarily attributable to the model altering content in a way that is incorrect, hallucinated or distorted, rather than simply deleting content.

\section{Document Characteristics Analysis} \label{app:domain_chars}

To understand what document characteristics influence task difficulty for current LLMs, we analyzed the relationship between measurable document characteristics and reconstruction scores for GPT~5.2. As with the semantic operation analysis (Appendix~\ref{app:semantic_ops}), we use single-round-trip results from the edit testing phase (Appendix~\ref{app:edit_qa}) to isolate document-level effects from multi-round error accumulation.

\subsection{Document Characteristic Metrics}

\begin{wraptable}{r}{0.45\textwidth}
    \vspace{-1\intextsep}
    \centering
    \input{tables/domain_characteristics}
    \caption{Category-level document characteristics and mean reconstruction scores (GPT~5.2, single round trips). LLMs perform best in Science \& Engineering domains (97.3\%) and worst in Everyday domains (89.9\%).}
    \label{tab:domain_characteristics}
    \vspace{-1\intextsep}
\end{wraptable}

We compute five properties from the initial document of each work environment, capturing different aspects of document structure and content:

\paragraph{Naturalness.} Ratio of function words (determiners, prepositions, pronouns, conjunctions, and auxiliary verbs) to total words, normalized so that the typical prose rate (${\sim}$45\% function words) maps to 1.0. High values indicate natural language prose; low values indicate code, data, or markup.

\paragraph{Numerical fraction.} Fraction of whitespace-delimited tokens that contain at least one digit. Captures the prevalence of numeric data (coordinates, timestamps, quantities) in the document.

\paragraph{Vocabulary richness.} Type-token ratio: number of unique lowercased words divided by total words. Higher values indicate diverse vocabulary (e.g., prose), while lower values indicate repetitive token usage (e.g., structured records with recurring field names).

\paragraph{Repetitiveness.} Fraction of 5-grams that appear more than once in the document. High values indicate documents with repeated structural patterns (e.g., tabular rows, chemical records); low values indicate documents with mostly unique phrasing.

\paragraph{Structural density.} Fraction of characters that are neither alphabetic nor whitespace (e.g., punctuation, brackets, operators, delimiters). High values indicate markup-heavy or code-heavy content; low values indicate natural prose.

\subsection{Category-Level Overview}

Table~\ref{tab:domain_characteristics} reports mean reconstruction scores and document characteristics for each of the five domain categories defined in Section~\ref{sec:delegate52}, and Figure~\ref{fig:domain_characteristics_main} (main text) shows the Cohen's $d$ effect sizes for each property on reconstruction score.

\subsection{Findings}

Two document characteristics show the largest effects. \textit{Repetitiveness} ($d = {+}0.261$, $p < 0.001$) is the strongest positive predictor: LLMs degrade less on documents with highly repetitive structure (e.g., tabular data, chemical records). \textit{Naturalness} ($d = {-}0.260$, $p < 0.001$) is the strongest negative predictor: LLMs degrade more on documents dominated by natural language prose.

\textit{Numerical fraction} ($d = {+}0.159$, $p < 0.001$) and \textit{structural density} ($d = {+}0.119$, $p < 0.001$) are also associated with less degradation. \textit{Vocabulary richness} ($d = {-}0.209$) has a notable effect size but does not reach statistical significance ($p = 0.18$), possibly due to its correlation with naturalness.

In summary, LLMs degrade least on documents that are repetitive, numerical, and structurally dense---properties typical of formal and machine-oriented formats---and most on documents that are natural and lexically diverse---properties typical of human-authored prose. This provides actionable advice for knowledge work delegation: current LLMs are more performant at manipulating structured files (Science \& Engineering, Code \& Configuration) than natural language documents (Everyday, Creative \& Media).

\section{Semantic Operation Analysis} \label{app:semantic_ops}

Each editing task in \benchmark{} is annotated with the semantic operations it requires (see Appendix~\ref{app:env_scaling}), drawn from a set of 11 operations. We report the distribution of these operations across all editing tasks in the benchmark, and study the relative difficulty of editing tasks based on these operations for current LLMs.

\begin{figure*}[t]
    \centering
    \begin{minipage}{0.8\textwidth}
    \centering
    \begin{subfigure}[t]{0.60\textwidth}
        \vspace{0pt}
        \centering
        \includegraphics[width=\textwidth]{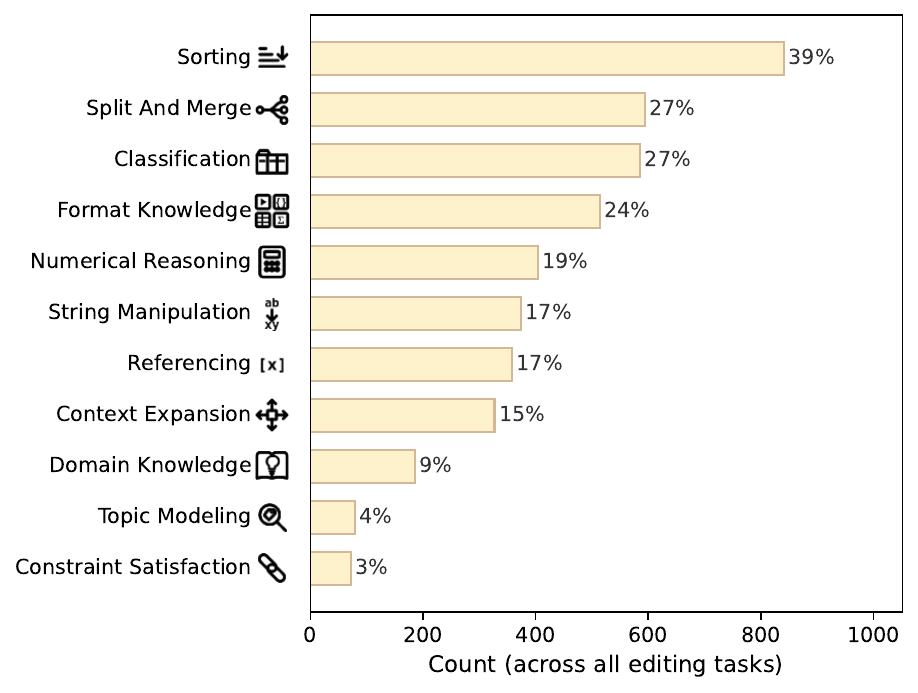}
        \subcaption{Operation Frequency}
        \label{fig:semantic_ops_freq}
    \end{subfigure}\hfill
    \begin{subfigure}[t]{0.38\textwidth}
        \vspace{0pt}
        \centering
        \includegraphics[width=\textwidth]{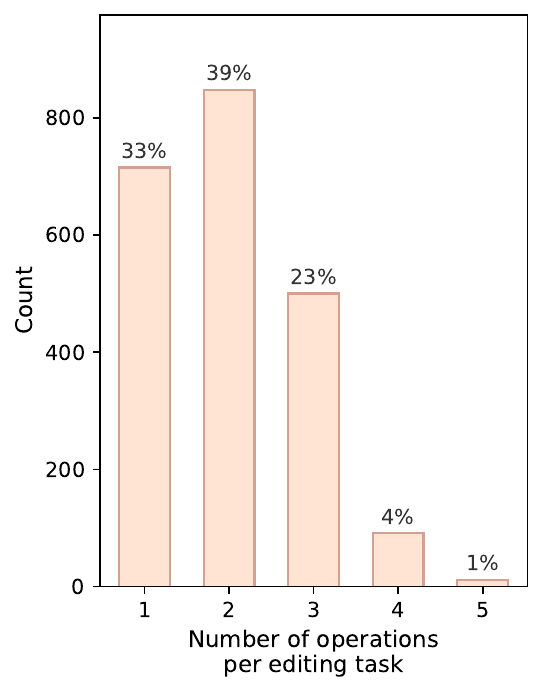}
        \subcaption{Operations per Task}
        \label{fig:semantic_ops_dist}
    \end{subfigure}
    \end{minipage}
    \caption{Semantic operation analysis of the editing tasks in \benchmark{}.
    (\subref{fig:semantic_ops_freq})~Operation frequency across editing tasks: some operations (sorting) are more common than others (constraint satisfaction).
    (\subref{fig:semantic_ops_dist})~Editing tasks typically involve 2 (39\%) or 1\% (33\%) operations, but some require up to 5 simultaneously, with a mean of 2.0 operations per task.
    The operation difficulty analysis is presented in Figure~\ref{fig:semantic_ops_main} (main text).}
    \label{fig:semantic_operations}
\end{figure*}

\paragraph{Operation distribution.} Figure~\ref{fig:semantic_ops_freq} shows the frequency of each semantic operation across editing tasks, and Figure~\ref{fig:semantic_ops_dist} shows that most editing tasks involve 1 or 2 semantic operations (72\% combined), and occasionally involve 4 or more (5\%).

\paragraph{Operation difficulty.} To study the relationship between semantic operations and editing difficulty in isolation, we use single-round-trip results from the edit testing phase (Appendix~\ref{app:edit_qa}) rather than sequential multi-round simulations. This isolates the difficulty of each edit task independent of error accumulation from prior rounds in the work environment. We ran individual round trips with GPT~5.2 and compute the point-biserial correlation between each operation's presence (binary) and the backward reconstruction score across 14,973 round trips. Figure~\ref{fig:semantic_ops_main} (main text) presents the results as a forest plot.

Three operations are significantly associated with \textit{lower} reconstruction scores: \textit{split and merge} ($r_{pb} = {-}0.080$, $p < 0.001$), \textit{classification} ($r_{pb} = {-}0.076$, $p < 0.001$), and \textit{format knowledge} ($r_{pb} = {-}0.060$, $p < 0.001$). These operations often require \textit{global} document restructuring, where the model must reason about the full document structure and where information can be silently dropped or misrouted.

Conversely, three operations are significantly associated with \textit{higher} scores: \textit{string manipulation} ($r_{pb} = {+}0.068$, $p < 0.001$), \textit{referencing} ($r_{pb} = {+}0.043$, $p < 0.001$), and \textit{context expansion} ($r_{pb} = {+}0.023$, $p < 0.01$). These might involve a larger number of \textit{local} operations where the model can operate on individual tokens or passages without needing global document understanding.

\paragraph{Number of operations.} We further find that number of semantic operations an editing task involves is negatively correlated with reconstruction score (Spearman $r = {-}0.043$, $p < 0.001$). Mean scores decline monotonically from 94.0\% for single-operation tasks to 82.6\% for tasks requiring five simultaneous operations, suggesting that compound tasks are more challenging as operations must be coordinated.

In summary, this analysis suggests that the \textit{type of cognitive operation} the editing task requires influences the difficulty of the task for current LLMs. We note that since this analysis is based on experiments involving a single model, the reported correlations may not generalize to all models. The results should be interpreted as preliminary evidence rather than a generalizable pattern.

\section{Context-Size Experiment} \label{app:context_size_exp}

This section details the experimental setup for the context-size ablation reported in Section~\ref{sec:context_size}, which isolates the effect of document size on degradation during simulated delegated workflows.

\subsection{Domain and Size Selection}

We selected five domains for this experiment, one from each benchmark category: Accounting (Structured Record), Calendar (Everyday), Playlist (Creative \& Media), Satellite (Science \& Engineering), and Spreadsheet (Code \& Configuration). Domains were selected based on their ability to be scaled (they consisted of a list of entries that could be removed without losing document coherence).

For each domain, we produced six document-size variants at approximately 1k, 2k, 4k, 6k, 8k, and 10k tokens, yielding 30 work environment variants in total. The experiment was run with a single model (GPT~5.4) under the same conditions as the main experiment (10 round-trips, no tools), with document size as the only varying parameter.

\subsection{Work Environment Construction}

For each domain, we first constructed a document of approximately 10,000 tokens from a real-world document, sourced in a process that is similar to main work environments in the benchmark. Smaller size variants were then derived from the 10k document through balanced ablation: entries were grouped by a primary categorical field in the domain (e.g., expense category, calendar track, rotation status) and proportionally downsampled at each target size, ensuring that the distribution of categories is preserved and that the resulting document remains well-formed.

A single set of six reversible edits was written per domain based on the 10k token document and shared identically across all six size variants. Edits were designed to be \textit{size-agnostic}: they reference structural properties of the document (e.g., ``split by category,'' ``sort by date'') rather than specific entries or hardcoded counts, so that each edit remains feasible and well-defined at every scale. By construction, the same edit prompts applied to documents ranging from ${\sim}$850 to ${\sim}$10,000 tokens produce analogous transformations, enabling direct comparison of degradation across document sizes. No distractor context was included in this experiment.

\section{Image Domain} \label{app:image_domain}

This section details the image editing domain introduced in Section~\ref{sec:image_domain}, which extends \benchmark{} beyond textual documents.

\paragraph{Image Selection.} We selected 6 photographs from Wikipedia (all public domain), spanning diverse visual subjects. Each image was resized to 512$\times$512 pixels (PNG).

\paragraph{Edit Design.} Each work environment has 6--7 forward/backward edit pairs that matches the structure of textual domains. Edits target domain-specific visual transformations such as color changes (e.g., ``change foliage to autumn colors''), style transfers (e.g., ``re-render in Van Gogh's style''), lighting modifications (e.g., ``add Rembrandt lighting''), object replacement (e.g., ``replace chicken with salmon''), and atmospheric effects (e.g., ``add monsoon rain''). Each forward edit is paired with a reverse instruction designed to recover the original image (e.g., ``change autumn foliage to more spring-like green'').

\paragraph{Execution.} We evaluated models with dedicated image generation capabilities. The model receives the current image together with a text prompt describing the requested edit. The prompt template instructs the model to \textit{``change as little as possible apart from what is explicitly requested.''} Each work environment consists of a single 512$\times$512 image with no distractor context.

\paragraph{Evaluation.} We use a composite perceptual similarity metric that compares the generated image against the original reference. The metric combines three components:

\begin{itemize}[nosep]
    \item \textbf{SSIM} (structural similarity on RGB channels): 50\% weight --- captures structural degradation while preserving sensitivity to color changes.
    \item \textbf{HSV histogram correlation}: 25\% weight --- captures global color distribution fidelity across hue, saturation, and value channels.
    \item \textbf{Pixel similarity} ($1 - \text{RMSE} \times 2.5$, clamped to $[0, 1]$): 25\% weight --- captures per-pixel deviations with a steep penalty curve.
\end{itemize}

\noindent The weights were calibrated through ablation testing to ensure appropriate metric behavior: identical images score 1.0, minor distortions (e.g., JPEG compression, light blur) score 0.87--0.97, moderate transformations (e.g., grayscale conversion) score 0.60--0.80, and severe distortions (e.g., random noise) score below 0.10.

\begin{figure*}[p]
    \centering
    \includegraphics[width=\textwidth]{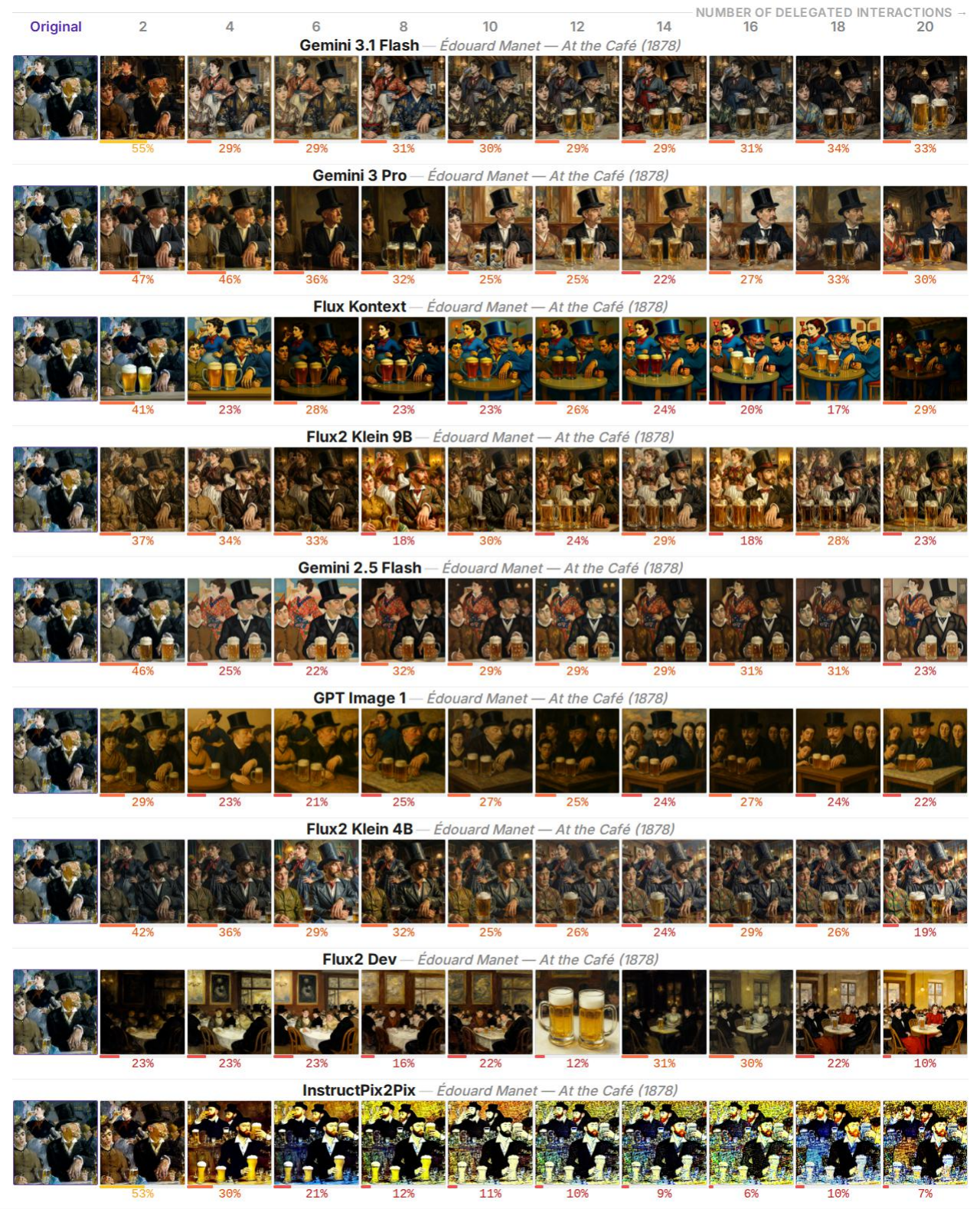}
    \caption{Image domain gallery: visual degradation over 20 delegated interactions (10 round trips) for 9 image generation models. Each row shows one model's output after each round trip, starting from the original image (left). All models exhibit progressive degradation, with weaker models losing fidelity within the first few interactions.}
    \label{fig:image_domain_gallery}
    \label{app:image_domain_gallery}
\end{figure*}

\paragraph{Models.} We tested 9 models with image generation capabilities from four families: Instruct~Pix2Pix~\cite{Brooks2022InstructPix2PixLT}, GPT~Image~1~\cite{Hurst2024GPT4oSC} (OpenAI), Flux~Kontext~\cite{Labs2025FLUX1KF}, Flux2~Dev, Flux2~Klein~4B, and Flux2~Klein~9B~\cite{flux-2-2025} , and Gemini~2.5~Flash~Image, Gemini~3~Pro~Image, and Gemini~3.1~Flash~Image~\cite{Comanici2025Gemini2P} (Google). Figure~\ref{fig:image_domain_gallery} shows representative outputs from all 9 models over 10 round trips on a single work environment, illustrating the visual degradation patterns.

\section{Dataset Creation Process} \label{app:dataset_creation}

\begin{figure*}[t]
    \centering
    \includegraphics[width=\textwidth]{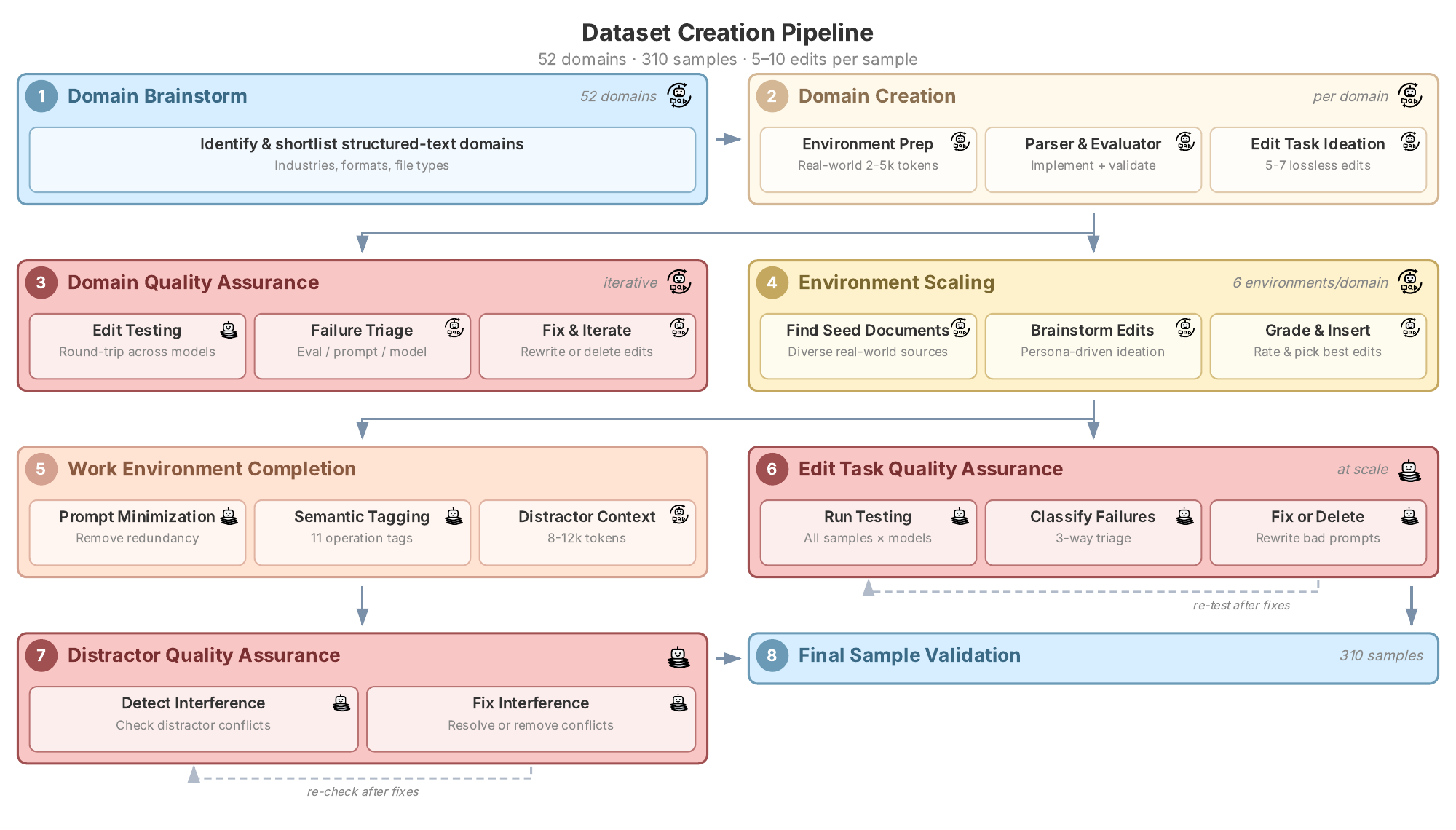}
    \caption{\benchmark{} was created using a human-directed, semi-automated agentic workflows (delegated work). The project was primarily implemented in Python in Visual Studio Code, with Claude 4.5 Opus and Claude 4.6 Opus as the main LLM agents.}
    \label{fig:dataset_creation_process}
\end{figure*}

The creation of \benchmark{} followed an eight-stage pipeline, illustrated in Figure~\ref{fig:dataset_creation_process}. The process combined human-directed oversight with semi-automated agentic workflows: at each stage, LLM-powered subagents performed structured subtasks (brainstorming, classification, prompt writing) while a human researcher reviewed outputs, made design decisions, and triggered iteration. We describe each stage below.

\subsection{Stage 1: Domain Brainstorm} \label{app:domain_brainstorm}
Domain identification followed a two-phase hierarchical brainstorm. In Phase~1, we enumerated approximately 50 general areas or industries likely to contain domain-specific textual artifacts used in professional knowledge work. In Phase~2, for each general area, subagents independently brainstormed specific document types within that area, evaluating candidates on four criteria: (1)~the area involves creating and editing structured textual artifacts, (2)~the area uses domain-specific file formats or notations (not generic office documents), (3)~real-world examples are publicly available online, and (4)~the domain is distinct from those already selected. This process produced over 100 candidate domains. We shortlisted candidates through additional prioritization filters: well-defined text format with clear syntax, publicly downloadable with minimal friction, support for multiple interesting edit types, moderate complexity (challenging but tractable for evaluation), and diversity across professions. We additionally measured domain popularity on GitHub by counting files with domain-specific extensions, as a proxy for LLM training data density. The final selection of 52 domains spans five categories (Science \& Engineering, Code \& Configuration, Creative \& Media, Structured Record, and Everyday), covering a broad range of structured textual formats from crystallography files to textile patterns.

\subsection{Stage 2: Domain Creation} \label{app:domain_creation}
For each of the 52 domains, we created an initial work environment consisting of a seed document, a domain-specific parser and evaluator, and a set of initial edit tasks. This stage is broken into three sub-steps.

\textit{First Environment Preparation.} Seed documents were sourced primarily through GitHub code search, filtered by file extension and size. Each candidate was validated against the desiderata listed in \autoref{tab:document_desiderata}. When raw documents exceeded the 2--5k token target, they were preprocessed (trimming, comment removal, normalization) using domain-specific Python scripts. Provenance metadata (source URL, license, search query used) was recorded for each selected document.

\textit{Parser \& Evaluator.} For each domain, we implemented a Python module exposing three methods: \texttt{parse\_context} (converts raw document files into a structured representation), \texttt{compute\_domain\_statistics} (extracts summary metrics), and \texttt{evaluate\_context} (scores semantic equivalence between a candidate and reference document on a $[0, 1]$ scale). Implementation prioritized existing parsing libraries (e.g., \texttt{python-chess} for PGN, \texttt{python-ly} for LilyPond, \texttt{icalendar} for iCal) over custom parsers, to maximize robustness across the diversity of real-world documents. Domain modules also include an optional \texttt{preprocess\_context} method that normalizes common syntax errors observed in small-scale LLM simulations (e.g., extra whitespace, alternate quoting styles, code fences in output, case differences in keywords), making the parser more forgiving of minor syntax-sugar deviations without inflating scores for genuinely incorrect outputs. Each evaluator was validated through ablation testing: removing $K$ out of $N$ logical blocks from a document must produce a score no higher than $1 - K/N$, ensuring proportional sensitivity to content loss.

\textit{Edit Task Ideation.} For the initial environment, we designed 5--7 reversible edit tasks. Edits were brainstormed using a persona-driven approach: for each document, we identified 3--5 realistic personas (e.g., auditor, grant writer, board treasurer for an accounting ledger) and designed edits reflecting the transformations each persona would plausibly request. Each edit task consists of a forward instruction and a backward instruction that must be lossless when composed. Nine rules governed edit design: edits must be domain-specific (Rule~1), fully reversible with no information loss (Rule~2), transformative rather than purely expansive (Rules~3, 4, 9), written in natural language without mentioning reversibility (Rules~5, 6), use wildcards when filenames should not be revealed (Rule~7), and include ordering metadata when splitting or merging (Rule~8). 

\subsection{Stage 3: Domain Quality Assurance} \label{app:domain_qa}
After creating the initial work environment for each domain, we ran an iterative quality assurance loop to validate and refine the parser, and evaluator of the domain. The loop consists of three sub-steps.

\textit{Edit Testing.} Each forward--backward edit pair was tested in isolation by running round-trip evaluations: starting from the seed document, applying the forward edit with an LLM, then applying the backward edit to the result, and scoring the reconstruction against the original. Tests were run with two models (GPT~5.2 and GPT~5 Mini) for 5 runs per edit pair, providing statistical reliability.

\textit{Failure Triage.} Low-scoring round trips (score $< 0.1$) were inspected and classified into three categories: (a)~\textit{warranted} failures, where the LLM output was genuinely poor, (b)~\textit{parser failures}, where the output was semantically correct but the evaluator rejected it due to minor formatting differences, and (c)~\textit{ambiguous} cases requiring deeper investigation.

\textit{Fix \& Iterate.} Based on the triage, we applied targeted fixes. For parser failures, we added preprocessing normalization to the evaluator (e.g., tolerating whitespace variants, alternate quoting) to reduce false negatives without inflating scores for genuinely incorrect outputs. After fixes, edit testing was re-run to confirm improvements, and the loop repeated until all identified issues were resolved.

\subsection{Stage 4: Environment Scaling} \label{app:env_scaling}
The domain creation stage produced one work environment per domain. In this stage, we scaled to six work environments per domain by finding five additional seed documents and creating edit tasks for each. The process follows three sub-steps.

\textit{Find Seed Documents.} For each domain, we searched for diverse real-world documents using GitHub code search, public data portals, and open-source project repositories. Diversity was explicitly targeted along four axes: subject matter, structural complexity, size within the 2--5k token range, and authorship style. Each candidate was validated against the document desiderata (\autoref{tab:document_desiderata}), tested for parser compatibility, and documented with full provenance.

\textit{Brainstorm Edits.} Edit tasks for the new work environments were generated using persona-driven ideation, following the same approach as the initial domain creation.

\subsection{Stage 5: Work Environment Completion} \label{app:env_completion}
Before quality assurance at scale, each work environment was completed with three additional components.

\textit{Prompt Minimization.} Forward and backward prompts were minimized by an LLM to remove redundant information: standard domain conversions that any competent practitioner would know, repeated phrasings, and patterns obvious from a single example. The objective was to ensure each prompt is realistic as users are unlikely to provide obvious information in instructions, and that instructions contain exactly the information needed to perform the edit. Motivational or background context (e.g., ``customers have trouble pronouncing dish names'') was retained in instructions, as real users often include such framing.

\textit{Semantic Tagging.} Each edit task was tagged with the semantic operations it requires, drawn from a set of 11 operations: \textit{numerical reasoning}, \textit{constraint satisfaction}, \textit{split and merge}, \textit{topic modeling}, \textit{classification}, \textit{domain knowledge}, \textit{format knowledge}, \textit{string manipulation}, \textit{sorting}, \textit{context expansion}, and \textit{referencing}. Operations were assigned by an LLM based on the actual prompt text, with strict criteria: an operation was applied only if the edit thoroughly involves it, not merely incidentally. Operations are assigned for an edit task, considering both the forward and backward edit instructions.

\textit{Distractor Context.} For each work environment, we curated 1--5 distractor documents totaling 8--12k tokens. Distractor documents are topically related to the seed document but must not interfere with any edit task. Distractor creation followed eleven criteria (D1--D11), including: topical relatedness, non-interference with edits, heterogeneous file formats, licensed for redistribution, real-world sourced (not synthetic), not overly famous, and at most 50\% sourced from Wikipedia to ensure source diversity. Distractors were sourced from GitHub repositories, Wikipedia, government documents, and open data portals.

\subsection{Stage 6: Edit Task Quality Assurance} \label{app:edit_qa}
After implementing 310 work environments in 52 domains, we ran edit testing on the 2,125 edit tasks of the benchmark. The process mirrors the domain-level QA (Stage~3) but operates at the dataset level, and focuses on correctness of edit tasks rather than the domain parsing and evaluation.

\textit{Run Testing.} Forward--backward round trips were executed for all edit tasks with two models (GPT~5.2 and GPT~5 Mini), with 5 runs per pair. Results were stored as structured logs for automated analysis.

\textit{Classify Failures.} Edits where all tested models scored below a threshold (80\%) were flagged as problematic. Each flagged edit was classified by an LLM into one of three categories: (a)~\textit{eval\_parser\_error}, where the output is semantically correct but the evaluator fails to capture the similarity, (b)~\textit{prompt\_edit\_error}, where the prompt design makes the round-trip infeasible even for a perfect model, or (c)~\textit{model\_error}, where the prompts are well-defined but the model fails due to its own limitations.

\textit{Fix or Delete.} For edits classified as parser errors, the evaluator was patched with additional preprocessing. For prompt errors, an LLM suggested either a rewrite (clarifying ambiguity, adding preservation instructions) or deletion (if the edit was fundamentally not reversible). After applying fixes, stale test logs were purged and the testing loop was repeated until convergence.

\subsection{Stage 7: Distractor Quality Assurance} \label{app:distractor_qa}
A dedicated QA stage verified that distractor documents do not interfere with edit tasks. This stage uses an LLM-based detection pipeline.

\textit{Detect Interference.} For each work environment, an LLM reviewed all edit prompts alongside the full set of basic-state and distractor files, classifying each edit as either \textit{no\_interference} or \textit{interference}. Five interference types were checked: filename collision (distractor shares a name with a task file), content confusion (distractor content so similar it could be incorporated into the edit), prompt scope ambiguity (prompt language that could apply to distractors, e.g., ``merge all CSV files''), information leakage (distractor reveals parts of the expected target), and structural interference (distractor alters interpretation of the file set).

\textit{Fix Interference.} Flagged cases were resolved through one of four actions, in order of preference: (a)~\textit{dismiss} the flag as a false positive, (b)~\textit{modify the edit task instructions} to reference files by exact filename rather than generic descriptions, (c)~\textit{delete the problematic distractor file} (provided the remaining distractor context retained at least 5,000 tokens), or (d)~\textit{delete the edit} as a last resort. After fixes, the detection pipeline was re-run to confirm resolution.

\subsection{Stage 8: Final Work Environment Validation} \label{app:final_validation}
Before finalizing the benchmark, every work environment passed a comprehensive validation suite of 23 automated checks. The checks covered structural integrity (valid JSON, required keys, state graph connectivity, no orphan states), context properties (token count within the 2--5k range, no triple backticks, files exist on disk), edit requirements (minimum 4 forward edits, prompts non-empty and free of reversibility-revealing language), metadata completeness (provenance URL, license, redistribution status, semantic operation tags, state summaries $\leq$25 words), distractor integrity (metadata matches files on disk, token budget met), and domain API verification (self-evaluation of the seed document scores exactly 1.0). The final benchmark comprises 310 validated work environments across 52 domains, totaling 2,125 edit tasks.

\section{Model Details} \label{app:model_details}

Table~\ref{tab:model_details} lists the exact model versions and providers used in our experiments.

\begin{table*}[h!]
\input{tables/model_details}
\caption{Model details for the 19 LLMs evaluated in \benchmark{}.}
\label{tab:model_details}
\end{table*}

\section{Agentic Harness (Operating Models with Tools)} \label{app:agentic_harness}

This section details the agentic harness used in the experiments of Section~\ref{sec:agentic}, where models are given tools and allowed to iteratively edit workspace files rather than generating modified documents in a single instruction response.

\subsection{Tools and Execution Environment}

The agentic harness provides the model with five tools via the OpenAI function-calling interface:

\begin{enumerate}
    \item \texttt{read\_file(filename)}: Read the full contents of a workspace file.
    \item \texttt{write\_file(filename, content)}: Create or overwrite a file with the given content.
    \item \texttt{delete\_file(filename)}: Remove a file from the workspace.
    \item \texttt{run\_python(code)}: Execute a Python script with read/write access to workspace files.
    \item \texttt{finish()}: Signal completion of the task.
\end{enumerate}

Files are backed by an in-memory virtual filesystem. The \texttt{run\_python} tool materializes workspace files to a temporary directory, executes the script, and syncs any file changes back to the virtual filesystem. Python execution is sandboxed using \texttt{bubblewrap} (\texttt{bwrap}): system libraries and the Python interpreter are mounted read-only, the script has read-write access only to the temporary workspace directory, and network access is disabled (\texttt{--unshare-net}). Scripts are subject to a 30-second timeout, and their output is truncated to 10,000 characters to prevent context blowup.

\subsection{Iteration Budget and Termination}

The agentic loop runs for a maximum of 25 LLM round-trips (turns), with an additional hard stop at 500,000 cumulative tokens. In practice, all tested models complete their editing tasks well within this budget: the top-performing models reach the \texttt{finish()} signal in over 99\% of runs before exhausting the turn limit.

The harness enforces a \textit{write-before-finish} constraint: the model must perform at least one write operation (\texttt{write\_file} or \texttt{run\_python}) before the \texttt{finish()} tool is accepted. If the model calls \texttt{finish()} prematurely, it receives an error message instructing it to read the files, apply the requested edits, and then call \texttt{finish()}. If the model stops issuing tool calls without having written anything, the harness sends a nudge message asking it to use the tools.

\subsection{Distractor Handling}

In the single-shot setting, the full content of all files (including distractors) is provided in the initial prompt; the model must process the entire context to produce its output. In the agentic setting, only the \textit{filename} of all workspace files (including distractors) is listed in the initial prompt; the model must explicitly call \texttt{read\_file} to inspect any file's contents. The agentic setting is therefore structurally advantaged with respect to distractor handling: the model can choose to read only the files relevant to the task and ignore the rest, whereas the single-shot model is forced to process all content. We log every \texttt{read\_file} call to track which files (including distractors) each model reads, enabling post-hoc analysis of distractor interaction in the agentic setting. We find in our simulations that models tend to read an average of 20\% of distractor files during simulation, confirming the advantage the agentic setting has over the single-shot setting.

At the end of each agentic run, distractor files are stripped from the output snapshot to prevent context bloat across successive round-trips. Distractor files are then reintroduced in the next simulated interaction. In other words, modifications to distractor files by the model are not preserved. 

\subsection{Logging and Metadata}

For each agentic run, we log the following metadata:

\begin{itemize}
    \item \textbf{Token usage}: Cumulative prompt tokens, completion tokens, and reasoning tokens (when available) across all turns.
    \item \textbf{Cost}: Total monetary cost in USD, based on model per-token pricing.
    \item \textbf{Latency}: Wall-clock time for each LLM call and cumulative latency.
    \item \textbf{Turn count}: Number of LLM round-trips used.
    \item \textbf{Tool call log}: For each tool invocation, we record the tool name, the turn number, and the argument keys (file contents and code are omitted for compactness).
    \item \textbf{Operation sequence}: Ordered list of tool names, enabling high-level tracing of the model's editing strategy (e.g., read-read-write-finish vs.\ read-run\_python-finish).
    \item \textbf{Files read}: Every filename passed to \texttt{read\_file}, used for distractor analysis.
    \item \textbf{Clean finish}: A boolean indicating whether the model called \texttt{finish()} (or stopped issuing tool calls after writing) versus hitting the turn or token budget.
\end{itemize}

\section{Document Desiderata} \label{app:document_desiderata}

Each document in \benchmark{} was curated according to the desiderata listed in \autoref{tab:document_desiderata}.

\begin{table*}[h!]
\input{tables/document_desiderata}
\caption{Desiderata guiding the curation of documents in \benchmark{}.}
\label{tab:document_desiderata}
\end{table*}

\end{document}

%% file: shortcut.tex
\newcommand{\benchmark}{\texttt{DELEGATE-52}}

\newcommand{\ccol}[1]{\cellcolor[HTML]{#1}}

\usepackage{soul}
\newcommand{\chbox}[2]{{\definecolor{chbox@tmp}{HTML}{#1}\sethlcolor{chbox@tmp}\hl{#2}}}

\newcolumntype{/}{!{\color{white}\vline width 7pt}}

\newcommand{\cmark}{\textcolor{black!40}{$\checkmark$}}

\newcommand{\arrowfill}{%
  \leavevmode\leaders\hrule height 0.4pt\hfill\kern0pt\ensuremath{\rightarrow}%
}

\newcommand{\symbolimg}[2][0.35cm]{%
  \ensuremath{\vcenter{\hbox{\includegraphics[height=#1]{#2}}}}%
}
\newcommand{\gemini}{\symbolimg[0.35cm]{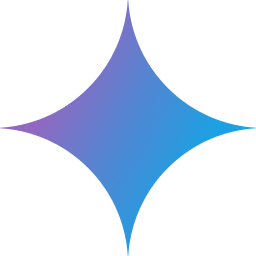}}
\newcommand{\claude}{\symbolimg[0.35cm]{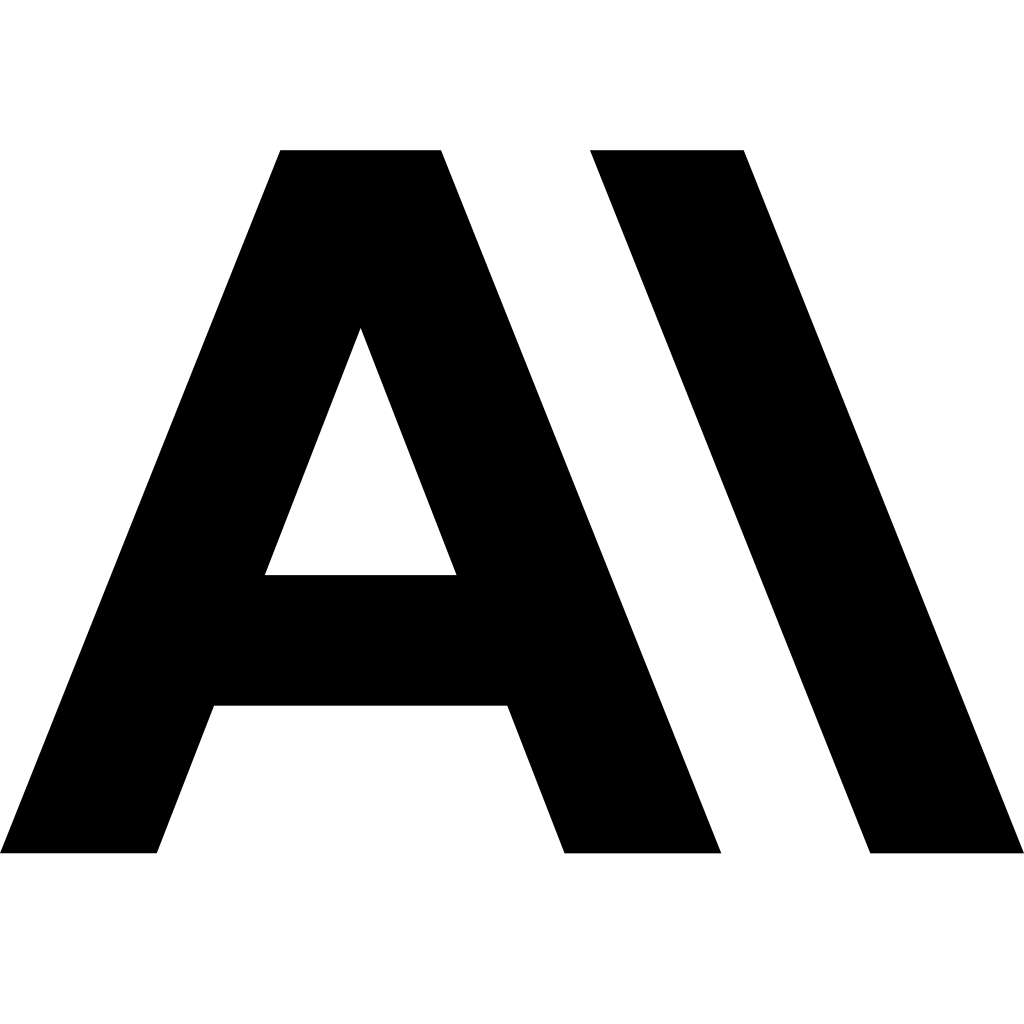}}
\newcommand{\openai}{\symbolimg[0.35cm]{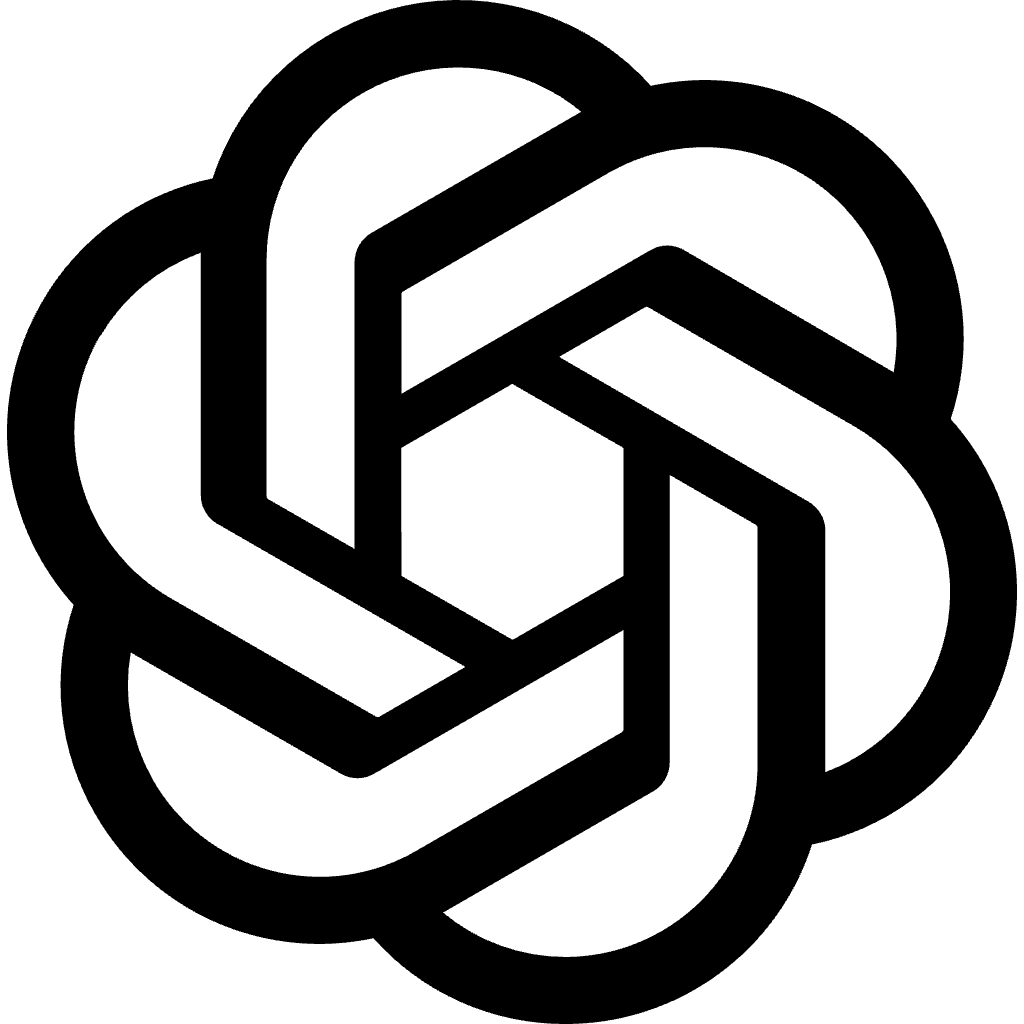}}
\newcommand{\xai}{\symbolimg[0.35cm]{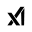}}
\newcommand{\mistral}{\symbolimg[0.35cm]{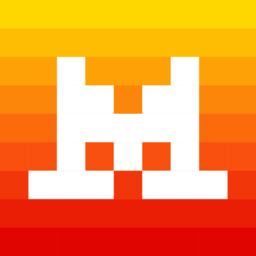}}
\newcommand{\bfl}{\symbolimg[0.35cm]{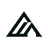}}
\newcommand{\berkeley}{\symbolimg[0.35cm]{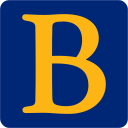}}

\newcommand{\moonshot}{\symbolimg[0.35cm]{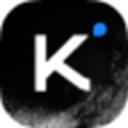}}

\newcommand{\domainicon}[1]{\symbolimg[0.3cm]{assets/domain_icons_black/#1.pdf}}

\newcommand{\icondirect}{\symbolimg[0.45cm]{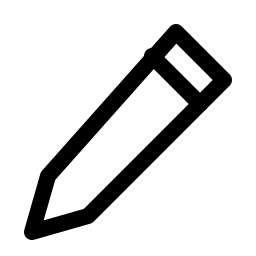}}
\newcommand{\iconagentic}{\symbolimg[0.45cm]{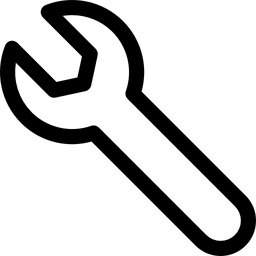}}

\newcommand{\iconroundrobin}{\symbolimg[0.45cm]{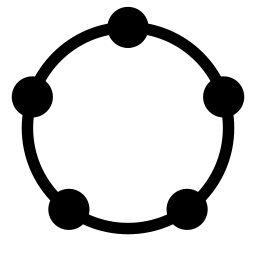}}
\newcommand{\iconsingleedit}{\symbolimg[0.45cm]{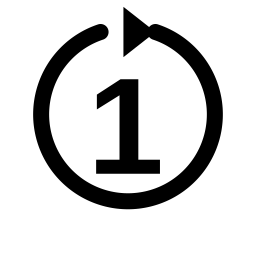}}

\newcommand{\ctxdot}[1]{\tikz[baseline=-0.6ex]\fill[black] circle (#1);}

\newcommand{\iconexec}{\symbolimg[0.4cm]{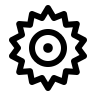}}
\newcommand{\iconwrite}{\symbolimg[0.4cm]{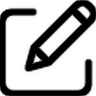}}

\newcommand{\domainheader}[2]{%
  \begin{tabular}[b]{@{}c@{}}
    \rotatebox{90}{\scriptsize #2}\\[-2pt]
    \domainicon{#1}
  \end{tabular}%
}

\definecolor{darkblue}{rgb}{0, 0, 0.5}

\hypersetup{colorlinks=true, citecolor=darkblue, linkcolor=darkblue, urlcolor=darkblue}

%% file: tables/overall_by_round.tex
\centering
\resizebox{\linewidth}{!}{%
\renewcommand{\arraystretch}{1.6}
\begin{tabular}{lrrrrrrrrrr}
 & \multicolumn{10}{c}{\Large{\textbf{\benchmark}}} \\[-6pt]
 & \multicolumn{10}{c}{\textbf{Workflow length $k$ (\# interactions)}} \\[-11pt]

& & & & & & & & & & \multicolumn{1}{r@{\kern-1pt}}{$\rightarrow$} \\[-10pt]
\cmidrule[1.0pt]{2-11}
\textbf{Model} & \multicolumn{1}{c}{2} & \multicolumn{1}{c}{4} & \multicolumn{1}{c}{6} & \multicolumn{1}{c}{8} & \multicolumn{1}{c}{10} & \multicolumn{1}{c}{12} & \multicolumn{1}{c}{14} & \multicolumn{1}{c}{16} & \multicolumn{1}{c}{18} & \multicolumn{1}{c}{20} \\
\cmidrule(r){1-1}\cmidrule{2-11}
\openai\hspace{1pt} GPT 5 Nano & \ccol{E19898} 30.3 & \ccol{D88F8F} 17.4 & \ccol{D58C8C} 12.8 & \ccol{D58B8B} 12.2 & \ccol{D48B8B} 11.4 & \ccol{D48A8A} 11.1 & \ccol{D48A8A} 10.5 & \ccol{D48A8A} 10.3 & \ccol{D48A8A} 10.1 & \ccol{D48A8A} 10.0 \\
\openai\hspace{1pt} GPT 4o & \ccol{EEAEAE} 45.6 & \ccol{E19898} 29.6 & \ccol{DD9494} 23.9 & \ccol{DA9191} 19.9 & \ccol{D99090} 18.8 & \ccol{D88F8F} 17.3 & \ccol{D88E8E} 16.5 & \ccol{D88E8E} 16.2 & \ccol{D78E8E} 15.6 & \ccol{D68D8D} 14.7 \\
\openai\hspace{1pt} OSS 120B & \ccol{FEE7D0} 73.1 & \ccol{F6BFBF} 52.4 & \ccol{E59D9D} 36.5 & \ccol{E09797} 28.3 & \ccol{DD9494} 25.0 & \ccol{DC9292} 22.2 & \ccol{DA9191} 20.3 & \ccol{DA9191} 20.2 & \ccol{DA9191} 19.8 & \ccol{D99090} 19.2 \\
\mistral\hspace{1pt} Large 3 & \ccol{FEF3D3} 82.4 & \ccol{FEE5D1} 71.8 & \ccol{FACFCA} 59.8 & \ccol{F7C3C3} 53.8 & \ccol{EFAFAF} 46.1 & \ccol{EBA8A8} 43.4 & \ccol{E8A1A1} 40.7 & \ccol{E69D9D} 37.4 & \ccol{E59C9C} 35.9 & \ccol{E59C9C} 35.5 \\
\gemini\hspace{1pt} 3 Flash & \ccol{FEECCE} 76.0 & \ccol{FBD3CB} 61.6 & \ccol{F9C9C7} 57.1 & \ccol{F2B8B8} 49.6 & \ccol{F0B3B3} 47.5 & \ccol{EBA7A7} 42.8 & \ccol{E9A2A2} 41.1 & \ccol{E79F9F} 39.5 & \ccol{E59D9D} 36.6 & \ccol{E59C9C} 35.8 \\
\openai\hspace{1pt} GPT 5 Mini & \ccol{FEF6E1} 86.3 & \ccol{FEEACF} 75.1 & \ccol{FDDBCF} 66.2 & \ccol{FBD0CA} 60.4 & \ccol{F9C6C6} 55.2 & \ccol{F3BABA} 50.5 & \ccol{F1B4B4} 48.1 & \ccol{EFB1B1} 47.0 & \ccol{EEAEAE} 45.6 & \ccol{EDACAC} 45.1 \\
\openai\hspace{1pt} GPT 5 Chat & \ccol{FEF4D7} 83.3 & \ccol{FEE8D0} 73.3 & \ccol{FDDBCF} 66.0 & \ccol{FBD0CA} 60.2 & \ccol{F9C9C7} 56.4 & \ccol{F6C1C1} 53.0 & \ccol{F4BBBB} 50.9 & \ccol{F2B6B6} 49.1 & \ccol{F0B3B3} 47.8 & \ccol{EFB0B0} 46.8 \\
\openai\hspace{1pt} o1 & \ccol{FEF6E1} 86.4 & \ccol{FEECCE} 76.7 & \ccol{FEE0D1} 68.6 & \ccol{FCD5CD} 63.3 & \ccol{FACBC8} 57.6 & \ccol{F7C3C3} 53.9 & \ccol{F6C1C1} 53.2 & \ccol{F3B9B9} 50.2 & \ccol{F2B7B7} 49.2 & \ccol{F1B4B4} 48.1 \\
\openai\hspace{1pt} o3 & \ccol{FEF5DC} 85.2 & \ccol{FEEACF} 75.2 & \ccol{FDDBCF} 65.9 & \ccol{FBD1CA} 60.7 & \ccol{FACBC8} 58.1 & \ccol{F7C2C2} 53.5 & \ccol{F4BBBB} 50.8 & \ccol{F2B7B7} 49.4 & \ccol{F2B6B6} 48.9 & \ccol{F1B5B5} 48.2 \\
\openai\hspace{1pt} GPT 5 & \ccol{FFFAF0} 91.5 & \ccol{FEF2CF} 80.9 & \ccol{FEE5D1} 71.6 & \ccol{FDDBCF} 66.3 & \ccol{FBD3CC} 62.1 & \ccol{FACDC9} 58.5 & \ccol{F9C7C6} 55.9 & \ccol{F7C2C2} 53.3 & \ccol{F4BCBC} 51.4 & \ccol{F1B5B5} 48.3 \\
\openai\hspace{1pt} GPT 4.1 & \ccol{FEF8E8} 88.9 & \ccol{FEF1CC} 79.8 & \ccol{FEE4D2} 70.9 & \ccol{FEDED0} 67.7 & \ccol{FBD3CC} 62.2 & \ccol{F9C9C7} 56.8 & \ccol{F8C5C5} 54.8 & \ccol{F5BDBD} 51.7 & \ccol{F2B8B8} 49.8 & \ccol{F2B7B7} 49.5 \\
\xai\hspace{1pt} Grok 4 & \ccol{FFFAF0} 91.7 & \ccol{FEF5DD} 85.4 & \ccol{FEF0CC} 78.5 & \ccol{FEE9D0} 74.0 & \ccol{FEE1D2} 69.0 & \ccol{FDDDD0} 67.2 & \ccol{FDDACF} 65.4 & \ccol{FBD3CC} 62.1 & \ccol{FBD2CB} 61.4 & \ccol{FACEC9} 59.3 \\
\openai\hspace{1pt} GPT 5.1 & \ccol{FFF9EE} 90.8 & \ccol{FEF3D5} 82.8 & \ccol{FEEECD} 78.0 & \ccol{FEE9D0} 74.0 & \ccol{FEE3D2} 69.9 & \ccol{FDDCD0} 66.7 & \ccol{FCD9CE} 64.9 & \ccol{FCD5CD} 62.9 & \ccol{FBD3CB} 61.7 & \ccol{FBD0CA} 60.5 \\
\moonshot\hspace{1pt} Kimi K2.5 & \ccol{FFF9EE} 91.1 & \ccol{FEF6E0} 86.1 & \ccol{FEF4D6} 83.0 & \ccol{FEEBCF} 75.6 & \ccol{FEE8D0} 73.3 & \ccol{FEE3D2} 70.0 & \ccol{FEE1D2} 68.8 & \ccol{FDDCD0} 66.4 & \ccol{FCD9CE} 64.9 & \ccol{FCD7CD} 64.1 \\
\claude\hspace{1pt} 4.6 Sonnet & \ccol{FFFAF1} 92.2 & \ccol{FEF6DE} 85.7 & \ccol{FEF3D1} 81.8 & \ccol{FEEFCD} 78.2 & \ccol{FEEACF} 74.9 & \ccol{FEE5D1} 71.7 & \ccol{FEE3D2} 70.2 & \ccol{FEE1D2} 69.1 & \ccol{FDDDD0} 66.9 & \ccol{FDDBCF} 66.0 \\
\openai\hspace{1pt} GPT 5.2 & \ccol{FFFBF2} 92.7 & \ccol{FEF6E2} 86.9 & \ccol{FEF3D3} 82.2 & \ccol{FEEECD} 77.9 & \ccol{FEE9CF} 74.4 & \ccol{FEE5D1} 71.6 & \ccol{FEE3D2} 70.0 & \ccol{FEE0D1} 68.5 & \ccol{FDDDD0} 67.1 & \ccol{FDDBCF} 66.1 \\
\openai\hspace{1pt} GPT 5.4 & \ccol{FFFCF6} 94.3 & \ccol{FEF8E9} 89.3 & \ccol{FEF5DD} 85.4 & \ccol{FEF3D2} 82.0 & \ccol{FEF1CC} 79.4 & \ccol{FEECCE} 76.4 & \ccol{FEE9CF} 74.6 & \ccol{FEE7D0} 73.1 & \ccol{FEE6D1} 72.1 & \ccol{FEE5D2} 71.5 \\
\claude\hspace{1pt} 4.6 Opus & \ccol{FFFCF5} 94.2 & \ccol{FFF9EC} 90.1 & \ccol{FEF6E2} 86.8 & \ccol{FEF3D3} 82.5 & \ccol{FEF1CC} 79.5 & \ccol{FEEECD} 78.0 & \ccol{FEECCE} 76.3 & \ccol{FEEACF} 75.2 & \ccol{FEE9CF} 74.3 & \ccol{FEE7D0} 73.1 \\
\gemini\hspace{1pt} 3.1 Pro & \ccol{FFFEFC} 96.8 & \ccol{FFFBF5} 93.5 & \ccol{FFFAEF} 91.4 & \ccol{FEF8E8} 88.9 & \ccol{FEF6E1} 86.6 & \ccol{FEF4D9} 83.9 & \ccol{FEF3D2} 82.2 & \ccol{FEF3D0} 81.2 & \ccol{FEF2CF} 80.9 & \ccol{FEF2CF} 80.9 \\
\bottomrule
\end{tabular}
}

%% file: tables/round10_by_domain.tex
\centering
\resizebox{\textwidth}{!}{%
\setlength{\tabcolsep}{0.5pt}
\renewcommand{\arraystretch}{1.6}
\begin{tabular}{l/ccccccccccc/ccccccccccc/ccccccccccc/ccccccccccc/cccccccc}
 & \multicolumn{11}{c}{\large\textbf{Code \& Config.}} & \multicolumn{11}{c}{\large\textbf{Science \& Eng.}} & \multicolumn{11}{c}{\large\textbf{Creative \& Media}} & \multicolumn{11}{c}{\large\textbf{Structured Rec.}} & \multicolumn{8}{c}{\large\textbf{Everyday}} \\
\cmidrule(r){2-12} \cmidrule(r){13-23} \cmidrule(r){24-34} \cmidrule(r){35-45} \cmidrule{46-53}
 & \domainheader{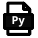}{Python} & \domainheader{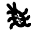}{Malware} & \domainheader{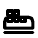}{Docker} & \domainheader{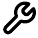}{Makefile} & \domainheader{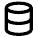}{DB Schema} & \domainheader{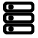}{Infra} & \domainheader{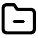}{Filesystem} & \domainheader{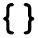}{JSON} & \domainheader{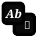}{Translation} & \domainheader{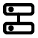}{DNS} & \domainheader{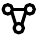}{Graphviz} & \domainheader{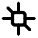}{Circuit} & \domainheader{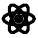}{Quantum} & \domainheader{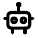}{Robotics} & \domainheader{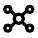}{Molecule} & \domainheader{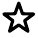}{Star Cat} & \domainheader{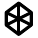}{Crystal} & \domainheader{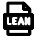}{Math Lean} & \domainheader{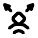}{Satellite} & \domainheader{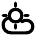}{Weather} & \domainheader{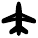}{Aviation} & \domainheader{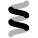}{Protein} & \domainheader{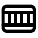}{Screenplay} & \domainheader{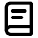}{Fiction} & \domainheader{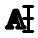}{Font Eng} & \domainheader{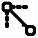}{Vector} & \domainheader{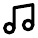}{Music} & \domainheader{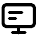}{Slides} & \domainheader{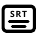}{Subtitles} & \domainheader{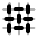}{Weaving} & \domainheader{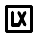}{LaTeX} & \domainheader{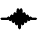}{Audio Syn} & \domainheader{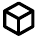}{3D Obj} & \domainheader{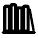}{Lib Catalog} & \domainheader{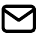}{Emails} & \domainheader{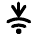}{Ham Radio} & \domainheader{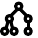}{Treebank} & \domainheader{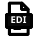}{EDIFACT} & \domainheader{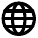}{Geodata} & \domainheader{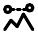}{Geotrack} & \domainheader{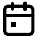}{Calendar} & \domainheader{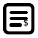}{Accounting} & \domainheader{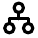}{Genealogy} & \domainheader{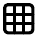}{Spreadsheet} & \domainheader{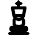}{Chess} & \domainheader{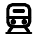}{Transit} & \domainheader{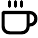}{Food Menu} & \domainheader{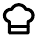}{Recipe} & \domainheader{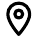}{Landmarks} & \domainheader{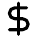}{Earnings} & \domainheader{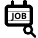}{Job Board} & \domainheader{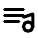}{Playlist} \\
\cmidrule(r){1-1} \cmidrule(r){2-12} \cmidrule(r){13-23} \cmidrule(r){24-34} \cmidrule(r){35-45} \cmidrule{46-53}
\openai\hspace{1pt} GPT 5 Nano & \ccol{D6EEFF}{\cmark} & \ccol{D48A8A}{} & \ccol{D48A8A}{} & \ccol{D48A8A}{} & \ccol{D48A8A}{} & \ccol{D48A8A}{} & \ccol{D48A8A}{} & \ccol{D48A8A}{} & \ccol{D48A8A}{} & \ccol{D48A8A}{} & \ccol{D48A8A}{} & \ccol{D48A8A}{} & \ccol{D48A8A}{} & \ccol{D48A8A}{} & \ccol{D48A8A}{} & \ccol{D48A8A}{} & \ccol{D48A8A}{} & \ccol{D48A8A}{} & \ccol{D48A8A}{} & \ccol{D48A8A}{} & \ccol{D48A8A}{} & \ccol{D48A8A}{} & \ccol{E8A0A0}{} & \ccol{F9C6C6}{} & \ccol{D48A8A}{} & \ccol{D48A8A}{} & \ccol{D48A8A}{} & \ccol{D48A8A}{} & \ccol{D48A8A}{} & \ccol{D48A8A}{} & \ccol{D48A8A}{} & \ccol{D48A8A}{} & \ccol{D48A8A}{} & \ccol{D48A8A}{} & \ccol{D48A8A}{} & \ccol{D48A8A}{} & \ccol{D48A8A}{} & \ccol{D48A8A}{} & \ccol{D48A8A}{} & \ccol{D48A8A}{} & \ccol{D48A8A}{} & \ccol{D48A8A}{} & \ccol{D48A8A}{} & \ccol{D48A8A}{} & \ccol{D48A8A}{} & \ccol{D48A8A}{} & \ccol{D48A8A}{} & \ccol{D48A8A}{} & \ccol{D48A8A}{} & \ccol{D48A8A}{} & \ccol{D48A8A}{} & \ccol{D48A8A}{} \\
\openai\hspace{1pt} GPT 4o & \ccol{D6EEFF}{\cmark} & \ccol{D48A8A}{} & \ccol{D48A8A}{} & \ccol{D48A8A}{} & \ccol{D48A8A}{} & \ccol{D48A8A}{} & \ccol{D48A8A}{} & \ccol{D48A8A}{} & \ccol{D48A8A}{} & \ccol{D48A8A}{} & \ccol{D48A8A}{} & \ccol{D48A8A}{} & \ccol{D48A8A}{} & \ccol{D48A8A}{} & \ccol{D48A8A}{} & \ccol{D48A8A}{} & \ccol{D48A8A}{} & \ccol{D48A8A}{} & \ccol{D48A8A}{} & \ccol{D48A8A}{} & \ccol{D48A8A}{} & \ccol{D48A8A}{} & \ccol{F9C6C6}{} & \ccol{D48A8A}{} & \ccol{D48A8A}{} & \ccol{D48A8A}{} & \ccol{D48A8A}{} & \ccol{D48A8A}{} & \ccol{D48A8A}{} & \ccol{D48A8A}{} & \ccol{D48A8A}{} & \ccol{D48A8A}{} & \ccol{D48A8A}{} & \ccol{D48A8A}{} & \ccol{D48A8A}{} & \ccol{D48A8A}{} & \ccol{D48A8A}{} & \ccol{D48A8A}{} & \ccol{D48A8A}{} & \ccol{D48A8A}{} & \ccol{D48A8A}{} & \ccol{D48A8A}{} & \ccol{D48A8A}{} & \ccol{D48A8A}{} & \ccol{D48A8A}{} & \ccol{D48A8A}{} & \ccol{D48A8A}{} & \ccol{D48A8A}{} & \ccol{D48A8A}{} & \ccol{D48A8A}{} & \ccol{D48A8A}{} & \ccol{D48A8A}{} \\
\openai\hspace{1pt} OSS 120B & \ccol{D6EEFF}{\cmark} & \ccol{D48A8A}{} & \ccol{D48A8A}{} & \ccol{D48A8A}{} & \ccol{D48A8A}{} & \ccol{D48A8A}{} & \ccol{D48A8A}{} & \ccol{D48A8A}{} & \ccol{D48A8A}{} & \ccol{D48A8A}{} & \ccol{D48A8A}{} & \ccol{D48A8A}{} & \ccol{D48A8A}{} & \ccol{E8A0A0}{} & \ccol{D48A8A}{} & \ccol{E8A0A0}{} & \ccol{D48A8A}{} & \ccol{D48A8A}{} & \ccol{D48A8A}{} & \ccol{D48A8A}{} & \ccol{D48A8A}{} & \ccol{D48A8A}{} & \ccol{FFE3D3}{} & \ccol{E8A0A0}{} & \ccol{D48A8A}{} & \ccol{D48A8A}{} & \ccol{D48A8A}{} & \ccol{D48A8A}{} & \ccol{D48A8A}{} & \ccol{D48A8A}{} & \ccol{D48A8A}{} & \ccol{D48A8A}{} & \ccol{D48A8A}{} & \ccol{D48A8A}{} & \ccol{D48A8A}{} & \ccol{D48A8A}{} & \ccol{D48A8A}{} & \ccol{D48A8A}{} & \ccol{D48A8A}{} & \ccol{D48A8A}{} & \ccol{D48A8A}{} & \ccol{D48A8A}{} & \ccol{D48A8A}{} & \ccol{D48A8A}{} & \ccol{D48A8A}{} & \ccol{D48A8A}{} & \ccol{D48A8A}{} & \ccol{D48A8A}{} & \ccol{D48A8A}{} & \ccol{D48A8A}{} & \ccol{D48A8A}{} & \ccol{D48A8A}{} \\
\mistral\hspace{1pt} Large 3 & \ccol{D6EEFF}{\cmark} & \ccol{E8A0A0}{} & \ccol{D48A8A}{} & \ccol{E8A0A0}{} & \ccol{D48A8A}{} & \ccol{E8A0A0}{} & \ccol{D48A8A}{} & \ccol{D48A8A}{} & \ccol{D48A8A}{} & \ccol{D48A8A}{} & \ccol{D48A8A}{} & \ccol{F9C6C6}{} & \ccol{F9C6C6}{} & \ccol{FEF2CC}{} & \ccol{D48A8A}{} & \ccol{D48A8A}{} & \ccol{F9C6C6}{} & \ccol{D48A8A}{} & \ccol{D48A8A}{} & \ccol{D48A8A}{} & \ccol{D48A8A}{} & \ccol{D48A8A}{} & \ccol{F9C6C6}{} & \ccol{E8A0A0}{} & \ccol{D48A8A}{} & \ccol{D48A8A}{} & \ccol{D48A8A}{} & \ccol{D48A8A}{} & \ccol{D48A8A}{} & \ccol{D48A8A}{} & \ccol{D48A8A}{} & \ccol{D48A8A}{} & \ccol{D48A8A}{} & \ccol{D48A8A}{} & \ccol{D48A8A}{} & \ccol{F9C6C6}{} & \ccol{D48A8A}{} & \ccol{D48A8A}{} & \ccol{D48A8A}{} & \ccol{D48A8A}{} & \ccol{D48A8A}{} & \ccol{D48A8A}{} & \ccol{D48A8A}{} & \ccol{D48A8A}{} & \ccol{D48A8A}{} & \ccol{D48A8A}{} & \ccol{D48A8A}{} & \ccol{D48A8A}{} & \ccol{D48A8A}{} & \ccol{D48A8A}{} & \ccol{D48A8A}{} & \ccol{D48A8A}{} \\
\gemini\hspace{1pt} 3 Flash & \ccol{D6EEFF}{\cmark} & \ccol{D48A8A}{} & \ccol{E8A0A0}{} & \ccol{D48A8A}{} & \ccol{D48A8A}{} & \ccol{D48A8A}{} & \ccol{D48A8A}{} & \ccol{D48A8A}{} & \ccol{D48A8A}{} & \ccol{D48A8A}{} & \ccol{D48A8A}{} & \ccol{D48A8A}{} & \ccol{FFE3D3}{} & \ccol{E8A0A0}{} & \ccol{D48A8A}{} & \ccol{D48A8A}{} & \ccol{E8A0A0}{} & \ccol{FFE3D3}{} & \ccol{D48A8A}{} & \ccol{D48A8A}{} & \ccol{D48A8A}{} & \ccol{D48A8A}{} & \ccol{FFE3D3}{} & \ccol{E8A0A0}{} & \ccol{FFE3D3}{} & \ccol{F9C6C6}{} & \ccol{D48A8A}{} & \ccol{D48A8A}{} & \ccol{D48A8A}{} & \ccol{D48A8A}{} & \ccol{D48A8A}{} & \ccol{D48A8A}{} & \ccol{D48A8A}{} & \ccol{D48A8A}{} & \ccol{D48A8A}{} & \ccol{D48A8A}{} & \ccol{D48A8A}{} & \ccol{D48A8A}{} & \ccol{D48A8A}{} & \ccol{D48A8A}{} & \ccol{D48A8A}{} & \ccol{D48A8A}{} & \ccol{D48A8A}{} & \ccol{D48A8A}{} & \ccol{D48A8A}{} & \ccol{D48A8A}{} & \ccol{D48A8A}{} & \ccol{D48A8A}{} & \ccol{D48A8A}{} & \ccol{D48A8A}{} & \ccol{D48A8A}{} & \ccol{D48A8A}{} \\
\openai\hspace{1pt} GPT 5 Mini & \ccol{D6EEFF}{\cmark} & \ccol{F9C6C6}{} & \ccol{FFE3D3}{} & \ccol{E8A0A0}{} & \ccol{D48A8A}{} & \ccol{D48A8A}{} & \ccol{D48A8A}{} & \ccol{D48A8A}{} & \ccol{D48A8A}{} & \ccol{D48A8A}{} & \ccol{D48A8A}{} & \ccol{FEF2CC}{} & \ccol{FEF2CC}{} & \ccol{FFE3D3}{} & \ccol{D48A8A}{} & \ccol{E8A0A0}{} & \ccol{D48A8A}{} & \ccol{E8A0A0}{} & \ccol{D48A8A}{} & \ccol{D48A8A}{} & \ccol{E8A0A0}{} & \ccol{D48A8A}{} & \ccol{F9C6C6}{} & \ccol{FEF2CC}{} & \ccol{F9C6C6}{} & \ccol{F9C6C6}{} & \ccol{D48A8A}{} & \ccol{D48A8A}{} & \ccol{D48A8A}{} & \ccol{D48A8A}{} & \ccol{D48A8A}{} & \ccol{D48A8A}{} & \ccol{D48A8A}{} & \ccol{FFE3D3}{} & \ccol{D48A8A}{} & \ccol{D48A8A}{} & \ccol{D48A8A}{} & \ccol{E8A0A0}{} & \ccol{D48A8A}{} & \ccol{D48A8A}{} & \ccol{D48A8A}{} & \ccol{D48A8A}{} & \ccol{D48A8A}{} & \ccol{D48A8A}{} & \ccol{D48A8A}{} & \ccol{F9C6C6}{} & \ccol{D48A8A}{} & \ccol{D48A8A}{} & \ccol{D48A8A}{} & \ccol{D48A8A}{} & \ccol{D48A8A}{} & \ccol{D48A8A}{} \\
\openai\hspace{1pt} GPT 5 Chat & \ccol{D6EEFF}{\cmark} & \ccol{F9C6C6}{} & \ccol{E8A0A0}{} & \ccol{D48A8A}{} & \ccol{E8A0A0}{} & \ccol{E8A0A0}{} & \ccol{D48A8A}{} & \ccol{D48A8A}{} & \ccol{D48A8A}{} & \ccol{D48A8A}{} & \ccol{D48A8A}{} & \ccol{FEF2CC}{} & \ccol{FFE3D3}{} & \ccol{FFE3D3}{} & \ccol{F9C6C6}{} & \ccol{E8A0A0}{} & \ccol{D48A8A}{} & \ccol{D48A8A}{} & \ccol{D48A8A}{} & \ccol{E8A0A0}{} & \ccol{D48A8A}{} & \ccol{D48A8A}{} & \ccol{FFE3D3}{} & \ccol{E8A0A0}{} & \ccol{D48A8A}{} & \ccol{FFE3D3}{} & \ccol{E8A0A0}{} & \ccol{D48A8A}{} & \ccol{D48A8A}{} & \ccol{D48A8A}{} & \ccol{D48A8A}{} & \ccol{D48A8A}{} & \ccol{D48A8A}{} & \ccol{FFE3D3}{} & \ccol{E8A0A0}{} & \ccol{E8A0A0}{} & \ccol{D48A8A}{} & \ccol{D48A8A}{} & \ccol{D48A8A}{} & \ccol{D48A8A}{} & \ccol{D48A8A}{} & \ccol{D48A8A}{} & \ccol{D48A8A}{} & \ccol{D48A8A}{} & \ccol{D48A8A}{} & \ccol{D48A8A}{} & \ccol{D48A8A}{} & \ccol{D48A8A}{} & \ccol{D48A8A}{} & \ccol{D48A8A}{} & \ccol{D48A8A}{} & \ccol{D48A8A}{} \\
\openai\hspace{1pt} o1 & \ccol{FFF9EC}{} & \ccol{F9C6C6}{} & \ccol{E8A0A0}{} & \ccol{E8A0A0}{} & \ccol{F9C6C6}{} & \ccol{E8A0A0}{} & \ccol{D48A8A}{} & \ccol{D48A8A}{} & \ccol{D48A8A}{} & \ccol{D48A8A}{} & \ccol{D48A8A}{} & \ccol{FFE3D3}{} & \ccol{FFE3D3}{} & \ccol{D48A8A}{} & \ccol{E8A0A0}{} & \ccol{FFE3D3}{} & \ccol{FFE3D3}{} & \ccol{D48A8A}{} & \ccol{E8A0A0}{} & \ccol{D48A8A}{} & \ccol{E8A0A0}{} & \ccol{D48A8A}{} & \ccol{F9C6C6}{} & \ccol{F9C6C6}{} & \ccol{D48A8A}{} & \ccol{F9C6C6}{} & \ccol{D48A8A}{} & \ccol{D48A8A}{} & \ccol{D48A8A}{} & \ccol{D48A8A}{} & \ccol{D48A8A}{} & \ccol{D48A8A}{} & \ccol{D48A8A}{} & \ccol{FFE3D3}{} & \ccol{E8A0A0}{} & \ccol{FFE3D3}{} & \ccol{D48A8A}{} & \ccol{E8A0A0}{} & \ccol{D48A8A}{} & \ccol{D48A8A}{} & \ccol{D48A8A}{} & \ccol{D48A8A}{} & \ccol{D48A8A}{} & \ccol{D48A8A}{} & \ccol{D48A8A}{} & \ccol{D48A8A}{} & \ccol{D48A8A}{} & \ccol{D48A8A}{} & \ccol{D48A8A}{} & \ccol{D48A8A}{} & \ccol{D48A8A}{} & \ccol{D48A8A}{} \\
\openai\hspace{1pt} o3 & \ccol{FEF2CC}{} & \ccol{FFE3D3}{} & \ccol{D48A8A}{} & \ccol{D48A8A}{} & \ccol{E8A0A0}{} & \ccol{D48A8A}{} & \ccol{D48A8A}{} & \ccol{D48A8A}{} & \ccol{D48A8A}{} & \ccol{D48A8A}{} & \ccol{D48A8A}{} & \ccol{F9C6C6}{} & \ccol{FEF2CC}{} & \ccol{D48A8A}{} & \ccol{FEF2CC}{} & \ccol{FEF2CC}{} & \ccol{D48A8A}{} & \ccol{D48A8A}{} & \ccol{D48A8A}{} & \ccol{D48A8A}{} & \ccol{E8A0A0}{} & \ccol{D48A8A}{} & \ccol{FFE3D3}{} & \ccol{FFE3D3}{} & \ccol{E8A0A0}{} & \ccol{D48A8A}{} & \ccol{E8A0A0}{} & \ccol{D48A8A}{} & \ccol{D48A8A}{} & \ccol{E8A0A0}{} & \ccol{D48A8A}{} & \ccol{D48A8A}{} & \ccol{D48A8A}{} & \ccol{FFE3D3}{} & \ccol{D48A8A}{} & \ccol{FEF2CC}{} & \ccol{F9C6C6}{} & \ccol{FFE3D3}{} & \ccol{E8A0A0}{} & \ccol{D48A8A}{} & \ccol{D48A8A}{} & \ccol{D48A8A}{} & \ccol{D48A8A}{} & \ccol{D48A8A}{} & \ccol{D48A8A}{} & \ccol{FEF2CC}{} & \ccol{D48A8A}{} & \ccol{D48A8A}{} & \ccol{D48A8A}{} & \ccol{D48A8A}{} & \ccol{D48A8A}{} & \ccol{D48A8A}{} \\
\openai\hspace{1pt} GPT 5 & \ccol{D6EEFF}{\cmark} & \ccol{F9C6C6}{} & \ccol{F9C6C6}{} & \ccol{F9C6C6}{} & \ccol{F9C6C6}{} & \ccol{D48A8A}{} & \ccol{D48A8A}{} & \ccol{D48A8A}{} & \ccol{D48A8A}{} & \ccol{D48A8A}{} & \ccol{D48A8A}{} & \ccol{FFE3D3}{} & \ccol{FFE3D3}{} & \ccol{FFE3D3}{} & \ccol{FEF2CC}{} & \ccol{D48A8A}{} & \ccol{F9C6C6}{} & \ccol{FFE3D3}{} & \ccol{D48A8A}{} & \ccol{D48A8A}{} & \ccol{D48A8A}{} & \ccol{D48A8A}{} & \ccol{F9C6C6}{} & \ccol{FFE3D3}{} & \ccol{E8A0A0}{} & \ccol{E8A0A0}{} & \ccol{D48A8A}{} & \ccol{D48A8A}{} & \ccol{D48A8A}{} & \ccol{E8A0A0}{} & \ccol{D48A8A}{} & \ccol{D48A8A}{} & \ccol{D48A8A}{} & \ccol{D48A8A}{} & \ccol{E8A0A0}{} & \ccol{E8A0A0}{} & \ccol{E8A0A0}{} & \ccol{E8A0A0}{} & \ccol{D48A8A}{} & \ccol{D48A8A}{} & \ccol{D48A8A}{} & \ccol{D48A8A}{} & \ccol{D48A8A}{} & \ccol{D48A8A}{} & \ccol{D48A8A}{} & \ccol{D48A8A}{} & \ccol{D48A8A}{} & \ccol{D48A8A}{} & \ccol{D48A8A}{} & \ccol{D48A8A}{} & \ccol{D48A8A}{} & \ccol{D48A8A}{} \\
\openai\hspace{1pt} GPT 4.1 & \ccol{D6EEFF}{\cmark} & \ccol{E8A0A0}{} & \ccol{FFE3D3}{} & \ccol{D48A8A}{} & \ccol{F9C6C6}{} & \ccol{E8A0A0}{} & \ccol{D48A8A}{} & \ccol{E8A0A0}{} & \ccol{D48A8A}{} & \ccol{D48A8A}{} & \ccol{D48A8A}{} & \ccol{FFF9EC}{} & \ccol{F9C6C6}{} & \ccol{FEF2CC}{} & \ccol{E8A0A0}{} & \ccol{D48A8A}{} & \ccol{D48A8A}{} & \ccol{F9C6C6}{} & \ccol{D48A8A}{} & \ccol{D48A8A}{} & \ccol{D48A8A}{} & \ccol{D48A8A}{} & \ccol{FEF2CC}{} & \ccol{F9C6C6}{} & \ccol{D48A8A}{} & \ccol{E8A0A0}{} & \ccol{E8A0A0}{} & \ccol{D48A8A}{} & \ccol{E8A0A0}{} & \ccol{D48A8A}{} & \ccol{D48A8A}{} & \ccol{D48A8A}{} & \ccol{D48A8A}{} & \ccol{F9C6C6}{} & \ccol{D48A8A}{} & \ccol{E8A0A0}{} & \ccol{D48A8A}{} & \ccol{F9C6C6}{} & \ccol{D48A8A}{} & \ccol{D48A8A}{} & \ccol{D48A8A}{} & \ccol{D48A8A}{} & \ccol{D48A8A}{} & \ccol{D48A8A}{} & \ccol{E8A0A0}{} & \ccol{D48A8A}{} & \ccol{D48A8A}{} & \ccol{D48A8A}{} & \ccol{D48A8A}{} & \ccol{D48A8A}{} & \ccol{D48A8A}{} & \ccol{D48A8A}{} \\
\xai\hspace{1pt} Grok 4 & \ccol{D6EEFF}{\cmark} & \ccol{F9C6C6}{} & \ccol{E8A0A0}{} & \ccol{E8A0A0}{} & \ccol{F9C6C6}{} & \ccol{E8A0A0}{} & \ccol{E8A0A0}{} & \ccol{E8A0A0}{} & \ccol{D48A8A}{} & \ccol{D48A8A}{} & \ccol{D48A8A}{} & \ccol{FFE3D3}{} & \ccol{D48A8A}{} & \ccol{E8A0A0}{} & \ccol{D6EEFF}{\cmark} & \ccol{F9C6C6}{} & \ccol{F9C6C6}{} & \ccol{D48A8A}{} & \ccol{E8A0A0}{} & \ccol{E8A0A0}{} & \ccol{D48A8A}{} & \ccol{F9C6C6}{} & \ccol{FFE3D3}{} & \ccol{F9C6C6}{} & \ccol{FEF2CC}{} & \ccol{F9C6C6}{} & \ccol{F9C6C6}{} & \ccol{E8A0A0}{} & \ccol{E8A0A0}{} & \ccol{D48A8A}{} & \ccol{D48A8A}{} & \ccol{E8A0A0}{} & \ccol{D48A8A}{} & \ccol{D48A8A}{} & \ccol{FFE3D3}{} & \ccol{F9C6C6}{} & \ccol{D48A8A}{} & \ccol{D48A8A}{} & \ccol{F9C6C6}{} & \ccol{D48A8A}{} & \ccol{FEF2CC}{} & \ccol{E8A0A0}{} & \ccol{D48A8A}{} & \ccol{D48A8A}{} & \ccol{FFE3D3}{} & \ccol{D48A8A}{} & \ccol{E8A0A0}{} & \ccol{D48A8A}{} & \ccol{D48A8A}{} & \ccol{D48A8A}{} & \ccol{D48A8A}{} & \ccol{D48A8A}{} \\
\openai\hspace{1pt} GPT 5.1 & \ccol{D6EEFF}{\cmark} & \ccol{FFE3D3}{} & \ccol{FFE3D3}{} & \ccol{F9C6C6}{} & \ccol{F9C6C6}{} & \ccol{F9C6C6}{} & \ccol{F9C6C6}{} & \ccol{D48A8A}{} & \ccol{D48A8A}{} & \ccol{D48A8A}{} & \ccol{D48A8A}{} & \ccol{FFF9EC}{} & \ccol{FEF2CC}{} & \ccol{FFF9EC}{} & \ccol{FEF2CC}{} & \ccol{D48A8A}{} & \ccol{FFE3D3}{} & \ccol{F9C6C6}{} & \ccol{E8A0A0}{} & \ccol{E8A0A0}{} & \ccol{D48A8A}{} & \ccol{D48A8A}{} & \ccol{FFE3D3}{} & \ccol{F9C6C6}{} & \ccol{FEF2CC}{} & \ccol{FFE3D3}{} & \ccol{E8A0A0}{} & \ccol{E8A0A0}{} & \ccol{E8A0A0}{} & \ccol{D48A8A}{} & \ccol{D48A8A}{} & \ccol{D48A8A}{} & \ccol{D48A8A}{} & \ccol{FFE3D3}{} & \ccol{FFE3D3}{} & \ccol{E8A0A0}{} & \ccol{E8A0A0}{} & \ccol{E8A0A0}{} & \ccol{F9C6C6}{} & \ccol{D48A8A}{} & \ccol{E8A0A0}{} & \ccol{D48A8A}{} & \ccol{D48A8A}{} & \ccol{D48A8A}{} & \ccol{E8A0A0}{} & \ccol{E8A0A0}{} & \ccol{E8A0A0}{} & \ccol{D48A8A}{} & \ccol{D48A8A}{} & \ccol{D48A8A}{} & \ccol{D48A8A}{} & \ccol{D48A8A}{} \\
\moonshot\hspace{1pt} Kimi K2.5 & \ccol{D6EEFF}{\cmark} & \ccol{FEF2CC}{} & \ccol{F9C6C6}{} & \ccol{F9C6C6}{} & \ccol{F9C6C6}{} & \ccol{E8A0A0}{} & \ccol{E8A0A0}{} & \ccol{E8A0A0}{} & \ccol{F9C6C6}{} & \ccol{E8A0A0}{} & \ccol{D48A8A}{} & \ccol{FFF9EC}{} & \ccol{E8A0A0}{} & \ccol{FFF9EC}{} & \ccol{FFF9EC}{} & \ccol{FFE3D3}{} & \ccol{FFE3D3}{} & \ccol{FFE3D3}{} & \ccol{F9C6C6}{} & \ccol{D48A8A}{} & \ccol{E8A0A0}{} & \ccol{E8A0A0}{} & \ccol{FEF2CC}{} & \ccol{FFE3D3}{} & \ccol{F9C6C6}{} & \ccol{FEF2CC}{} & \ccol{F9C6C6}{} & \ccol{D48A8A}{} & \ccol{D48A8A}{} & \ccol{D48A8A}{} & \ccol{D48A8A}{} & \ccol{D48A8A}{} & \ccol{D48A8A}{} & \ccol{FFF9EC}{} & \ccol{E8A0A0}{} & \ccol{D48A8A}{} & \ccol{FFE3D3}{} & \ccol{F9C6C6}{} & \ccol{E8A0A0}{} & \ccol{D48A8A}{} & \ccol{E8A0A0}{} & \ccol{F9C6C6}{} & \ccol{D48A8A}{} & \ccol{D48A8A}{} & \ccol{D48A8A}{} & \ccol{D48A8A}{} & \ccol{D48A8A}{} & \ccol{D48A8A}{} & \ccol{D48A8A}{} & \ccol{D48A8A}{} & \ccol{E8A0A0}{} & \ccol{D48A8A}{} \\
\claude\hspace{1pt} 4.6 Sonnet & \ccol{D6EEFF}{\cmark} & \ccol{FFE3D3}{} & \ccol{F9C6C6}{} & \ccol{FEF2CC}{} & \ccol{E8A0A0}{} & \ccol{FFE3D3}{} & \ccol{FEF2CC}{} & \ccol{E8A0A0}{} & \ccol{D48A8A}{} & \ccol{D48A8A}{} & \ccol{D48A8A}{} & \ccol{D6EEFF}{\cmark} & \ccol{FFF9EC}{} & \ccol{FFF9EC}{} & \ccol{FFF9EC}{} & \ccol{FFE3D3}{} & \ccol{FFE3D3}{} & \ccol{FFE3D3}{} & \ccol{D48A8A}{} & \ccol{F9C6C6}{} & \ccol{E8A0A0}{} & \ccol{D48A8A}{} & \ccol{FFE3D3}{} & \ccol{F9C6C6}{} & \ccol{FFF9EC}{} & \ccol{E8A0A0}{} & \ccol{E8A0A0}{} & \ccol{F9C6C6}{} & \ccol{E8A0A0}{} & \ccol{E8A0A0}{} & \ccol{E8A0A0}{} & \ccol{D48A8A}{} & \ccol{D48A8A}{} & \ccol{FFF9EC}{} & \ccol{FFE3D3}{} & \ccol{E8A0A0}{} & \ccol{F9C6C6}{} & \ccol{E8A0A0}{} & \ccol{D48A8A}{} & \ccol{D48A8A}{} & \ccol{F9C6C6}{} & \ccol{F9C6C6}{} & \ccol{E8A0A0}{} & \ccol{D48A8A}{} & \ccol{E8A0A0}{} & \ccol{FFE3D3}{} & \ccol{D48A8A}{} & \ccol{D48A8A}{} & \ccol{D48A8A}{} & \ccol{D48A8A}{} & \ccol{D48A8A}{} & \ccol{D48A8A}{} \\
\openai\hspace{1pt} GPT 5.2 & \ccol{D6EEFF}{\cmark} & \ccol{FFE3D3}{} & \ccol{FFE3D3}{} & \ccol{FFE3D3}{} & \ccol{E8A0A0}{} & \ccol{FFE3D3}{} & \ccol{FFE3D3}{} & \ccol{D48A8A}{} & \ccol{E8A0A0}{} & \ccol{E8A0A0}{} & \ccol{D48A8A}{} & \ccol{D6EEFF}{\cmark} & \ccol{F9C6C6}{} & \ccol{FFF9EC}{} & \ccol{FFE3D3}{} & \ccol{FFE3D3}{} & \ccol{F9C6C6}{} & \ccol{FFE3D3}{} & \ccol{F9C6C6}{} & \ccol{F9C6C6}{} & \ccol{D48A8A}{} & \ccol{E8A0A0}{} & \ccol{FEF2CC}{} & \ccol{FEF2CC}{} & \ccol{FFF9EC}{} & \ccol{E8A0A0}{} & \ccol{E8A0A0}{} & \ccol{F9C6C6}{} & \ccol{E8A0A0}{} & \ccol{D48A8A}{} & \ccol{E8A0A0}{} & \ccol{D48A8A}{} & \ccol{D48A8A}{} & \ccol{FFE3D3}{} & \ccol{F9C6C6}{} & \ccol{FFE3D3}{} & \ccol{F9C6C6}{} & \ccol{F9C6C6}{} & \ccol{FFE3D3}{} & \ccol{D48A8A}{} & \ccol{D48A8A}{} & \ccol{D48A8A}{} & \ccol{D48A8A}{} & \ccol{D48A8A}{} & \ccol{E8A0A0}{} & \ccol{F9C6C6}{} & \ccol{D48A8A}{} & \ccol{D48A8A}{} & \ccol{D48A8A}{} & \ccol{D48A8A}{} & \ccol{D48A8A}{} & \ccol{D48A8A}{} \\
\openai\hspace{1pt} GPT 5.4 & \ccol{D6EEFF}{\cmark} & \ccol{FEF2CC}{} & \ccol{F9C6C6}{} & \ccol{FEF2CC}{} & \ccol{FEF2CC}{} & \ccol{FFE3D3}{} & \ccol{F9C6C6}{} & \ccol{E8A0A0}{} & \ccol{E8A0A0}{} & \ccol{F9C6C6}{} & \ccol{D48A8A}{} & \ccol{FFE3D3}{} & \ccol{FEF2CC}{} & \ccol{FEF2CC}{} & \ccol{F9C6C6}{} & \ccol{FFE3D3}{} & \ccol{F9C6C6}{} & \ccol{FFE3D3}{} & \ccol{F9C6C6}{} & \ccol{F9C6C6}{} & \ccol{E8A0A0}{} & \ccol{F9C6C6}{} & \ccol{FEF2CC}{} & \ccol{FFF9EC}{} & \ccol{FFF9EC}{} & \ccol{F9C6C6}{} & \ccol{E8A0A0}{} & \ccol{F9C6C6}{} & \ccol{F9C6C6}{} & \ccol{E8A0A0}{} & \ccol{E8A0A0}{} & \ccol{E8A0A0}{} & \ccol{D48A8A}{} & \ccol{FFE3D3}{} & \ccol{E8A0A0}{} & \ccol{FFF9EC}{} & \ccol{FFE3D3}{} & \ccol{E8A0A0}{} & \ccol{FEF2CC}{} & \ccol{D48A8A}{} & \ccol{E8A0A0}{} & \ccol{F9C6C6}{} & \ccol{D48A8A}{} & \ccol{D48A8A}{} & \ccol{E8A0A0}{} & \ccol{FEF2CC}{} & \ccol{E8A0A0}{} & \ccol{D48A8A}{} & \ccol{E8A0A0}{} & \ccol{D48A8A}{} & \ccol{D48A8A}{} & \ccol{D48A8A}{} \\
\claude\hspace{1pt} 4.6 Opus & \ccol{D6EEFF}{\cmark} & \ccol{FFF9EC}{} & \ccol{FFE3D3}{} & \ccol{FEF2CC}{} & \ccol{E8A0A0}{} & \ccol{FFE3D3}{} & \ccol{FEF2CC}{} & \ccol{E8A0A0}{} & \ccol{FEF2CC}{} & \ccol{FFE3D3}{} & \ccol{D48A8A}{} & \ccol{D6EEFF}{\cmark} & \ccol{FFF9EC}{} & \ccol{FFF9EC}{} & \ccol{FFF9EC}{} & \ccol{FEF2CC}{} & \ccol{FFE3D3}{} & \ccol{FFE3D3}{} & \ccol{D48A8A}{} & \ccol{FFE3D3}{} & \ccol{E8A0A0}{} & \ccol{E8A0A0}{} & \ccol{FFE3D3}{} & \ccol{FFE3D3}{} & \ccol{D6EEFF}{\cmark} & \ccol{F9C6C6}{} & \ccol{F9C6C6}{} & \ccol{F9C6C6}{} & \ccol{D48A8A}{} & \ccol{F9C6C6}{} & \ccol{F9C6C6}{} & \ccol{E8A0A0}{} & \ccol{D48A8A}{} & \ccol{D6EEFF}{\cmark} & \ccol{FFE3D3}{} & \ccol{F9C6C6}{} & \ccol{E8A0A0}{} & \ccol{D6EEFF}{\cmark} & \ccol{D48A8A}{} & \ccol{E8A0A0}{} & \ccol{F9C6C6}{} & \ccol{E8A0A0}{} & \ccol{E8A0A0}{} & \ccol{D48A8A}{} & \ccol{F9C6C6}{} & \ccol{FFE3D3}{} & \ccol{E8A0A0}{} & \ccol{F9C6C6}{} & \ccol{D48A8A}{} & \ccol{E8A0A0}{} & \ccol{D48A8A}{} & \ccol{D48A8A}{} \\
\gemini\hspace{1pt} 3.1 Pro & \ccol{D6EEFF}{\cmark} & \ccol{D6EEFF}{\cmark} & \ccol{FFE3D3}{} & \ccol{FEF2CC}{} & \ccol{FFF9EC}{} & \ccol{FEF2CC}{} & \ccol{FFF9EC}{} & \ccol{E8A0A0}{} & \ccol{F9C6C6}{} & \ccol{FFE3D3}{} & \ccol{D48A8A}{} & \ccol{D6EEFF}{\cmark} & \ccol{FFF9EC}{} & \ccol{FFF9EC}{} & \ccol{FFF9EC}{} & \ccol{D6EEFF}{\cmark} & \ccol{FFE3D3}{} & \ccol{D48A8A}{} & \ccol{D6EEFF}{\cmark} & \ccol{FFE3D3}{} & \ccol{E8A0A0}{} & \ccol{D6EEFF}{\cmark} & \ccol{FEF2CC}{} & \ccol{FEF2CC}{} & \ccol{FFE3D3}{} & \ccol{FFF9EC}{} & \ccol{E8A0A0}{} & \ccol{F9C6C6}{} & \ccol{F9C6C6}{} & \ccol{FFE3D3}{} & \ccol{FFE3D3}{} & \ccol{F9C6C6}{} & \ccol{E8A0A0}{} & \ccol{D6EEFF}{\cmark} & \ccol{FFF9EC}{} & \ccol{D6EEFF}{\cmark} & \ccol{D6EEFF}{\cmark} & \ccol{D6EEFF}{\cmark} & \ccol{FEF2CC}{} & \ccol{E8A0A0}{} & \ccol{E8A0A0}{} & \ccol{FFE3D3}{} & \ccol{E8A0A0}{} & \ccol{D48A8A}{} & \ccol{D6EEFF}{\cmark} & \ccol{FFE3D3}{} & \ccol{E8A0A0}{} & \ccol{E8A0A0}{} & \ccol{D48A8A}{} & \ccol{E8A0A0}{} & \ccol{D48A8A}{} & \ccol{D48A8A}{} \\
\bottomrule
\end{tabular}
}

%% file: tables/agentic_scores.tex
\centering
\resizebox{\linewidth}{!}{%
\renewcommand{\arraystretch}{1.6}
\begin{tabular}{p{1.5cm}rrrrrrrrrr}
 & \multicolumn{10}{c}{\textbf{Workflow length $k$ (\# interactions)}} \\[-11pt]

& & & & & & & & & & \multicolumn{1}{r@{\kern-1pt}}{$\rightarrow$} \\[-10pt]
\cmidrule[1.0pt]{2-11}
\textbf{Model} & \multicolumn{1}{c}{2} & \multicolumn{1}{c}{4} & \multicolumn{1}{c}{6} & \multicolumn{1}{c}{8} & \multicolumn{1}{c}{10} & \multicolumn{1}{c}{12} & \multicolumn{1}{c}{14} & \multicolumn{1}{c}{16} & \multicolumn{1}{c}{18} & \multicolumn{1}{c}{20} \\
\cmidrule(r){1-1}\cmidrule{2-11}
\openai\hspace{1pt} 5.4 \hfill \icondirect & \ccol{FFFAEF} 94.3 & \ccol{FEF3D2} 89.3 & \ccol{FEEACF} 85.4 & \ccol{FEE0D1} 82.0 & \ccol{FCD7CD} 79.4 & \ccol{FACCC8} 76.4 & \ccol{F9C6C6} 74.6 & \ccol{F6BFBF} 73.1 & \ccol{F3BABA} 72.1 & \ccol{F2B7B7} 71.5 \\
\openai\hspace{1pt} 5.4 \hfill \iconagentic & \ccol{FEF3D2} 89.2 & \ccol{FEECCE} 86.2 & \ccol{FEE0D1} 82.0 & \ccol{FCD5CD} 79.0 & \ccol{F9C9C7} 75.4 & \ccol{F2B8B8} 71.7 & \ccol{EFAFAF} 69.8 & \ccol{EDACAC} 69.1 & \ccol{EBA7A7} 68.2 & \ccol{EBA8A8} 68.3 \\
\addlinespace[5pt]
\openai\hspace{1pt} 5.2 \hfill \icondirect & \ccol{FEF7E6} 92.7 & \ccol{FEEECD} 86.9 & \ccol{FEE1D2} 82.2 & \ccol{FBD1CB} 77.9 & \ccol{F8C5C5} 74.4 & \ccol{F2B8B8} 71.6 & \ccol{EFB0B0} 70.0 & \ccol{ECA9A9} 68.5 & \ccol{E9A3A3} 67.1 & \ccol{E79F9F} 66.1 \\
\openai\hspace{1pt} 5.2 \hfill \iconagentic & \ccol{FEF5DB} 90.7 & \ccol{FEE9CF} 85.1 & \ccol{FBD1CB} 77.9 & \ccol{F8C5C5} 74.5 & \ccol{EEAEAE} 69.4 & \ccol{EAA5A5} 67.5 & \ccol{E69D9D} 65.0 & \ccol{E49C9C} 63.6 & \ccol{E49B9B} 63.2 & \ccol{E49B9B} 63.4 \\
\addlinespace[5pt]
\openai\hspace{1pt} 5.1 \hfill \icondirect & \ccol{FEF5DB} 90.8 & \ccol{FEE3D2} 82.8 & \ccol{FBD2CB} 78.0 & \ccol{F7C3C3} 74.0 & \ccol{EFB0B0} 69.9 & \ccol{E8A0A0} 66.7 & \ccol{E69D9D} 64.9 & \ccol{E39B9B} 62.9 & \ccol{E29999} 61.7 & \ccol{E09797} 60.5 \\
\openai\hspace{1pt} 5.1 \hfill \iconagentic & \ccol{FEE4D2} 83.2 & \ccol{F9C8C7} 75.2 & \ccol{EDABAB} 68.8 & \ccol{E49B9B} 63.3 & \ccol{DF9696} 59.7 & \ccol{DD9494} 58.1 & \ccol{DB9292} 56.1 & \ccol{D98F8F} 54.4 & \ccol{D78D8D} 53.0 & \ccol{D68C8C} 52.1 \\
\addlinespace[5pt]
\openai\hspace{1pt} 4.1 \hfill \icondirect & \ccol{FEF3D0} 88.9 & \ccol{FCD8CE} 79.8 & \ccol{F1B5B5} 70.9 & \ccol{EAA6A6} 67.7 & \ccol{E29A9A} 62.2 & \ccol{DC9393} 56.8 & \ccol{D99090} 54.8 & \ccol{D58C8C} 51.7 & \ccol{D48A8A} 49.8 & \ccol{D48A8A} 49.5 \\
\openai\hspace{1pt} 4.1 \hfill \iconagentic & \ccol{FEE7D0} 84.4 & \ccol{F3B9B9} 71.9 & \ccol{E49B9B} 63.5 & \ccol{DC9393} 57.1 & \ccol{D78D8D} 52.8 & \ccol{D48A8A} 48.8 & \ccol{D48A8A} 46.1 & \ccol{D48A8A} 43.6 & \ccol{D48A8A} 42.0 & \ccol{D48A8A} 40.4 \\
\bottomrule
\end{tabular}
}

%% file: tables/agentic_behavior.tex
\centering
\resizebox{\linewidth}{!}{%
\setlength{\tabcolsep}{4pt}
\renewcommand{\arraystretch}{1.6}
\begin{tabular}{lrrrrrrrrrr}
 & \multicolumn{5}{c}{\textbf{Overhead (with tools / no tools)}} & \multicolumn{4}{c}{\textbf{Edit Method}} &  \\
\cmidrule(lr){2-6}\cmidrule(lr){7-10}
\textbf{Model} & \multicolumn{1}{c}{\#Tools} & \multicolumn{1}{c}{Inp$\times$} & \multicolumn{1}{c}{Out$\times$} & \multicolumn{1}{c}{\$$\times$} & \multicolumn{1}{c}{Lat$\times$} & \multicolumn{1}{c}{\iconexec} & \multicolumn{1}{c}{\iconwrite} & \multicolumn{1}{c}{\iconexec\;+\;\iconwrite} & \multicolumn{1}{c}{$\varnothing$} & \multicolumn{1}{c}{\%Dist} \\
\cmidrule(r){1-1}\cmidrule(lr){2-6}\cmidrule(lr){7-10}\cmidrule(l){11-11}
\openai\hspace{1pt} 5.4 & 8.3 & \ccol{F2C3C3} 2.1$\times$ & \ccol{D8E7F2} 0.6$\times$ & \ccol{FDFDFE} 1.0$\times$ & \ccol{A7C8E0} 0.4$\times$ & 45\% & 49\% & 6\% & 0\% & 16\% \\
\openai\hspace{1pt} 5.2 & 11.7 & \ccol{E4A7A7} 3.2$\times$ & \ccol{E8F1F7} 0.8$\times$ & \ccol{FAE6E6} 1.4$\times$ & \ccol{FCF2F2} 1.2$\times$ & 39\% & 47\% & 14\% & 0\% & 18\% \\
\openai\hspace{1pt} 5.1 & 7.5 & \ccol{F3C7C7} 2.0$\times$ & \ccol{F2F7FA} 0.9$\times$ & \ccol{FDF5F5} 1.1$\times$ & \ccol{E8ADAD} 2.9$\times$ & 15\% & 81\% & 4\% & 0\% & 22\% \\
\openai\hspace{1pt} 4.1 & 9.6 & \ccol{D78F8F} 4.6$\times$ & \ccol{FBEDED} 1.2$\times$ & \ccol{F1BEBE} 2.2$\times$ & \ccol{FAE8E8} 1.3$\times$ & 10\% & 75\% & 14\% & 1\% & 7\% \\
\bottomrule
\end{tabular}
}

%% file: tables/context_size.tex
\centering
\resizebox{\linewidth}{!}{%
\renewcommand{\arraystretch}{1.6}
\begin{tabular}{c@{\kern5pt}lrrrrrrrrrr}
 & & \multicolumn{10}{c}{\textbf{Workflow length $k$ (\# interactions)}} \\[-11pt]

& & & & & & & & & & & \multicolumn{1}{r@{\kern-1pt}}{$\rightarrow$} \\[-10pt]
\cmidrule[1.0pt]{3-12}
 & \textbf{Doc.\ Size} & \multicolumn{1}{c}{2} & \multicolumn{1}{c}{4} & \multicolumn{1}{c}{6} & \multicolumn{1}{c}{8} & \multicolumn{1}{c}{10} & \multicolumn{1}{c}{12} & \multicolumn{1}{c}{14} & \multicolumn{1}{c}{16} & \multicolumn{1}{c}{18} & \multicolumn{1}{c}{20} \\
\cmidrule(r){2-2}\cmidrule{3-12}
\makebox[0pt]{$\uparrow$} & \ctxdot{1.2pt} 1k tokens & \ccol{FFFEFE} 97.3 & \ccol{FFFDFB} 96.4 & \ccol{FFFDFA} 96.1 & \ccol{FFFDFA} 96.2 & \ccol{FFFDFA} 95.9 & \ccol{FFFDF9} 95.4 & \ccol{FFFDF9} 95.4 & \ccol{FFFAF1} 91.8 & \ccol{FFFAF1} 91.8 & \ccol{FFFAEF} 91.4 \\
 & \ctxdot{1.6pt} 2k tokens & \ccol{FFFFFF} 98.4 & \ccol{FFFEFC} 96.7 & \ccol{FFFDF9} 95.3 & \ccol{FFFBF2} 92.6 & \ccol{FFFAF1} 91.9 & \ccol{FFFAEF} 91.3 & \ccol{FFF9EE} 90.9 & \ccol{FFF9ED} 90.7 & \ccol{FFF9EC} 90.0 & \ccol{FFF9EC} 89.9 \\
\makebox[0pt]{\smash{\rule[-50pt]{0.6pt}{90pt}}}\makebox[0pt]{\smash{\raisebox{2pt}{\rule{6pt}{0.6pt}}}} & \ctxdot{2.2pt} 4k tokens & \ccol{FFFCF5} 94.1 & \ccol{FFFAF1} 92.1 & \ccol{FFFAEF} 91.4 & \ccol{FEF4D9} 84.0 & \ccol{FEF4D6} 83.1 & \ccol{FEF3D3} 82.4 & \ccol{FEF2CF} 81.0 & \ccol{FEF2CC} 79.9 & \ccol{FEF1CC} 79.1 & \ccol{FEF0CC} 79.0 \\
 & \ctxdot{2.8pt} 6k tokens & \ccol{FFFFFF} 98.7 & \ccol{FFFBF4} 93.4 & \ccol{FFF9ED} 90.5 & \ccol{FEF7E3} 87.2 & \ccol{FEF4D9} 84.0 & \ccol{FEF2CE} 80.7 & \ccol{FEF1CC} 79.3 & \ccol{FEF2CC} 79.9 & \ccol{FEEECD} 77.7 & \ccol{FEE6D1} 72.3 \\
 & \ctxdot{3.4pt} 8k tokens & \ccol{FFFCF5} 94.1 & \ccol{FFFAF1} 92.1 & \ccol{FEF4D8} 83.9 & \ccol{FEF1CC} 79.7 & \ccol{FEE9D0} 74.2 & \ccol{FEE7D1} 72.9 & \ccol{FEE7D1} 72.6 & \ccol{FEE1D2} 69.0 & \ccol{FDDDD0} 67.0 & \ccol{FDDDD0} 67.4 \\
\makebox[0pt]{$\downarrow$} & \ctxdot{4.0pt} 10k tokens & \ccol{FFF9EC} 90.1 & \ccol{FEF6E1} 86.4 & \ccol{FEF4D6} 83.0 & \ccol{FEEECD} 77.7 & \ccol{FEE6D1} 72.2 & \ccol{FEE4D2} 70.7 & \ccol{FDDDD0} 67.1 & \ccol{FCD6CD} 63.6 & \ccol{FCD7CD} 63.7 & \ccol{FBCFCA} 59.9 \\
\bottomrule
\end{tabular}
}

%% file: tables/fifty_rounds.tex
\centering
\resizebox{\linewidth}{!}{%
\renewcommand{\arraystretch}{1.6}
\begin{tabular}{lrr/rrrrrrrr}
 & \multicolumn{10}{c}{\textbf{Workflow length $k$ (\# interactions)}} \\[-11pt]

& & & \llap{\smash{\rule[-1.5pt]{2pt}{7pt}}\kern28pt} & & & & & & & \multicolumn{1}{r@{\kern-1pt}}{$\rightarrow$} \\[-10pt]
\cmidrule[1.0pt]{2-3}\cmidrule[1.0pt]{4-11}
\textbf{Model} & \multicolumn{1}{c}{10} & \multicolumn{1}{c}{20} & \multicolumn{1}{c}{30} & \multicolumn{1}{c}{40} & \multicolumn{1}{c}{50} & \multicolumn{1}{c}{60} & \multicolumn{1}{c}{70} & \multicolumn{1}{c}{80} & \multicolumn{1}{c}{90} & \multicolumn{1}{c}{100} \\
\cmidrule(r){1-1}\cmidrule(r){2-3}\cmidrule{4-11}
\openai\hspace{1pt} GPT 5.4 & \ccol{FEF1CC} 79.7 & \ccol{FEE7D1} 72.9 & \ccol{FEE2D2} 69.7 & \ccol{FDDDD0} 66.8 & \ccol{FDDBCF} 66.2 & \ccol{FCD5CC} 62.9 & \ccol{FBD3CC} 62.2 & \ccol{FBD3CC} 62.0 & \ccol{FBD1CA} 60.6 & \ccol{FACDC9} 58.7 \\
\openai\hspace{1pt} GPT 5.2 & \ccol{FEEBCF} 75.6 & \ccol{FDDDD0} 67.1 & \ccol{FCD6CD} 63.5 & \ccol{FBCFCA} 60.0 & \ccol{F9C9C7} 56.5 & \ccol{F9C9C7} 56.6 & \ccol{F6C1C1} 53.1 & \ccol{F3BABA} 50.6 & \ccol{F4BCBC} 51.5 & \ccol{F3BABA} 50.4 \\
\openai\hspace{1pt} GPT 5.1 & \ccol{FEE0D1} 68.5 & \ccol{FBD1CB} 60.9 & \ccol{F9C9C7} 56.7 & \ccol{F6BFBF} 52.4 & \ccol{F2B8B8} 49.5 & \ccol{EFB1B1} 47.0 & \ccol{EDADAD} 45.3 & \ccol{EDADAD} 45.2 & \ccol{EBA8A8} 43.4 & \ccol{EAA6A6} 42.6 \\
\openai\hspace{1pt} GPT 4.1 & \ccol{FACCC8} 58.4 & \ccol{F2B7B7} 49.3 & \ccol{EDABAB} 44.4 & \ccol{E9A3A3} 41.5 & \ccol{E79F9F} 39.3 & \ccol{E69E9E} 38.2 & \ccol{E69D9D} 37.2 & \ccol{E59C9C} 35.6 & \ccol{E49B9B} 34.2 & \ccol{E39A9A} 33.3 \\
\bottomrule
\end{tabular}
}

%% file: tables/distractor_effect.tex
\centering
\resizebox{\linewidth}{!}{%
\renewcommand{\arraystretch}{1.6}
\begin{tabular}{lrrrrrrrrrr}
 & \multicolumn{10}{c}{\textbf{Workflow length $k$ (\# interactions)}} \\[-11pt]

& & & & & & & & & & \multicolumn{1}{r@{\kern-1pt}}{$\rightarrow$} \\[-10pt]
\cmidrule[1.0pt]{2-11}
\textbf{Model} & \multicolumn{1}{c}{2} & \multicolumn{1}{c}{4} & \multicolumn{1}{c}{6} & \multicolumn{1}{c}{8} & \multicolumn{1}{c}{10} & \multicolumn{1}{c}{12} & \multicolumn{1}{c}{14} & \multicolumn{1}{c}{16} & \multicolumn{1}{c}{18} & \multicolumn{1}{c}{20} \\
\cmidrule(r){1-1}\cmidrule{2-11}
\openai\hspace{1pt} GPT 5.4 & \ccol{FFFCF6} 94.3 & \ccol{FEF8E9} 89.3 & \ccol{FEF5DD} 85.4 & \ccol{FEF3D2} 82.0 & \ccol{FEF1CC} 79.4 & \ccol{FEECCE} 76.4 & \ccol{FEE9CF} 74.6 & \ccol{FEE7D0} 73.1 & \ccol{FEE6D1} 72.1 & \ccol{FEE5D2} 71.5 \\
\quad{\footnotesize\textcolor{gray}{$-$ distractor}} & \ccol{FFFCF7} 94.7 & \ccol{FFF9EE} 90.9 & \ccol{FEF7E6} 88.4 & \ccol{FEF6E1} 86.4 & \ccol{FEF4D7} 83.5 & \ccol{FEF3D2} 82.1 & \ccol{FEF2CF} 81.0 & \ccol{FEF1CC} 79.8 & \ccol{FEF1CC} 79.2 & \ccol{FEEECD} 77.8 \\
\addlinespace[5pt]
\openai\hspace{1pt} GPT 5.2 & \ccol{FFFBF2} 92.7 & \ccol{FEF6E2} 86.9 & \ccol{FEF3D3} 82.2 & \ccol{FEEECD} 77.9 & \ccol{FEE9CF} 74.4 & \ccol{FEE5D1} 71.6 & \ccol{FEE3D2} 70.0 & \ccol{FEE0D1} 68.5 & \ccol{FDDDD0} 67.1 & \ccol{FDDBCF} 66.1 \\
\quad{\footnotesize\textcolor{gray}{$-$ distractor}} & \ccol{FFFBF4} 93.4 & \ccol{FEF8E8} 88.8 & \ccol{FEF6DE} 85.8 & \ccol{FEF4D6} 83.0 & \ccol{FEF2CF} 80.9 & \ccol{FEF0CC} 78.5 & \ccol{FEECCE} 76.0 & \ccol{FEEACF} 75.1 & \ccol{FEEACF} 74.7 & \ccol{FEE9CF} 74.5 \\
\addlinespace[5pt]
\openai\hspace{1pt} GPT 5.1 & \ccol{FFF9EE} 90.8 & \ccol{FEF3D5} 82.8 & \ccol{FEEECD} 78.0 & \ccol{FEE9D0} 74.0 & \ccol{FEE3D2} 69.9 & \ccol{FDDCD0} 66.7 & \ccol{FCD9CE} 64.9 & \ccol{FCD5CD} 62.9 & \ccol{FBD3CB} 61.7 & \ccol{FBD0CA} 60.5 \\
\quad{\footnotesize\textcolor{gray}{$-$ distractor}} & \ccol{FFFCF5} 94.1 & \ccol{FEF7E5} 87.8 & \ccol{FEF4D9} 84.1 & \ccol{FEF1CC} 79.8 & \ccol{FEECCE} 76.3 & \ccol{FEE6D1} 72.4 & \ccol{FEE3D2} 70.4 & \ccol{FEE2D2} 69.8 & \ccol{FEDFD1} 67.8 & \ccol{FDDDD0} 67.0 \\
\addlinespace[5pt]
\openai\hspace{1pt} GPT 4.1 & \ccol{FEF8E8} 88.9 & \ccol{FEF1CC} 79.8 & \ccol{FEE4D2} 70.9 & \ccol{FEDED0} 67.7 & \ccol{FBD3CC} 62.2 & \ccol{F9C9C7} 56.8 & \ccol{F8C5C5} 54.8 & \ccol{F5BDBD} 51.7 & \ccol{F2B8B8} 49.8 & \ccol{F2B7B7} 49.5 \\
\quad{\footnotesize\textcolor{gray}{$-$ distractor}} & \ccol{FEF7E6} 88.1 & \ccol{FEF1CC} 79.2 & \ccol{FEE8D0} 73.6 & \ccol{FEDED0} 67.6 & \ccol{FCD8CE} 64.4 & \ccol{FBD0CA} 60.4 & \ccol{F9CAC7} 57.4 & \ccol{F9C8C7} 56.1 & \ccol{F8C5C5} 54.7 & \ccol{F6BFBF} 52.3 \\
\bottomrule
\end{tabular}
}

%% file: tables/image_domain.tex
\centering
\resizebox{\linewidth}{!}{%
\renewcommand{\arraystretch}{1.6}
\begin{tabular}{lrrrrrrrrrr}
 & \multicolumn{10}{c}{\textbf{Workflow length $k$ (\# interactions)}} \\[-11pt]

& & & & & & & & & & \multicolumn{1}{r@{\kern-1pt}}{$\rightarrow$} \\[-10pt]
\cmidrule[1.0pt]{2-11}
\textbf{Model} & \multicolumn{1}{c}{2} & \multicolumn{1}{c}{4} & \multicolumn{1}{c}{6} & \multicolumn{1}{c}{8} & \multicolumn{1}{c}{10} & \multicolumn{1}{c}{12} & \multicolumn{1}{c}{14} & \multicolumn{1}{c}{16} & \multicolumn{1}{c}{18} & \multicolumn{1}{c}{20} \\
\cmidrule(r){1-1}\cmidrule{2-11}
\berkeley\hspace{1pt} Instruct Pix2pix & \ccol{E49B9B} 34.3 & \ccol{DD9494} 24.3 & \ccol{DB9292} 21.1 & \ccol{D78E8E} 15.9 & \ccol{D98F8F} 17.7 & \ccol{D78E8E} 16.1 & \ccol{D48A8A} 11.2 & \ccol{D48A8A} 9.6 & \ccol{D58B8B} 12.1 & \ccol{D48B8B} 11.4 \\
\bfl\hspace{1pt} Flux2 Dev & \ccol{E29999} 31.3 & \ccol{DE9595} 25.4 & \ccol{DC9393} 23.4 & \ccol{DA9191} 20.6 & \ccol{DA9191} 19.7 & \ccol{D99090} 19.2 & \ccol{DE9595} 25.7 & \ccol{DC9393} 23.1 & \ccol{DC9393} 23.2 & \ccol{DA9090} 19.5 \\
\openai\hspace{1pt} GPT Image 1 & \ccol{E19898} 30.3 & \ccol{DF9696} 27.6 & \ccol{DB9191} 20.9 & \ccol{DC9292} 22.0 & \ccol{DA9090} 19.6 & \ccol{D99090} 19.1 & \ccol{DA9090} 19.5 & \ccol{DB9191} 20.8 & \ccol{DC9393} 22.5 & \ccol{DB9191} 20.9 \\
\bfl\hspace{1pt} Flux Kontext & \ccol{F8C3C3} 54.0 & \ccol{E59C9C} 35.5 & \ccol{E49C9C} 35.0 & \ccol{DF9696} 27.2 & \ccol{DF9696} 26.6 & \ccol{DD9494} 25.1 & \ccol{DC9393} 22.6 & \ccol{DC9292} 22.1 & \ccol{DB9292} 21.0 & \ccol{DB9292} 22.0 \\
\bfl\hspace{1pt} Flux2 Klein 4b & \ccol{E79F9F} 38.8 & \ccol{E29999} 31.3 & \ccol{E09797} 28.7 & \ccol{DF9696} 26.9 & \ccol{DD9494} 24.4 & \ccol{DC9393} 23.1 & \ccol{DC9393} 22.5 & \ccol{DD9494} 24.0 & \ccol{DB9292} 21.8 & \ccol{DC9393} 22.4 \\
\bfl\hspace{1pt} Flux2 Klein 9b & \ccol{F8C4C4} 54.6 & \ccol{E59D9D} 37.0 & \ccol{E29999} 31.6 & \ccol{E09797} 29.0 & \ccol{DE9595} 25.8 & \ccol{DC9393} 23.7 & \ccol{DE9595} 26.2 & \ccol{DD9494} 25.1 & \ccol{DF9696} 27.2 & \ccol{DE9595} 26.2 \\
\gemini\hspace{1pt} 2.5 Flash Image & \ccol{F9C6C6} 55.1 & \ccol{ECA9A9} 43.8 & \ccol{E8A0A0} 40.0 & \ccol{E49C9C} 35.1 & \ccol{E49C9C} 34.8 & \ccol{E29A9A} 32.7 & \ccol{E09898} 29.5 & \ccol{E09797} 28.9 & \ccol{E09797} 29.0 & \ccol{DF9696} 27.7 \\
\gemini\hspace{1pt} 3 Pro Image & \ccol{FCD5CD} 63.2 & \ccol{F1B4B4} 48.1 & \ccol{E59D9D} 36.7 & \ccol{E09797} 28.5 & \ccol{DF9797} 27.9 & \ccol{DF9696} 27.6 & \ccol{E09797} 29.1 & \ccol{E09797} 28.7 & \ccol{E09797} 28.7 & \ccol{E19898} 30.0 \\
\gemini\hspace{1pt} 3.1 Flash Image & \ccol{F9C7C6} 55.6 & \ccol{EDABAB} 44.4 & \ccol{E59C9C} 35.5 & \ccol{E29A9A} 32.7 & \ccol{E29A9A} 32.3 & \ccol{E29999} 32.0 & \ccol{E19898} 30.4 & \ccol{E29999} 31.5 & \ccol{E39A9A} 33.2 & \ccol{E19898} 30.4 \\
\bottomrule
\end{tabular}
}

%% file: tables/critical_errors_compact.tex
\centering
\resizebox{\linewidth}{!}{%
\renewcommand{\arraystretch}{1.6}
\begin{tabular}{l/rrrrr/r}
 & \multicolumn{5}{c}{\textbf{\% Relays with 1+}} & \\[-2pt]
 & \multicolumn{5}{c}{\textbf{Critical Error by Interaction}} & \\[-11pt]
 & & & & & \multicolumn{1}{r@{\kern-1pt}}{$\rightarrow$} & \\[-10pt]
\cmidrule(lr){2-6}\cmidrule(l){7-7}
\textbf{Model} & \multicolumn{1}{c}{2} & \multicolumn{1}{c}{6} & \multicolumn{1}{c}{10} & \multicolumn{1}{c}{14} & \multicolumn{1}{c}{20} & \% Critical \\
\cmidrule(r){1-1}\cmidrule(lr){2-6}\cmidrule(l){7-7}
\gemini\hspace{1pt} 3.1 Pro & \ccol{FFFAF1} 6.5 & \ccol{FEF4D7} 16.4 & \ccol{FEE9CF} 26.6 & \ccol{FEDFD1} 33.9 & \ccol{FCD8CE} 38.1 & \ccol{DB9292} 86.3 \\
\claude\hspace{1pt} 4.6 Opus & \ccol{FEF6DE} 13.7 & \ccol{FEE9D0} 27.0 & \ccol{FBD4CC} 40.4 & \ccol{F9CAC7} 46.4 & \ccol{F8C3C3} 49.7 & \ccol{DC9292} 86.1 \\
\claude\hspace{1pt} 4.6 Sonnet & \ccol{FEF4D6} 17.0 & \ccol{FEE0D1} 33.4 & \ccol{FBD0CA} 42.7 & \ccol{F8C5C5} 48.8 & \ccol{F4BCBC} 53.2 & \ccol{DB9292} 86.4 \\
\openai\hspace{1pt} GPT 5.4 & \ccol{FEF6DE} 13.9 & \ccol{FEE5D2} 30.1 & \ccol{FBD1CA} 42.5 & \ccol{F9C7C6} 48.2 & \ccol{F2B7B7} 55.2 & \ccol{DF9696} 80.9 \\
\openai\hspace{1pt} GPT 5.2 & \ccol{FEF4D9} 15.8 & \ccol{FDDCD0} 35.7 & \ccol{F9C7C6} 48.3 & \ccol{F3B9B9} 54.6 & \ccol{EDABAB} 60.7 & \ccol{DD9494} 83.4 \\
\xai\hspace{1pt} Grok 4 & \ccol{FEF5DB} 14.9 & \ccol{FEDFD1} 34.2 & \ccol{F9C9C7} 46.6 & \ccol{F4BBBB} 53.8 & \ccol{ECAAAA} 61.1 & \ccol{D88F8F} 92.0 \\
\moonshot\hspace{1pt} Kimi K2.5 & \ccol{FEF3D1} 18.5 & \ccol{FEDFD1} 33.7 & \ccol{F9C7C6} 48.3 & \ccol{F2B7B7} 55.1 & \ccol{ECA9A9} 61.3 & \ccol{DB9292} 87.2 \\
\openai\hspace{1pt} GPT 5.1 & \ccol{FEF0CC} 21.9 & \ccol{FACDC9} 44.9 & \ccol{EFB1B1} 58.1 & \ccol{E9A3A3} 64.4 & \ccol{E69D9D} 68.8 & \ccol{DD9494} 84.1 \\
\openai\hspace{1pt} GPT 5 & \ccol{FEF5DB} 14.9 & \ccol{FBD1CB} 42.1 & \ccol{F0B3B3} 57.4 & \ccol{E9A2A2} 64.5 & \ccol{E49C9C} 71.5 & \ccol{D88E8E} 92.9 \\
\openai\hspace{1pt} GPT 5 Mini & \ccol{FEEDCD} 23.5 & \ccol{F5BEBE} 52.3 & \ccol{E79F9F} 66.8 & \ccol{E29A9A} 74.5 & \ccol{E19898} 77.1 & \ccol{D98F8F} 90.8 \\
\openai\hspace{1pt} GPT 5 Chat & \ccol{FEE4D2} 31.1 & \ccol{EFB1B1} 58.1 & \ccol{E69D9D} 69.0 & \ccol{E29A9A} 74.4 & \ccol{E09898} 77.8 & \ccol{DA9090} 89.2 \\
\openai\hspace{1pt} o1 & \ccol{FEE7D1} 28.7 & \ccol{F2B7B7} 55.2 & \ccol{E69E9E} 68.5 & \ccol{E49B9B} 72.0 & \ccol{E09898} 78.0 & \ccol{D99090} 90.2 \\
\openai\hspace{1pt} o3 & \ccol{FEE3D2} 31.4 & \ccol{EFB0B0} 58.6 & \ccol{E59D9D} 69.3 & \ccol{E29999} 75.9 & \ccol{E09797} 79.1 & \ccol{D88F8F} 91.5 \\
\openai\hspace{1pt} GPT 4.1 & \ccol{FEEBCE} 25.3 & \ccol{F3B9B9} 54.5 & \ccol{E59D9D} 69.2 & \ccol{E29999} 75.3 & \ccol{DF9797} 79.5 & \ccol{DB9292} 86.6 \\
\gemini\hspace{1pt} 3 Flash & \ccol{FEE4D2} 30.5 & \ccol{F0B3B3} 57.2 & \ccol{E69E9E} 68.6 & \ccol{E39B9B} 73.4 & \ccol{DF9696} 80.6 & \ccol{D68C8C} 95.7 \\
\mistral\hspace{1pt} Large 3 & \ccol{FEDFD1} 34.2 & \ccol{EAA5A5} 63.3 & \ccol{E19898} 77.1 & \ccol{DF9696} 81.2 & \ccol{DC9393} 85.8 & \ccol{D99090} 90.5 \\
\openai\hspace{1pt} OSS 120B & \ccol{F9C9C7} 47.1 & \ccol{DF9696} 80.1 & \ccol{D99090} 90.0 & \ccol{D88F8F} 91.6 & \ccol{D88E8E} 92.9 & \ccol{D68D8D} 95.0 \\
\openai\hspace{1pt} GPT 4o & \ccol{E19898} 77.2 & \ccol{D88F8F} 92.1 & \ccol{D68C8C} 95.4 & \ccol{D68C8C} 96.1 & \ccol{D68C8C} 96.4 & \ccol{D68C8C} 95.7 \\
\openai\hspace{1pt} GPT 5 Nano & \ccol{DE9595} 81.8 & \ccol{D68C8C} 96.0 & \ccol{D58B8B} 96.9 & \ccol{D58B8B} 97.0 & \ccol{D58B8B} 97.2 & \ccol{D58B8B} 97.0 \\
\bottomrule
\end{tabular}
}

%% file: tables/round_robin.tex
\centering
\resizebox{\linewidth}{!}{%
\renewcommand{\arraystretch}{1.6}
\begin{tabular}{p{1.5cm}rrrrrrrrrr}
 & \multicolumn{10}{c}{\textbf{Workflow length $k$ (\# interactions)}} \\[-11pt]

& & & & & & & & & & \multicolumn{1}{r@{\kern-1pt}}{$\rightarrow$} \\[-10pt]
\cmidrule[1.0pt]{2-11}
\textbf{Model} & \multicolumn{1}{c}{2} & \multicolumn{1}{c}{4} & \multicolumn{1}{c}{6} & \multicolumn{1}{c}{8} & \multicolumn{1}{c}{10} & \multicolumn{1}{c}{12} & \multicolumn{1}{c}{14} & \multicolumn{1}{c}{16} & \multicolumn{1}{c}{18} & \multicolumn{1}{c}{20} \\
\cmidrule(r){1-1}\cmidrule{2-11}
\openai\hspace{1pt} 5.4 \hfill \iconroundrobin & \ccol{FFFCF7} 94.7 & \ccol{FFFAF0} 91.5 & \ccol{FEF5DD} 85.5 & \ccol{FEF4D7} 83.5 & \ccol{FEF3D0} 81.4 & \ccol{FEE5D1} 71.8 & \ccol{FEE6D1} 71.9 & \ccol{FDDDD0} 67.3 & \ccol{FDD9CE} 65.1 & \ccol{FDDDD0} 66.9 \\
\openai\hspace{1pt} 5.4 \hfill \iconsingleedit & \ccol{FFFCF5} 94.0 & \ccol{FFFAF1} 92.1 & \ccol{FFFAEF} 91.3 & \ccol{FFF9EC} 90.4 & \ccol{FEF8E9} 89.3 & \ccol{FEF8E8} 88.7 & \ccol{FEF8E8} 88.8 & \ccol{FEF8E7} 88.5 & \ccol{FEF7E4} 87.7 & \ccol{FEF7E3} 87.1 \\
\addlinespace[5pt]
\openai\hspace{1pt} 5.2 \hfill \iconroundrobin & \ccol{FFFCF5} 94.1 & \ccol{FEF8E8} 88.7 & \ccol{FEF4D6} 83.2 & \ccol{FEEFCD} 78.3 & \ccol{FEEACF} 74.9 & \ccol{FEE5D2} 71.4 & \ccol{FEE3D2} 69.8 & \ccol{FEE0D1} 68.5 & \ccol{FDDCD0} 66.7 & \ccol{FDDBCF} 66.0 \\
\openai\hspace{1pt} 5.2 \hfill \iconsingleedit & \ccol{FFFBF2} 92.5 & \ccol{FEF8EB} 89.7 & \ccol{FEF8E8} 89.0 & \ccol{FEF6E2} 86.8 & \ccol{FEF7E3} 87.0 & \ccol{FEF6E1} 86.6 & \ccol{FEF6E0} 86.0 & \ccol{FEF5DC} 85.3 & \ccol{FEF5DB} 84.7 & \ccol{FEF4D9} 84.0 \\
\addlinespace[5pt]
\openai\hspace{1pt} 5.1 \hfill \iconroundrobin & \ccol{FEF8E8} 89.0 & \ccol{FEF0CC} 78.7 & \ccol{FEE6D1} 72.4 & \ccol{FEE3D2} 70.0 & \ccol{FDDBCF} 66.1 & \ccol{FBD4CC} 62.3 & \ccol{FBD3CC} 62.0 & \ccol{FBD0CA} 60.5 & \ccol{FACDC9} 59.1 & \ccol{F9C9C7} 56.8 \\
\openai\hspace{1pt} 5.1 \hfill \iconsingleedit & \ccol{FEF8E9} 89.4 & \ccol{FEF5DD} 85.6 & \ccol{FEF4D8} 83.8 & \ccol{FEF3D5} 82.6 & \ccol{FEF3D3} 82.4 & \ccol{FEF2CE} 80.7 & \ccol{FEF1CC} 79.3 & \ccol{FEEECD} 77.9 & \ccol{FEEECD} 77.9 & \ccol{FEEECD} 77.4 \\
\addlinespace[5pt]
\openai\hspace{1pt} 4.1 \hfill \iconroundrobin & \ccol{FEF6E0} 86.2 & \ccol{FEF2CD} 80.3 & \ccol{FEE1D2} 69.1 & \ccol{FEDFD1} 68.2 & \ccol{FDDCD0} 66.7 & \ccol{FBD1CB} 61.0 & \ccol{FACFCA} 59.7 & \ccol{F7C2C2} 53.4 & \ccol{F2B6B6} 49.0 & \ccol{F4BCBC} 51.4 \\
\openai\hspace{1pt} 4.1 \hfill \iconsingleedit & \ccol{FEF7E6} 88.1 & \ccol{FEF5DB} 84.8 & \ccol{FEF4D6} 83.0 & \ccol{FEF2CE} 80.8 & \ccol{FEF1CC} 79.3 & \ccol{FEEECD} 77.6 & \ccol{FEEDCD} 77.2 & \ccol{FEECCE} 76.7 & \ccol{FEEACF} 75.2 & \ccol{FEE8D0} 73.9 \\
\bottomrule
\end{tabular}
}

%% file: tables/critical_errors.tex
\centering
\resizebox{\linewidth}{!}{%
\renewcommand{\arraystretch}{1.6}
\begin{tabular}{l/rrrrrrrrrr/rrr}
 & \multicolumn{10}{c}{\textbf{\% Runs with 1+ Critical Error by Interaction}} & \multicolumn{3}{c}{\textbf{Degradation Breakdown}} \\[-11pt]
 & & & & & & & & & & \multicolumn{1}{r@{\kern-1pt}}{$\rightarrow$} & & & \\[-10pt]
\cmidrule(lr){2-11}\cmidrule(l){12-14}
\textbf{Model} & \multicolumn{1}{c}{2} & \multicolumn{1}{c}{4} & \multicolumn{1}{c}{6} & \multicolumn{1}{c}{8} & \multicolumn{1}{c}{10} & \multicolumn{1}{c}{12} & \multicolumn{1}{c}{14} & \multicolumn{1}{c}{16} & \multicolumn{1}{c}{18} & \multicolumn{1}{c}{20} & \% Critical & Avg Drop & Avg Crit Drop \\
\cmidrule(r){1-1}\cmidrule(lr){2-11}\cmidrule(l){12-14}
\gemini\hspace{1pt} 3.1 Pro & \ccol{FFFAF1} 6.5 & \ccol{FEF6E1} 13.2 & \ccol{FEF4D7} 16.4 & \ccol{FEF0CC} 22.0 & \ccol{FEE9CF} 26.6 & \ccol{FEE2D2} 32.1 & \ccol{FEDFD1} 33.9 & \ccol{FDDBCF} 36.2 & \ccol{FDDACF} 36.9 & \ccol{FCD8CE} 38.1 & \ccol{DB9292} 86.3 & \ccol{FEF0CC} 22.2 & \ccol{FEF2CF} 19.2 \\
\claude\hspace{1pt} 4.6 Opus & \ccol{FEF6DE} 13.7 & \ccol{FEF1CC} 20.9 & \ccol{FEE9D0} 27.0 & \ccol{FDDDD0} 35.1 & \ccol{FBD4CC} 40.4 & \ccol{FACEC9} 43.9 & \ccol{F9CAC7} 46.4 & \ccol{F9C9C7} 47.2 & \ccol{F9C7C6} 47.7 & \ccol{F8C3C3} 49.7 & \ccol{DC9292} 86.1 & \ccol{FEE4D2} 30.9 & \ccol{FEE9CF} 26.6 \\
\claude\hspace{1pt} 4.6 Sonnet & \ccol{FEF4D6} 17.0 & \ccol{FEEACF} 25.9 & \ccol{FEE0D1} 33.4 & \ccol{FCD5CC} 40.0 & \ccol{FBD0CA} 42.7 & \ccol{F9C7C6} 48.0 & \ccol{F8C5C5} 48.8 & \ccol{F8C4C4} 49.3 & \ccol{F5BDBD} 52.5 & \ccol{F4BCBC} 53.2 & \ccol{DB9292} 86.4 & \ccol{FCD7CD} 38.6 & \ccol{FEE0D1} 33.4 \\
\openai\hspace{1pt} GPT 5.4 & \ccol{FEF6DE} 13.9 & \ccol{FEEDCD} 23.5 & \ccol{FEE5D2} 30.1 & \ccol{FDD9CE} 37.2 & \ccol{FBD1CA} 42.5 & \ccol{F9C9C7} 47.2 & \ccol{F9C7C6} 48.2 & \ccol{F7C3C3} 50.3 & \ccol{F4BCBC} 53.4 & \ccol{F2B7B7} 55.2 & \ccol{DF9696} 80.9 & \ccol{FEE1D2} 32.7 & \ccol{FEEACF} 26.4 \\
\openai\hspace{1pt} GPT 5.2 & \ccol{FEF4D9} 15.8 & \ccol{FEE9D0} 27.1 & \ccol{FDDCD0} 35.7 & \ccol{FBD1CA} 42.3 & \ccol{F9C7C6} 48.3 & \ccol{F5BDBD} 52.5 & \ccol{F3B9B9} 54.6 & \ccol{F1B4B4} 57.0 & \ccol{EEAEAE} 59.2 & \ccol{EDABAB} 60.7 & \ccol{DD9494} 83.4 & \ccol{FCD6CD} 39.2 & \ccol{FEE1D2} 32.7 \\
\xai\hspace{1pt} Grok 4 & \ccol{FEF5DB} 14.9 & \ccol{FEECCE} 24.6 & \ccol{FEDFD1} 34.2 & \ccol{FCD5CC} 39.9 & \ccol{F9C9C7} 46.6 & \ccol{F8C5C5} 48.8 & \ccol{F4BBBB} 53.8 & \ccol{F0B3B3} 57.3 & \ccol{EFB0B0} 58.6 & \ccol{ECAAAA} 61.1 & \ccol{D88F8F} 92.0 & \ccol{F9C8C7} 47.5 & \ccol{FACFCA} 43.7 \\
\moonshot\hspace{1pt} Kimi K2.5 & \ccol{FEF3D1} 18.5 & \ccol{FEEACF} 26.2 & \ccol{FEDFD1} 33.7 & \ccol{FBCFCA} 43.3 & \ccol{F9C7C6} 48.3 & \ccol{F4BCBC} 53.1 & \ccol{F2B7B7} 55.1 & \ccol{EFB1B1} 57.9 & \ccol{EDACAC} 60.3 & \ccol{ECA9A9} 61.3 & \ccol{DB9292} 87.2 & \ccol{FBD0CA} 42.8 & \ccol{FDD9CE} 37.3 \\
\openai\hspace{1pt} GPT 5.1 & \ccol{FEF0CC} 21.9 & \ccol{FDDBCF} 36.5 & \ccol{FACDC9} 44.9 & \ccol{F6C0C0} 51.4 & \ccol{EFB1B1} 58.1 & \ccol{EBA8A8} 61.9 & \ccol{E9A3A3} 64.4 & \ccol{E8A0A0} 65.9 & \ccol{E79F9F} 67.3 & \ccol{E69D9D} 68.8 & \ccol{DD9494} 84.1 & \ccol{FACCC8} 45.3 & \ccol{FCD8CE} 38.1 \\
\openai\hspace{1pt} GPT 5 & \ccol{FEF5DB} 14.9 & \ccol{FEE6D1} 29.4 & \ccol{FBD1CB} 42.1 & \ccol{F7C2C2} 50.7 & \ccol{F0B3B3} 57.4 & \ccol{ECA9A9} 61.4 & \ccol{E9A2A2} 64.5 & \ccol{E79F9F} 67.2 & \ccol{E69D9D} 69.1 & \ccol{E49C9C} 71.5 & \ccol{D88E8E} 92.9 & \ccol{F2B8B8} 54.9 & \ccol{F6C1C1} 51.0 \\
\openai\hspace{1pt} GPT 5 Mini & \ccol{FEEDCD} 23.5 & \ccol{FBD3CB} 41.1 & \ccol{F5BEBE} 52.3 & \ccol{EDACAC} 60.4 & \ccol{E79F9F} 66.8 & \ccol{E49B9B} 72.0 & \ccol{E29A9A} 74.5 & \ccol{E29999} 75.3 & \ccol{E19999} 76.3 & \ccol{E19898} 77.1 & \ccol{D98F8F} 90.8 & \ccol{EBA8A8} 61.8 & \ccol{F1B5B5} 56.1 \\
\openai\hspace{1pt} GPT 5 Chat & \ccol{FEE4D2} 31.1 & \ccol{F9C7C6} 48.0 & \ccol{EFB1B1} 58.1 & \ccol{E8A1A1} 64.9 & \ccol{E69D9D} 69.0 & \ccol{E39B9B} 72.8 & \ccol{E29A9A} 74.4 & \ccol{E29999} 76.0 & \ccol{E19898} 77.2 & \ccol{E09898} 77.8 & \ccol{DA9090} 89.2 & \ccol{EEAEAE} 59.3 & \ccol{F4BCBC} 52.9 \\
\openai\hspace{1pt} o1 & \ccol{FEE7D1} 28.7 & \ccol{FACBC8} 45.9 & \ccol{F2B7B7} 55.2 & \ccol{EBA7A7} 62.7 & \ccol{E69E9E} 68.5 & \ccol{E59C9C} 70.4 & \ccol{E49B9B} 72.0 & \ccol{E29999} 75.3 & \ccol{E19898} 77.2 & \ccol{E09898} 78.0 & \ccol{D99090} 90.2 & \ccol{EDABAB} 60.8 & \ccol{F2B8B8} 54.9 \\
\openai\hspace{1pt} o3 & \ccol{FEE3D2} 31.4 & \ccol{FACBC8} 45.6 & \ccol{EFB0B0} 58.6 & \ccol{E9A2A2} 64.7 & \ccol{E59D9D} 69.3 & \ccol{E39A9A} 73.9 & \ccol{E29999} 75.9 & \ccol{E19898} 77.4 & \ccol{E09898} 77.8 & \ccol{E09797} 79.1 & \ccol{D88F8F} 91.5 & \ccol{EEAEAE} 59.1 & \ccol{F3BABA} 54.0 \\
\openai\hspace{1pt} GPT 4.1 & \ccol{FEEBCE} 25.3 & \ccol{FBD3CC} 40.7 & \ccol{F3B9B9} 54.5 & \ccol{ECA9A9} 61.7 & \ccol{E59D9D} 69.2 & \ccol{E49B9B} 72.6 & \ccol{E29999} 75.3 & \ccol{E19898} 77.6 & \ccol{E09797} 78.9 & \ccol{DF9797} 79.5 & \ccol{DB9292} 86.6 & \ccol{F0B2B2} 57.5 & \ccol{F8C3C3} 49.8 \\
\gemini\hspace{1pt} 3 Flash & \ccol{FEE4D2} 30.5 & \ccol{F9C8C7} 47.7 & \ccol{F0B3B3} 57.2 & \ccol{E8A0A0} 65.9 & \ccol{E69E9E} 68.6 & \ccol{E49C9C} 71.8 & \ccol{E39B9B} 73.4 & \ccol{E19999} 76.4 & \ccol{E09797} 79.2 & \ccol{DF9696} 80.6 & \ccol{D68C8C} 95.7 & \ccol{E49C9C} 71.5 & \ccol{E69E9E} 68.4 \\
\mistral\hspace{1pt} Large 3 & \ccol{FEDFD1} 34.2 & \ccol{F6C1C1} 51.0 & \ccol{EAA5A5} 63.3 & \ccol{E69D9D} 68.8 & \ccol{E19898} 77.1 & \ccol{E09797} 79.2 & \ccol{DF9696} 81.2 & \ccol{DD9494} 83.7 & \ccol{DC9393} 84.8 & \ccol{DC9393} 85.8 & \ccol{D99090} 90.5 & \ccol{E39B9B} 72.9 & \ccol{E8A0A0} 66.0 \\
\openai\hspace{1pt} OSS 120B & \ccol{F9C9C7} 47.1 & \ccol{E59D9D} 70.3 & \ccol{DF9696} 80.1 & \ccol{DA9191} 88.3 & \ccol{D99090} 90.0 & \ccol{D88E8E} 92.6 & \ccol{D88F8F} 91.6 & \ccol{D88F8F} 92.0 & \ccol{D88E8E} 92.4 & \ccol{D88E8E} 92.9 & \ccol{D68D8D} 95.0 & \ccol{DA9191} 88.6 & \ccol{DD9494} 84.2 \\
\openai\hspace{1pt} GPT 4o & \ccol{E19898} 77.2 & \ccol{DA9090} 89.2 & \ccol{D88F8F} 92.1 & \ccol{D68D8D} 94.7 & \ccol{D68C8C} 95.4 & \ccol{D68C8C} 96.0 & \ccol{D68C8C} 96.1 & \ccol{D68C8C} 96.4 & \ccol{D68C8C} 96.4 & \ccol{D68C8C} 96.4 & \ccol{D68C8C} 95.7 & \ccol{D78E8E} 93.7 & \ccol{D99090} 89.6 \\
\openai\hspace{1pt} GPT 5 Nano & \ccol{DE9595} 81.8 & \ccol{D88E8E} 92.9 & \ccol{D68C8C} 96.0 & \ccol{D58C8C} 96.7 & \ccol{D58B8B} 96.9 & \ccol{D58B8B} 96.9 & \ccol{D58B8B} 97.0 & \ccol{D58B8B} 97.1 & \ccol{D58B8B} 97.4 & \ccol{D58B8B} 97.2 & \ccol{D58B8B} 97.0 & \ccol{D68C8C} 96.3 & \ccol{D78E8E} 93.5 \\
\bottomrule
\end{tabular}
}

%% file: tables/domain_characteristics.tex
\centering
\resizebox{\linewidth}{!}{%
\renewcommand{\arraystretch}{1.6}
\begin{tabular}{lr/rrrrr}
 & \multicolumn{1}{c}{} & \multicolumn{5}{c}{\textbf{Document Properties (mean)}} \\
\cmidrule(l){3-7}
\textbf{Category} & Score & Nat. & Num. & Vocab & Rep. & Struct. \\
\cmidrule(r){1-1}\cmidrule(lr){2-2}\cmidrule(l){3-7}
Science \& Engineering & \ccol{FFFBF4} 97.29 & 0.54 & 0.17 & 0.22 & 0.07 & 0.18 \\
Code \& Configuration & \ccol{FEF3D0} 95.56 & 0.50 & 0.08 & 0.21 & 0.05 & 0.14 \\
Creative \& Media & \ccol{FEE8D0} 94.36 & 0.64 & 0.13 & 0.25 & 0.04 & 0.15 \\
Structured Records & \ccol{F8C3C3} 91.52 & 0.56 & 0.13 & 0.24 & 0.04 & 0.17 \\
Everyday & \ccol{EBA7A7} 89.87 & 0.60 & 0.10 & 0.25 & 0.03 & 0.14 \\
\bottomrule
\end{tabular}
}

%% file: tables/model_details.tex
\centering
\begin{tabular}{@{}lllc@{}}
\toprule
\textbf{Paper Name} & \textbf{Provider} & \textbf{Model ID} & \textbf{Ref.} \\
\midrule
\openai\hspace{1pt} GPT 4o & Azure & \texttt{gpt-4o\_2024-11-20} & \cite{Hurst2024GPT4oSC} \\
\openai\hspace{1pt} GPT 4.1 & Azure & \texttt{gpt-4.1\_2025-04-14} & \cite{Singh2025OpenAIGS} \\
\openai\hspace{1pt} GPT 5 Nano & Azure & \texttt{gpt-5-nano\_2025-08-07} & \cite{El-Kishky2024OpenAIOS} \\
\openai\hspace{1pt} GPT 5 Mini & Azure & \texttt{gpt-5-mini\_2025-08-07} & \cite{El-Kishky2024OpenAIOS} \\
\openai\hspace{1pt} GPT 5 Chat & Azure & \texttt{gpt-5-chat\_2025-10-03} & \cite{El-Kishky2024OpenAIOS} \\
\openai\hspace{1pt} GPT 5 & Azure & \texttt{gpt-5\_2025-08-07} & \cite{El-Kishky2024OpenAIOS} \\
\openai\hspace{1pt} GPT 5.1 & Azure & \texttt{gpt-5.1\_2025-11-13} & \cite{Singh2025OpenAIGS} \\
\openai\hspace{1pt} GPT 5.2 & Azure & \texttt{gpt-5.2\_2025-12-11} & \cite{Singh2025OpenAIGS} \\
\openai\hspace{1pt} GPT 5.4 & Azure & \texttt{gpt-5.4\_2026-03-05} & \cite{Singh2025OpenAIGS} \\
\openai\hspace{1pt} o1 & Azure & \texttt{o1\_2024-12-17} & \cite{Hurst2024GPT4oSC} \\
\openai\hspace{1pt} o3 & Azure & \texttt{o3\_2025-04-16} & \cite{Singh2025OpenAIGS} \\
\openai\hspace{1pt} OSS 120B & Azure & \texttt{gpt-oss-120b\_1} & \cite{El-Kishky2024OpenAIOS} \\
\midrule
\claude\hspace{1pt} 4.6 Sonnet & Anthropic & \texttt{claude-sonnet-4-6} & \cite{Anthropic2024Claude3} \\
\claude\hspace{1pt} 4.6 Opus & Anthropic & \texttt{claude-opus-4-6} & \cite{Anthropic2024Claude3} \\
\midrule
\gemini\hspace{1pt} 3 Flash & Google & \texttt{gemini-3-flash-preview} & \cite{Comanici2025Gemini2P} \\
\gemini\hspace{1pt} 3.1 Pro & Google & \texttt{gemini-3.1-pro-preview} & \cite{Comanici2025Gemini2P} \\
\midrule
\mistral\hspace{1pt} Large 3 & Azure & \texttt{Mistral-Large-3\_1} & \cite{Jiang2023Mistral7} \\
\midrule
\xai\hspace{1pt} Grok 4 & Azure & \texttt{grok-4\_1} & \cite{xAI2025Grok} \\
\midrule
\moonshot\hspace{1pt} Kimi K2.5 & DeepInfra & \texttt{moonshotai/Kimi-K2.5} & \cite{Team2025KimiKS} \\
\bottomrule
\end{tabular}

%% file: tables/document_desiderata.tex
\centering
\begin{tabular}{@{}lp{5cm}p{7.5cm}@{}}
\toprule
\textbf{ID} & \textbf{Desideratum} & \textbf{Description} \\
\midrule
D1 & Unencoded & The document should be unencoded and readable in plain text by the LLM. \\
D2 & Bounded Length & The document should be 2--5k tokens in length. \\
D3 & Semi-Structured & The document should be a mix of natural language and structured data or text. \\
D4 & Public \& Sharable & The document should be downloadable from online sources and publicly sharable. \\
D5 & Realistic & The document should consist of real-world data and content, not synthetically generated material. \\
D6 & Not Memorized & The document should not be overly famous, to avoid LLMs having memorized it (e.g., recent content is preferred). \\
D7 & Not Pedagogical & The document should avoid toy examples from libraries intended solely for learning and introductory purposes. \\
D8 & Minimal Comments & Comments, if present, should account for at most ${\sim}10\%$ of the content to avoid reducing the effective context size. \\
\bottomrule
\end{tabular}